%% file: ociamthesismain.tex
\documentclass[12pt]{ociamthesis}  % default square logo 

\usepackage{natbib}
\PassOptionsToPackage{numbers, compress}{natbib}
\usepackage{amssymb}
\usepackage{enumerate}
\usepackage{wrapfig, framed}
\usepackage[acronym]{glossaries}
\usepackage{authblk}
\usepackage{tikz}

\usepackage{lipsum}
\usepackage{hyperref}
\usepackage{url}
\usepackage{sidecap}
\usepackage{amsmath}

\usepackage{graphicx}
\usepackage{adjustbox}
\usepackage{caption,subcaption}
\usepackage{bbold}
\usepackage{cleveref}
\usepackage{bm}
\usepackage{booktabs}  
\usepackage{soul, color}
\usepackage{nicefrac}       % compact symbols for 1/2, etc.
\usepackage{microtype}      % microtypography
\include{macros}
\usepackage[normalem]{ulem}
\usepackage[titletoc]{appendix}

\newcommand\vartextvisiblespace[1][1.5em]{%
  \makebox[#1]{%
    \kern.07em
    \vrule height.3ex
    \hrulefill
    \vrule height.3ex
    \kern.07em
  }% <-- don't forget this one!
}

\usepackage{listings}
\usepackage{paralist}

\newcommand{\vf}{\textsc{SelPredVerif}}
\newcommand{\incf}{\textsc{InconsistVerif}}

\newcommand{\infersent}{\textsc{BiLSTM-Max}}
\newcommand{\inferSentAutoencoder}{\textsc{BiLSTM-Max-AutoEnc}}
\newcommand{\expzero}{\textsc{Hyp2Expl}}
\newcommand{\expzerol}{\textsc{Hyp2Lbl}}
\newcommand{\expone}{\textsc{BiLSTM-Max-PredExpl}}
\newcommand{\exptwo}{\textsc{BiLSTM-Max-ExplPred}}
\newcommand{\exptwoseqtoseq}{\textsc{BiLSTM-Max-ExplPred-Seq2Seq}}
\newcommand{\exptwoattention}{\textsc{BiLSTM-Max-ExplPred-Att}}

\newcommand{\rvj}{\textsc{RevExpl}}
\newcommand{\jtl}{\textsc{ExplToLbl}}

\newcommand{\sel}{\text{\textbf{S}}^m_{\textbf{x}}}
\newcommand{\scr}{\text{\textbf{SIR}}^m_{\textbf{x}}}
\newcommand{\nsel}{\text{\textbf{N}}^m_{\textbf{x}}}
\newcommand{\D}{\mathcal{D}}
\newcommand{\G}{\mathcal{G}^m}
\newcommand{\ga}{\mathcal{G}^a}
\newcommand{\sela}{\text{\textbf{S}}^a_{\textbf{x}}}
\newcommand{\scra}{\text{\textbf{SIR}}^a_{\textbf{x}}}
\newcommand{\nsela}{\text{\textbf{N}}^a_{\textbf{x}}}
\newcommand{\bfx}{\textbf{x}}

\newcommand{\bfr}{\mathbf{\hat{x}}}
\newcommand{\bfe}{\mathbf{e}}

\newcommand{\bfi}{\mathbf{\hat{e}}}

\newcommand{\bfxd}{\mathbf{x}_{c}}
\newcommand{\bfxi}{\mathbf{x}_{v}}
\newcommand{\bfri}{\mathbf{\hat{x}}_{v}}

\newcommand{\eg}{e.g.,}
\newcommand{\ie}{i.e.,}
\makeglossaries
 
\newacronym{gcd}{GCD}{Greatest Common Divisor}
 
\newacronym{lcm}{LCM}{Least Common Multiple}

\title{Explaining Deep Neural Networks}   %note \\[1ex] is a line break in the title

\author{Oana-Maria Camburu}             %your name
\college{Linacre College}  %your college

\degree{Doctor of Philosophy}     %the degree
\degreedate{Hilary 2020}         %the degree date

\begin{document}

%this baselineskip gives sufficient line spacing for an examiner to easily
%markup the thesis with comments
\baselineskip=22pt plus1pt % original 18pt plus1pt
%\linespread{1.6} %Oana added

%set the number of sectioning levels that get number and appear in the contents
\setcounter{secnumdepth}{3}
\setcounter{tocdepth}{3}

\maketitle                  % create a title page from the preamble info
\include{dedication}        % include a dedication.tex file
\include{acknowlegements}   % include an acknowledgements.tex file
\include{abstract}          % include the abstract
\include{publications}

\begin{romanpages}          % start roman page numbering
\tableofcontents            % generate and include a table of contents
\listoffigures              % generate and include a list of figures
\end{romanpages}            % end roman page numbering

%now include the files of latex for each of the chapters etc
%\include{acknowlegements}
%\include{abstract}
%\include{journey}

\include{introduction}
\include{preliminary}

\include{difficulty-explain}
\include{verify}

\include{esnli}
\include{inconsistencies}
\include{conclusions}

%now enable appendix numbering format and include any appendices
%\appendix
\begin{appendices}
\include{appendix0}
\include{appendix_esnli}
\include{appendix_inconsist}
\end{appendices}

%next line adds the Bibliography to the contents page
\addcontentsline{toc}{chapter}{Bibliography}
%uncomment next line to change bibliography name to references

\bibliographystyle{apalike}  %use the plain bibliography style
\bibliography{refs}        %use a bibtex bibliography file refs.bib
\end{document}

%% file: dedication.tex
\begin{dedication}
%This thesis is dedicated t
To my parents, Geta and Marin, for their enormous love and support. 
Parint\c{i}lor mei, Geta si Marin, pentru dragostea si sprijinul lor nem\u{a}rginit.
\end{dedication}

%% file: acknowlegements.tex
\begin{acknowledgements}
I am most grateful to my advisors, Phil Blunsom and Thomas Lukasiewicz, for their ongoing guidance and support. They always provided useful advice and perspective, forming me as a researcher.
I am also very grateful to Nando de Freitas for his guidance at the beginning of my DPhil. 

I am also very thankful to all my collaborators, and in particular to Ana-Maria Cre\c{t}u, Jakob Foerster, Eleonora Giunchiglia, Vid Kocijan, Pasquale Minervini, Tim Rockt\"aschel, and Brendan Shillingford (in alphabetical order) for the many inspiring research conversations.

I am grateful to my DPhil examiners, Shimon Whiteson and Kyunghyun Cho, for their careful examination and helpful comments on this thesis.

I also want to thank the Graduate Studies administrators, Julie Sheppard and Sarah Retz-Jones, for their ongoing helpful advice. 

I am also very grateful to Linacre College for becoming my home away from home, for the wonderful friends I made in college, and for the support and care I received from all the college staff, in particular, from Jane Hoverd and Karren Morris.

This work has been supported by an Engineering and Physical Sciences Research Council Scholarship and a J.P. Morgan PhD Fellowship.

Lastly, my gratitude goes to my parents, Geta and Marin, who always believed in me, encouraged me to persevere, and supported me unconditionally, to my mentoring family from the \'Ecole Polytechnique, Edouard and Ioana, who have been like a second family for me, and to all my friends who have been there for me throughout this journey. 
\end{acknowledgements}

%% file: abstract.tex
\begin{abstract}

Deep neural networks are becoming more and more popular due to their revolutionary success in diverse areas, such as computer vision, natural language processing, and speech recognition. However, the decision-making processes of these models are generally not interpretable to users. In various domains, such as healthcare, finance, or law, it is critical to know the reasons behind a decision made by an artificial intelligence system. Therefore, several directions for explaining neural models have recently been explored. 

In this thesis, I investigate two major directions for explaining deep neural networks. The first direction consists of feature-based post-hoc explanatory methods, that is, methods that aim to explain an already trained and fixed model (post-hoc), and that provide explanations in terms of input features, such as tokens for text and superpixels for images (feature-based). The second direction consists of self-explanatory neural models that generate natural language explanations, that is, models that have a built-in module that generates explanations for the predictions of the model. The contributions in these directions are as follows. 

First, I reveal certain difficulties of explaining even trivial models using only input features. I show that, despite the apparent implicit assumption that explanatory methods should look for one specific ground-truth feature-based explanation, there is often more than one such explanation for a prediction. I also show that two prevalent classes of explanatory methods target different types of ground-truth explanations without explicitly mentioning it. Moreover, I show that, sometimes, neither of these explanations is enough to provide a complete view of a decision-making process on an instance. 

Second, I introduce a framework for automatically verifying the faithfulness with which feature-based post-hoc explanatory methods describe the decision-making processes of the models that they aim to explain. This framework relies on the use of a particular type of model that is expected to provide insight into its decision-making process. I analyse potential limitations of this approach and introduce ways to alleviate them. 
The introduced verification framework is generic and can be instantiated on different tasks and domains to provide off-the-shelf sanity tests that can be used to test feature-based post-hoc explanatory methods. I instantiate this framework on a task of sentiment analysis and provide sanity tests\footnote{The sanity tests are available at \\ \url{https://github.com/OanaMariaCamburu/CanITrustTheExplainer}.} on which I present the performances of three popular explanatory methods. 

Third, to explore the direction of self-explanatory neural models that generate natural language explanations for their predictions, I collected a large dataset of $\sim\!\!570$K human-written natural language explanations on top of the influential Stanford Natural Language Inference (SNLI) dataset. I call this explanation-augmented dataset e-SNLI.\footnote{The dataset is publicly available at \url{https://github.com/OanaMariaCamburu/e-SNLI}.} 
I do a series of experiments that investigate both the capabilities of neural models to generate correct natural language explanations at test time, and the benefits of providing natural language explanations at training time.

Fourth, I show that current self-explanatory models that generate natural language explanations for their own predictions may generate inconsistent explanations, such as ``There is a dog in the image.'' and ``There is no dog in the [same] image.''. Inconsistent explanations reveal either that the explanations are not faithfully describing the decision-making process of the model or that the model learned a flawed decision-making process. 
I introduce a simple yet effective adversarial framework for sanity checking models against the generation of inconsistent natural language explanations. Moreover, as part of the framework, I address the problem of adversarial attacks with exact target sequences, a scenario that was not previously addressed in sequence-to-sequence attacks, and which can be useful for other tasks in natural language processing. I apply the framework on a state of the art neural model on e-SNLI and show that this model can generate a significant number of inconsistencies.

This work paves the way for obtaining more robust neural models accompanied by faithful explanations for their predictions.

\end{abstract}

%% file: publications.tex
\begin{publications}

This thesis is based on the following publications: %, which will be referred by their Roman numerals:
% \begin{enumerate}[I.]
%     \item \label{publication-esnli} Oana-Maria Camburu, Tim Rockt\"aschel, Thomas Lukasiewicz, Phil Blunsom. \textit{e-SNLI: Natural Language Inference with Natural Language Explanations.} In NeurIPS, 2018.
    
%      \item \label{publication-inconsist} Oana-Maria Camburu, Brendan Shillingford, Pasquale Minervini, Thomas Lukasiewicz, Phil Blunsom. \textit{Make Up Your Mind! Adversarial Generation of Inconsistent Natural Language Explanations.} In ACL, 2020.
    
%     \item \label{publication-verify} Oana-Maria Camburu\footnote{Equal first author.}, Eleonora Giunchiglia\footnotemark[3], Jakob Foerster, Thomas Lukasiewicz, Phil Blunsom. \textit{Can I Trust the Explainer? Verifying Post-hoc Explanatory Methods.} In Workshop on Safety and Robustness in Decision Making at NeurIPS, 2019.
    
% \end{enumerate}

\begin{itemize}
\item[\citep{esnli}:] Oana-Maria Camburu, Tim Rockt\"aschel, Thomas Lukasiewicz, Phil Blunsom. \textit{e-SNLI: Natural Language Inference with Natural Language Explanations.} In Advances in Neural Information Processing Systems 31, 2018.

\item[\citep{verify}:] Oana-Maria Camburu\footnote{Equal first author.} Eleonora Giunchiglia\footnotemark[3], Jakob Foerster, Thomas Lukasiewicz, Phil Blunsom. \textit{Can I Trust the Explainer? Verifying Post-hoc Explanatory Methods}. In the Neural Information Processing Systems (NeruIPS) Workshop Safety and Robustness in Decision Making, 2019.

\item[\citep{inconsistencies}:]  Oana-Maria Camburu, Brendan Shillingford, \\ Pasquale Minervini, Thomas Lukasiewicz, Phil Blunsom. \textit{Make Up Your Mind! Adversarial Generation of Inconsistent Natural Language Explanations} In Proceedings of the Annual Meeting of the Association for Computational Linguistics (ACL), 2020.

\end{itemize}

During my DPhil, I also co-authored the publications below, which are not incorporated in this thesis:

\begin{itemize}
    \item[\citep{Mao_2016_CVPR}:] Junhua Mao, Jonathan Huang, Alexander Toshev, Oana-Maria Camburu, Alan L. Yuille, Kevin Murphy. \textit{Generation and Comprehension of Unambiguous Object Descriptions.} In Proceedings of the IEEE Conference on Computer Vision and Pattern Recognition (CVPR), 2016.
    
    \item[\citep{vid-emnlp}:] Vid Kocijan, Oana-Maria Camburu, Ana-Maria Cre\c{t}u, Yordan Yordanov, Phil Blunsom, Thomas Lukasiewicz. \textit{WikiCREM: A Large Unsupervised Corpus for Coreference Resolution.} In Proceedings of the Conference on Empirical Methods in Natural Language Processing (EMNLP), 2019.
    
    \item[\citep{kocijan-etal-2019-surprisingly}:] Vid Kocijan, Ana-Maria Cre\c{t}u, Oana-Maria Camburu, Yordan Yordanov, Thomas Lukasiewicz. \textit{A Surprisingly Robust Trick for Winograd Schema Challenge.} In Proceedings of the Annual Meeting of the Association for Computational Linguistics (ACL), 2019.
    
    \item[\citep{yordan}:] Yordan Yordanov, Oana-Maria Camburu, Vid Kocijan, Thomas Lukasiewicz. \textit{Does the Objective Matter? Comparing Training Objectives for Pronoun Resolution.} In Proceedings of the Conference on Empirical Methods in Natural Language Processing (EMNLP), 2020.
    
    \item[\citep{do2020esnlive20}:] Virginie Do, Oana-Maria Camburu, Zeynep Akata, \\
    Thomas Lukasiewicz. \textit{e-SNLI-VE-2.0: Corrected Visual-Textual Entailment with Natural Language Explanations.} In the IEEE Computer Vision and Pattern Recognition (CVPR) Workshop on Fair, Data Efficient and Trusted Computer Vision, 2020.

\end{itemize}

Finally, during my DPhil, I also had the pleasure to co-supervise three MSc theses:

\begin{enumerate}[a.]

    \item \textit{Group-Sparse Sentence Representations}.  Thesis written by Ana-Maria Cre\c{t}u for the degree of Master of Science, Department of Computer Science, \'Ecole Polytechnique F\'ed\'erale de Lausanne, 2018.
    
    \item \textit{Natural Language Inference Dataset Generation using Natural Language Explanations}. Thesis written by Vlad Barvinko for the degree of Master of Science, Department of Computer Science, University of Oxford, 2019.
    
    \item \textit{Human-Readable Explanations of Neural Network Predictions}. Thesis written by Virginie Do for the degree of Master of Science, Oxford Internet Institute, University of Oxford, 2019.
\end{enumerate}

\end{publications}

%% file: introduction.tex
\chapter{Introduction}
\label{chap-intro}

Neural networks\footnote{This thesis only addresses artificial neural networks, hence there is no risk of confusion with biological neural networks.} are a class of computing systems designed to learn to perform tasks directly from raw data, such as images or text, without hard-coding task-specific knowledge. These systems have been gaining in popularity over recent years due to their versatility and revolutionary success in various domains. For example, in computer vision, deep neural networks improved the state of the art in object classification by a large margin \citep{Krizhevsky:2012:ICD:2999134.2999257}. Similarly, neural networks have revolutionised acoustic modelling \citep{hinton2012deep} and machine translation \citep{DBLP:journals/corr/SutskeverVL14}, %and other natural language processing tasks \citep{bert}, 
with significant improvements in performance over traditional machine learning methods. Learning without hard-coding task-specific knowledge is a breakthrough in artificial intelligence. For example, AlphaGo \citep{go} learned to play the complex game of Go solely by watching a large number of games played by humans and then repeatedly playing against itself. In 2016, AlphaGo beat the 18-time world champion Lee Sedol 4-1. Furthermore, it has also been shown that neural networks are able to acquire artistic capabilities and to create art. For example, neural networks can recognise artistic styles and transfer them to other images \citep{art}, and they can also generate music \citep{music, music2, music3}.

\section{The Importance of Explanations for Deep Neural Networks}
It has been shown that a key factor in the success of neural networks is their capability to be \textit{deep}, i.e., the fact that successful neural networks can be composed of a very large number of non-linear functions \citep{DBLP:journals/corr/Telgarsky16, DBLP:journals/corr/EldanS15, residual}. Intuitively, multiple layers of non-linear functions allow the network to learn features at various levels of abstraction between the raw data and the prediction. However, this comes at the cost of explainability, since providing a human intelligible interpretation to an intricate composition of a large number of non-linear functions is a difficult open question. Thus, in safety-critical applications, such as health diagnosis, credit allowance, or criminal justice, one may still prefer to employ less accurate but human-interpretable models, such as linear regression and decision trees. % \citep{bostrom2014ethics}. 

The doubts around the decision-making processes of neural networks are justified, as it has been shown that seemingly very accurate such systems can easily rely on spurious correlations in datasets (also called statistical bias, or artefacts) to provide correct answers \citep{pneumonia1, breaking, artifacts, artifacts-CNN, rightforwrong}. A notorious example is that of predicting dire outcomes for patients with pneumonia, on which neural networks outperformed traditional methods by a wide margin~\citep{pneumonia1}. However, it turned out that the training set contained the pattern that patients with a history of asthma were at lower risk of dying from pneumonia \citep{caruana-pneumonia}. This pattern appeared because asthmatic patients were given quicker and higher attention (hence, they had lower mortality rate), because they were, in fact, at higher risk. %The quicker and more intensive care received by these patients effectively reduced their mortality rates compared to the general population.
Obviously, it would be very dangerous, in practice, to use a model that relies on such correlations. 

Another source of mistrust in black-box systems comes from the potential subjective bias that these systems might develop, such as racism, sexism, or other kinds of discrimination and subjectivity \citep{skin}. For example, \citet{compas-fairness} cast doubt on the fairness of the widely used commercial risk assessment software COMPAS for recidivism prediction. Such biases may be learned either from under-representation or irrelevant statistical correlations in the datasets that we train and test our models on. For example, \citet{sexism} showed that word embeddings exhibit female/male gender stereotypes, while \citet{gap} showed that existing corpora and systems favour masculine entities. 

Moreover, a large number of adversarial attacks have shown the fragility of the apparently highly accurate neural networks. For example, image processing neural networks can be fooled into changing \textit{any} of their predictions into \textit{any} other possible prediction only by making imperceptible alterations to the pixels of an image \citep{adv-img1, adv-img2}. Adversarial attacks in neural networks have also had significant success rates in other domains, such as natural language processing \citep{adv-text1, adv-text2} and speech recognition \citep{adv-speech1}. The fragility of deep neural networks revealed by adversarial attacks casts doubt on the underlying learned decision-making processes of these methods. 

Therefore, for neural network systems to garner widespread public trust, as well as to ensure that these systems are indeed fair, we must have human-intelligible explanations for the decisions of these models \citep{role-trust}. 

Firstly, explanations are necessary for providing justifications to end-users for whom the decisions are being taken, such as patients and clients. Not only is it ethical to provide such justifications, but, in some countries, it is even required by law, for example with the introduction of the ``Right to explanation'' in GDPR 2016.\footnote{\url{https://www.privacy-regulation.eu/en/r71.htm}}

Secondly, explanations are useful for the employers of these systems such as doctors and judges, to better understand a system's strengths and limitations and to appropriately trust and employ the system's predictions.

Thirdly, explanations provide an excellent source of knowledge discovery. It is well known that neural networks are particularly good in finding patterns in data. Therefore, being able to explain the algorithms learned by neural networks may result in revealing valuable knowledge that would otherwise be difficult for humans to mine from extremely large datasets. For example, \citet{knowledge-discovery-bio} use decision trees to extract knowledge about conservation biology from trained neural networks. %A review on rule extraction algorithms for deep neural networks is given in \citet{DBLP:journals/corr/Hailesilassie16}.

Finally, being able to explain neural networks can help the developers/researchers of these methods to diagnose and further improve the systems. For example, \citet{lime} showed that, after seeing a number of explanations, even non-experts in machine learning were able to detect which words should be eliminated from a dataset in order to improve an untrustworthy classifier. 

Given the above-mentioned usages of explanations, there is currently an increasing demand for human-intelligible explanations for the predictions of artificial intelligent systems, which has been the motivation to dedicate my thesis to this topic. %In this thesis, I investigate the following question on explaining deep neural networks. %For example, the EU General Data Protection Regulation enforces, among others, the ``\textit{Right to explanation}''\footnote{https://www.privacy-regulation.eu/en/r71.htm} for users impacted by algorithmic decisions. 
%Current research efforts in explainability focus on two directions: (i)~developing post-hoc explanatory methods \citep{lime, shap, l2x, anchors, lrp, deeplift, maple}, i.e., methods that aim to explain already trained neural networks, and (ii)~developing neural networks that intrinsically exhibit a degree of self-explainability \citep{rcnn, invase, cars, zeynep}. This thesis brings contributions in both directions, as we describe below.

%\newpage
%\section{Contributions and Outline}
\section{Research Questions and Outline}
To advance the field of explaining deep neural networks, an appealing research direction is that of developing deep neural networks that provide their own natural language explanations for their decisions \citep{zeynep, math}. This type of model is similar to how humans both learn from explanations and explain their decisions. Indeed, humans do not learn solely from labelled examples supplied by a teacher, but seek a conceptual understanding of a task through both demonstrations and explanations \citep{psy1, psy2}. Moreover, natural language is readily comprehensible to an end-user who needs to assert the reliability of a model. It is also easiest for humans to provide free-form language, eliminating the additional effort of learning to produce formal language, thus making it simpler to collect such datasets. Lastly, natural language justifications might be mined from the existing large-scale free-form text, thus putting to advantage models that are capable of learning from natural language explanations. Therefore, two of the three main research questions I investigate in this thesis are: \textit{Do neural networks improve their behaviour and performance %exhibit improved behaviour 
if they are additionally given natural language explanations for the ground-truth label at training time?} and \textit{Can neural networks generate natural language explanations for their predictions at test time?} 
%exhibit improved behaviour OR improve their learning?

Given the scarcity of datasets of natural language explanations, I first collected a large dataset of ${\sim}570$K instances of human-written free-form natural language explanations on top of SNLI, an existing influential dataset of natural language inference \citep{snli}. I call this explanations-augmented dataset e-SNLI. For example, a natural language explanation for the fact that the sentence ``A woman is walking her dog in the park.'' entails the sentence ``A person is in the park.'' is that ``A woman is a person, and walking in the park implies being in the park.''. I~introduce neural models that learn from these explanations at training time and output such explanations at test time. I also investigate whether learning with the additional signal from natural language explanations can provide benefits in solving other downstream tasks. These contributions will be presented in Chapter \ref{chap-esnli}. 

While natural language explanations generated by a neural model can provide reassurance to users when they display correct argumentation for solving the task, they might not faithfully reflect the decision-making processes of the model. Therefore, in this thesis, I also investigate the following research question: \textit{Can we verify if explanations are faithfully describing the decision-making processes of the deep neural networks that they aim to explain?} To address this question for the class of neural models that generate natural language explanations, two different approaches can be considered. The first approach is to investigate whether the generated natural language explanations are consistent per model. For example, if a model generates explanations that are mutually contradictory, such as ``There is a dog in the image.'' and ``There is no dog in the [same] image.'', then there is a flaw in the model. The flaw can be either in the decision-making process or in the faithfulness with which the generated explanations reflect the decision-making process of the model (or in both). I~introduce a framework that checks whether models that generate natural language explanations can generate inconsistent explanations. As part of this framework, I~address the problem of adversarial attacks for full target sequences, a scenario that had not been previously addressed in sequence-to-sequence attacks and which can be useful for other kinds of tasks. I~instantiate the framework on a state-of-the-art neural model on e-SNLI, and show that this model is capable of generating a significant number of inconsistencies. These contributions will be presented in Chapter \ref{chap-inconsist}.

The second approach for explaining deep neural models consists of developing post-hoc explanatory methods, i.e., external methods that aim to explain already trained neural models. Post-hoc explanatory methods are widespread and currently form the majority of techniques for providing explanations for deep neural networks. While neural networks that generate their own natural language explanations may be corrupted by the same types of biases that cause the need for explaining such models, external explanatory methods may be less prone to these biases. Hence, an interesting approach in verifying the faithfulness of the natural language explanations generated by a model would be to investigate the correlation between the explanations given by an external explanatory method and the natural language explanations generated by the model itself. However, external explanatory methods may also not faithfully explain the decision-making processes of the models that they aim to explain. Therefore, in this thesis, I introduce a verification framework to verify the faithfulness of external explanatory methods. These contributions will be presented in Chapter \ref{chap-verify}. 

In the process of developing the verification framework mentioned above, I identified certain peculiarities of explaining a model using only input features, i.e., units of the input such as tokens for text and superpixels\footnote{A superpixel is a group of pixels with common characteristics, such as pixel intensity or proximity.} for images. For example, I show that sometimes there exists more than one ground-truth explanation for the same prediction of a model, contrary to what current works in the literature seem to imply. I also show that two prevalent classes of post-hoc explanatory methods target distinct ground-truth explanations and I reveal some of their strengths and limitations. These insights can have an important impact on how users choose explanatory methods to best suit their needs, as well as on verifying the faithfulness explanatory methods. These contributions will be presented in Chapter \ref{chap-difficulty}. %\footnote{The contributions are presented here in this order to make the reading more fluent.} %This is because I believe the reading will be more fluent in this way. 

To put into perspective the scope of these contributions, before diving into the contributions of this thesis, I provide background knowledge on explainability for deep neural networks in Chapter \ref{chap-taxo}. 

At the end of each contribution chapter, I present the conclusions and the open questions that the findings in the chapter lead to. Finally, in Chapter \ref{chap-concl}, I provide general conclusions and perspectives on explainability for deep neural networks.

%% file: preliminary.tex
\chapter{Background on Explanatory Methods for Deep Neural Networks}
\label{chap-taxo}
This thesis assumes and does not describe basic knowledge of machine learning and, in particular, of neural networks. 
Readers interested in an in-depth introduction into deep neural networks can see, for example, the book ``Deep Learning'' by \citet{dlbook}. In particular, I recommend Chapter 10 (``Sequence Modeling: Recurrent and Recursive Nets''), since recurrent neural networks are encountered extensively throughout this thesis. 

The main goal of this chapter is to provide the necessary background on the current state of explanatory methods, in order to ease the reading of the rest of the thesis as well as to put into perspective the contributions hereby brought to the field.

\section{Notation}
This section describes the notation used throughout this thesis. 

\paragraph{Variables.} Scalars are usually denoted with non-bold letters, while vectors are in bold. Thus, I use bold lower-case Roman letters, such as $\textbf{x}$, to denote 1-dimensional vectors, which are assumed to be column vectors. A component of a vector is denoted by subscripting the non-bold version of the variable letter with the index of the component. For example, the $i$-th component of the vector $\textbf{x}$ is denoted by $x_i$.

%Scalars are usually denoted with non-bold letters while vectors or higher-dimensional matrices are in bold. Thus, we use bold lower-case Roman letters, such as $\textbf{x}$, to denote 1-dimensional vectors, which are assumed to be column vectors. For higher-dimensional matrices, we use bold upper-case Roman letters, such as $\textbf{W}$. A component of a vector or a matrix is denoted by subscripting the non-bold version of the variable letter with the index or indices, respectively, of the component. For example, the $i$-th component of the vector $\textbf{x}$ is denoted by $x_i$, while $W_{i,j}$ denotes the component of $\textbf{W}$ in its $i$-th row and $j$-th column.

Hyperparameters are usually denoted by lower-case Greek letters. %, i,e., parameters influencing the architecture and training of the neural networks but whose values are set before the training of neural networks. 
For instance, $\mu$ and $\alpha$ typically refer to the learning rate and the coefficient of a loss term, respectively.

%A circumflex, or ``hat'', over a variable denotes a prediction for that variable. For example, $\hat{y}$ would typically be a neural network's prediction of the target $y$. 
%A star as superscript of a variable, e.g., $\lambda*$, (usually over the parameters or hyperparameters of the networks) is typically used to indicate the optimal value of the variable.

\paragraph{Data.} To denote datasets, I use calligraphic, upper-case Roman letters, for example, $\mathcal{D}$. % would typically refer to a dataset. 
To enumerate instances (also called samples or datapoints) in a dataset, I use parenthesised superscripts. For example, the dataset $\mathcal{D} = \{ (\textbf{x}^{(n)}, y^{(n)}) \}_{n=1,N}$ contains $N$ instances $\textbf{x}^{(n)}$ with their associated scalar or categorical targets $y^{(n)}$. Thus, to refer to the $i$-th component of the $n$-th instance in the dataset $\mathcal{D}$, I use $\textbf{x}_i^{(n)}$. Of particular importance in this thesis are datapoints that consist of variable length sequences, such as natural language text. For an input sequence $\textbf{x}$, $x_t$ denotes the $t$-th timestep in the sequence. For example, in a sentence $\textbf{x}$, $x_t$ can denote the $t$-th token or character. For discrete inputs, while the actual input into the neural network at each timestep $t$ is an embedding vector \citep{embed} of the token $x_t$, I use the scalar notation $x_t$ to refer to the raw input at timestep $t$. %, and $\text{embed}(x_t)$ to refer to its vector embedding.

\paragraph{Functions.}
In this thesis, the name of a function does not contain indications of the dimensionalities of its inputs and outputs. These dimensionalities are either stated or inferred from the context. Similarly to vectors, the components of the outputs of functions are referred to by using subscripts placed \textit{before} the argument (if present). For example, the $i$-th component of the vector output of function $f$ at point $\textbf{x}$ is referred to as $f_i(\textbf{x})$.

%A commonly used type of function in neural networks is a component-wise non-linearity, also called activation. I use $\sigma$ as a generic name for any activation function. Another convention is that I denote a loss function used for training neural networks by the calligraphical letter $\mathcal{L}$, which is typically a sum of loss terms. 

A commonly used type of function is the loss function used for training neural networks, which I denote by the uppercase letter $L$, which is typically a sum of loss terms. 

Finally, neural networks themselves are functions. Generic models will be commonly referred as $m$. For readability, I use abbreviations or acronyms as names for the (parts of the) neural networks. For example, the encoder and decoder of a sequence-to-sequence neural network can be referred as $\text{enc}(\cdot)$ and $\text{dec}(\cdot)$, respectively, while a multilayer perceptron can be referred as $\text{MLP}(\cdot)$.

\paragraph{Big O notation.} $O(\cdot)$ denotes the typical big O notation.

\paragraph{Vocabulary.} In this thesis, it is important to make the distinction between three types of algorithms. First, a \textit{target model} refers to a model that we want to explain. Second, an \textit{explanatory method} (also called \textit{explainer}) refers to a method that aims to explain a target model. Third, a \textit{verification framework} refers to a framework for verifying if an explanatory method faithfully explain a target model.

\section{Types of Explanatory Methods}
\label{types-expl}
Recently, an increasing number of diverse works are aiming to shed light on deep neural networks~\citep{lime, shap, lrp, lrp-rec, deeplift, anchors, maple, l2x, cars, zeynep, rcnn, invase}.\footnote{Some of these methods are not restricted to explaining only neural networks.} These methods differ substantially in various ways. 
Different groupings of explainers are presented below. However, I will not present all the possible groups of explanatory methods. This is because there is a large number of groups, and, since the aim of this chapter is to provide background for putting into perspective the rest of the thesis, I do not want to overwhelm the reader with information that would not be further necessary for the scope of this thesis. For the readers interested in more detailed reviews on categorisations of explanatory methods, I refer to the works of \citet{explaining-expls} and \citet{review-expl1}.

\subsection{Post-Hoc versus Self-Explanatory} 
One of the most prominent distinctions among current explanatory methods is to divide them into two types.

\begin{enumerate}[(i)]
    \item \textit{\textbf{Post-hoc explanatory methods}} are stand-alone methods that aim to explain already trained and fixed target models~\citep{lime, shap, l2x, anchors, saliency, lrp, lrp-rec, deeplift, integrated-grads}. For example, LIME \citep{lime} is a post-hoc explanatory method that explains a prediction of a target model by learning an interpretable model, such as a linear regression, on a neighbourhood around the prediction of the model (I~will explain the concept of neighbourhood of an instance in Section \ref{feature-based}). 
    
    \item \textit{\textbf{Self-explanatory models}} are target models which incorporate an explanation generation module into their architecture such that they provide explanations for their own predictions~\citep{rcnn, invase, tommi-neurips-rcnn, cars, zeynep, DBLP:conf/eccv/HendricksARDSD16, math, explainyourself}. At a high level, self-explanatory models have two inter-connected modules: (i)~a \textit{predictor} module, i.e., the part of the model that is dedicated to making a prediction for the task at hand, and (ii)~an \textit{explanation generator} module, i.e., the part of the model that is dedicated to providing the explanation for the prediction made by the predictor. For example, \citet{rcnn} introduced a self-explanatory neural network where the explanation generator selects a subset of the input features, which are then exclusively passed to the predictor that provides the final answer based solely on the selected features. Their model is also regularised such that the selection is short. Thus, the selected features are intended to form the explanation for the prediction. 
    
    Self-explanatory models do not necessarily need to have supervision on the explanations. For example, the models introduced by \citet{rcnn}, \citet{invase}, and \citet{tommi-neurips-rcnn} do not have supervision on the explanations but only at the final prediction. On the other hand, the models introduced by \citet{zeynep}, \citet{cars}, \citet{DBLP:conf/eccv/HendricksARDSD16}, and  \citet{explaining-expls} require explanation-level supervision.

\end{enumerate}

In general, for self-explanatory models, the predictor and explanation generator are trained jointly, hence the presence of the explanation generator is influencing the training of the predictor. This is not the case for post-hoc explanatory methods, which do not influence at all the predictions made by the already trained and fixed target models. Hence, for the cases where the augmentation of a neural network with an additional explanation generator results in a significantly lower task performance than that of the neural network trained only to perform the task, one may prefer to use the latter model followed by a post-hoc explanatory method. On the other hand, it can be the case that enhancing a neural network with an explanation generator and jointly training them results in a better performance on the task at hand. This can potentially be due to the additional guidance in the architecture of the model, or to the extra supervision on the explanations if available. For example, on the task of sentiment analysis, \citet{rcnn} obtained that adding an intermediate explanation generator module without any supervision on the explanations does not hurt performance. On the task of commonsense question answering, \citet{explainyourself} obtained a better performance by a self-explainable model with supervision on the explanations than by a neural network trained only to perform the task. Thus, the two types of explanatory methods have their advantages and disadvantages.

In Chapter \ref{chap-difficulty}, I will present a fundamental difference between two major types of post-hoc explanatory methods. In Chapter \ref{chap-verify}, I will introduce a verification framework for post-hoc explanatory methods which is based on self-explanatory target models. In Chapters \ref{chap-esnli} and \ref{chap-inconsist}, I will focus solely on self-explanatory models.

\subsection{Black-Box versus White-Box}
Another distinction among explanatory methods can be done in terms of the knowledge about the target model that the explainer requires. We have the following two categories.
\begin{enumerate}[(i)]
    \item \textit{\textbf{Black-box/model-agnostic explainers}} are explainers that assume access only to querying the target model on any input. LIME \citep{lime}, Anchors \citep{anchors}, KernalSHAP \citep{shap}, L2X \citep{l2x}, and LS-Tree \citep{lstree} are a few examples of model-agnostic explanatory methods.
    
    \item \textit{\textbf{White-box/model-dependant explainers}} are explainers that assume access to the architecture of the target model. For example, LRP \citep{lrp, lrp-rec}, DeepLIFT \citep{deeplift}, saliency maps \citep{saliency}, integrated gradients \citep{integrated-grads}, GradCAM \citep{gradcam}, and DeepSHAP and MaxSHAP \citep{shap} are a few examples of model-dependant explanatory methods. 
\end{enumerate}

This division mainly applies to post-hoc explanatory methods, since, by default, self-explanatory models exhibit a strong connection between the explanation generator and the predictor.  

Black-box explainers have a larger spectrum of applicability than white-box explainers, because the former can be applied in cases where one does not have access to the internal structure of the target model. Black-box explainers are also usually quicker to use in an off-the-shelf manner, since one does not have to adapt a model-dependant technique to a particular architecture of a model or to a new type of layer. However, these advantages come at the expense of potentially less accurate explanations,  since black-box explainers can infer correlations between inputs and predictions that do not necessarily reflect the true inner workings of the target model. 

The verification framework that I will introduce in Chapter \ref{chap-verify} can be applied to both black-box and white-box explainers.

\subsection{Instance-Wise versus Global} 
Another way of dividing explanatory methods is according to the scope of the explanation. We have the following two types of explainers.

\begin{enumerate}[(i)]
    \item \textit{\textbf{Instance-wise explainers}} provide an explanation for the prediction of the target model on any \textit{individual instance} \citep{lime, shap, l2x, maple, lrp, lrp-rec, saliency, deeplift}. For example, LIME \citep{lime} learns an instance-wise explanation for any instance by training a linear regression on a neighbourhood of the instance. 
    
    \item \textit{\textbf{Global explainers}} explain the high-level inner workings of the entire target model \citep{global-gam, global-distilation, global-partitions}. For example, \citet{global-distilation} provide global explanations by distilling a neural network into a soft decision tree. 
    
\end{enumerate}

Instance-wise explainers are particularly useful, for example, for use cases where end users require an explanation for the decisions taken for their particular case. Global explainers are particularly useful, for example, for quick model diagnostics of possible biases or for knowledge discovery. Since global explainers aim to explain the behaviour of the entire target model, usually via distilling the target model into an interpretable one, they implicitly provide instance-wise explanations as well. However, it is difficult, if not impossible, for an interpretable model to accurately capture all the irregularities learned by a highly non-linear model. Hence, instance-wise explanations derived from global explainers might not always be accurate. The majority of the current works in the literature focus on designing instance-wise explainers.

Conversely, instance-wise explanations can be a starting point for obtaining global explanations. For example, \citet{lime}, and \citet{anchors} introduced sub-modular pick techniques to derive a global explanation for the inner working of the model from instance-wise explanations, while \citet{global-gam} used clustering techniques to provide global explanations for sub-populations starting from instance-wise explanations. 

This division is mainly relevant for post-hoc explanatory methods. By default, self-explanatory models are instance-wise explainers, since the built-in explanation generation module would be applied to each instance. However, the above mentioned techniques for going from instance-wise to global explanations can equally be applied to self-explanatory models. 

This thesis addresses solely instance-wise explainers.

\subsection{Forms of Explanations}
Explanatory methods can also be divided into different groups depending on the form of the explanations that they provide. I will briefly describe below a few commonly used forms of explanations. I will discuss in more detail the feature-based and natural language explanations, since they are the focus of this thesis. % are: feature-based explanations, natural language explanations, concept-based explanations, and example-based explanations. %I only describe in detail the first two forms since they are the focus of this thesis. For the same reason, I also tailor the presentation of these forms of explanations to instance-wise explainers. 
%Four commonly used forms of explanations are: feature-based explanations, natural language explanations, concept-based explanations, and example-based explanations. I only describe in detail the first two forms since they are the focus of this thesis. For the same reason, I also tailor the presentation of these forms of explanations to instance-wise explainers.

%\subsubsection{Feature-Based Explanations}\label{feature-based}
\paragraph{Feature-based explanations.}
Feature-based explanations are currently the most widespread form of explanations and consist of assessing the importance/contribution that each feature of an instance has in the model prediction on that instance. Common features include tokens for text and super-pixels for images. 

There are two major types of feature-based explanations: importance weights and subsets of features. The importance weights explanations provide for each input feature in an instance a real number that represents the contribution of the feature to the prediction of the model on the instance. Subsets explanations provide for each instance the subset of most important features for the prediction of the model on the instance. For example, for a model that predicts that the sentence ``The movie was very good.'' has a positive sentiment of $4$ stars (out of $5$), a subset explanation could be formed of the features \{``very'', ``good''\}, while an importance weights explanation could, for example, attribute to the feature ``good'' an importance weight of $3$ and to the feature ``very'' an importance weight of $1$ (the sum of importance weights matches with the total prediction of $4$; this property is called feature-additivity and will be discussed in detail in the next section). I will describe each of these two types of feature-based explanations in Section \ref{feature-based}. 

There is a large number of explanatory methods providing feature-based explanations. For example, LIME \citep{lime}, SHAP \citep{shap}, L2X \citep{l2x},  LS-Tree \citep{lstree}, Anchors \citep{anchors}, LRP \citep{lrp, lrp-rec}, DeepLIFT \citep{deeplift}, Integrated Gradients \citep{integrated-grads}, and Grad-CAM \citep{gradcam} are just a few of the many feature-based post-hoc explanatory methods, while self-explanatory methods with feature-based explanations include RCNN \citep{rcnn}, INVASE \citep{invase}, CAR \citep{tommi-neurips-rcnn}, as well as the influential class of attention models \citep{attention}, among others.

I investigate feature-based explanations in Chapters \ref{chap-difficulty} and \ref{chap-verify}.

%\subsubsection{Natural Language Explanations}
\paragraph{Natural language explanations.}\label{nle}
Natural language explanations consist of natural language sentences that provide human-like arguments supporting a prediction. For example, a natural language explanation for the fact that the sentence ``A woman is walking her dog in the park.'' entails the sentence ``A person is in the park.'' is that ``A woman is a person, and walking in the park implies being in the park.''. %One may artificially transform feature-based explanations into full sentences via templates, such as ``Feature [X] contributed in the amount of [Y] to the prediction of the model.''. However, such explanations would not qualify as natural language explanations. 

To train explanatory methods to provide natural language explanations, new datasets of human-written natural language explanations have recently been collected to allow supervision on the explanations at training time, as well as to enable evaluation of the correctness of the generated explanations at test time \citep{zeynep, DBLP:conf/eccv/HendricksARDSD16, cars, math, world-tree, explainyourself}. For example, \citet{zeynep} introduced ACT-X and VQA-X, two datasets that contain (besides visual explanations) natural language explanations for the tasks of visual activity recognition and visual question-answering, respectively. Further, \citet{zeynep} also introduced the self-explanatory model PJ-X (Pointing and Justification Explanation), that is trained to jointly provide a prediction, a feature-based explanation, and a natural language explanation, with both explanations supporting the prediction.
Similarly, \citet{cars} introduced the BDD-X (Berkeley DeepDrive eXplanation) dataset consisting of natural language explanations supporting the decisions of a self-driving car. They train a self-explanatory model consisting of a vehicle controller (the predictor) and an explanation generator such that the decisions of a self-driving car are justified to the users with explanations such as ``The car moves back into the left lane because the school bus in front of it is stopping.''. Another relevant work is that of \citet{world-tree}, who provide a dataset of natural language explanation graphs for elementary science questions. However, their corpus is very small, only $1680$ pairs of questions and explanations. Similarly, \citet{math} introduced a dataset of textual explanations for solving mathematical problems. Yet, the focus of this dataset is narrow and it is arguably difficult to transfer to more general natural understanding tasks.

%Other works in the direction of self-explanatory methods with natural language explanations were introduced by \citet{DBLP:conf/eccv/HendricksARDSD16}, \citet{math}, and \citet{explainyourself}. 
While the majority of the natural language explanations methods are self-explanatory models, there are works, such as that of \citet{grounding}, that aim to improve the quality of the natural language explanations in a post-hoc manner. 

%On the other hand, the simple presence of a minimal amount of templating that does not influence the core foundation of the explanation, such as ``\textit{The prediction is [prediction] because [natural language explanation].}'' would also not discard the explanation from being categorized as natural language. Formally, we draw the line between template-based and natural language explanations as follows: if the main arguments of the explanation are generated in natural language by the explanatory methods as opposed to using 

In Chapter \ref{chap-esnli}, I introduce a new large dataset of ${\sim}570$K natural language explanations for the influential task of solving natural language inference \citep{snli}. In addition, I develop models that incorporate natural language explanations into their training process as well as generate such explanations at test time. Further, in Chapter \ref{chap-inconsist}, I draw attention to the fact that such models can generate inconsistent natural language explanations, which exposes flaws in the model. I introduce an adversarial framework for detecting pairs of inconsistent explanations and show that the best model trained in Chapter \ref{chap-esnli} can generate a significant amount of inconsistent explanations.  

%\subsubsection{Others}

\paragraph{Concept-based explanations.} Concept-based explainers aim to quantify the importance of a user-defined high-level concept, such as curls or strips, to the prediction of the model \citep{tcav, cocox}. % and they have been mainly applied to computer vision. %This type of explanation is useful when the 
    
\paragraph{Example-based explanations.} Example-based explainers provide instances from which one can derive insights into the model predictions. %The example-based explanations can also differ in terms of how the examples are supporting the prediction. 
For example, for a given instance, \citet{ex-based} aim to trace which instances from the training set influenced the most the model prediction on the current instance. \citet{ex-complemental} provide examples that show the model capabilities of distinguishing between the current instance and other instances which only have little yet essential differences with respect to the current instance.

\paragraph{Surrogate explanations.} Surrogate explainers aim to provide an interpretable surrogate model of the target model. For example, \citet{meijer} used Meijer G-functions to parametrize their surrogate explainer. %Feature-additivity explainers are also a type of local surrogate explainers. %, with the surrogate model being locally a linear model.

\paragraph{Combinations of forms of explanations.} A single explainer can also provide more than one type of explanations. For example, the PJ-X model introduced by \citet{zeynep} is a self-explanatory model that provides both feature-based and natural language explanations at the same time, and the joint training for both types of explanations was shown to improve each of the two types of explanations. Similarly, \citet{ex-complemental} introduce a self-explainable model that generates both natural language and example-based explanations.

\section{Feature-Based Explanations}\label{feature-based}
As mentioned above, there are two major types of feature-based explanations: importance weights and subsets. I will describe each of these below. Let $m$ be a model and $\textbf{x}$ an instance with a potentially variable number $n$ of features $\textbf{x} = (x_1, x_2, ... , x_n)$ (e.g., $x_i$ is the $i$-th token in the sentence $\textbf{x}$). A feature-based explanation can be seen as a mapping of $\textbf{x}$ to an output space. For importance weights explanations, each feature $x_i$ of $\bfx$ is mapped to a real number. For subsets explanations, $\bfx$ is mapped to a collection of subsets of $\bfx$, each of which is a potential alternative explanation. %the output space is $\mathbb{R}^n$, while for subsets explanations, the output space is the power set of the power set of the features, i.e., $\mathcal{P}(\mathcal{P}(\bfx))$, as we will see below.

\subsection{Importance Weights}\label{importance-weights}
Importance weights explanations attribute to each feature $x_i$ in an instance $\textbf{x}$ a weight $w_i(m, \textbf{x})$ which represents the importance (also called contribution) of feature $x_i$ for the prediction $m(\textbf{x})$ \citep{lime, shap, integrated-grads, lrp, lrp-rec, deeplift, saliency, lstree}. The weights are signed real numbers and the sign indicates whether the feature pulled the model towards the prediction (the same sign as the sign of the prediction) or against the prediction (the opposite of the prediction sign).

%In Chapter \ref{chap-verify} we will present a property, called feature-additivity, that is encountered in the majority of the current importance weights explainers.

\subsubsection{Feature-Additive Weights}\label{feature-additiv}
A property that has been encountered in a large number of the importance weights explainers is that of \textit{feature-additivity}, which requires the sum of the importance weights of all the features present in an instance to be equal to the prediction of the model on that instance minus the bias of the model. For classification tasks, the prediction of a model is considered to be the probability of the predicted class, usually the highest probability among the predicted probabilities of each possible class.
The bias of a model is the model prediction on an input that brings \textit{no information}, usually referred to as the \textit{reference} or \textit{baseline} input. For example, the all-black image is a common baseline in computer vision, while the zero-vector embedding is a common baseline in natural language processing \citep{l2x, integrated-grads}. Similarly, the information brought by any feature in an instance $\textbf{x}$ can be eliminated either by \textit{occlusion} (replacing the feature with a baseline feature, such as a black super-pixel or a zero-vector embedding) or by \textit{omission} (removing the feature completely) \citep{deletion} if that is technically possible, for example, for models that admit variable length input sequences, such as natural language processing models. Both occlusion and omission may result in out-of-distribution inputs that may lead to unreliable importance weights, as mentioned by \citet{integrated-grads}.  

Formally, for a model $m$ and an instance $\textbf{x}$, 
under the feature-additivity constraint, the features $\{x_i\}_i$ are attributed weights $\{w_i(m, \textbf{x})\}_i$ such that
\begin{equation}\label{eq:additiv}
\sum_{i=1}^{|\bfx|} w_i(m, \textbf{x}) = m(\textbf{x}) - m(\textbf{b}),
\end{equation}
where $\textbf{b}$ is the baseline input. For regression models, $m(\textbf{x})$ is the real value predicted by the model on instance $\textbf{x}$, while for classification tasks, $m(\textbf{x})$ is the probability of the predicted class. %While the equality in Equation \ref{eq:additiv} is the main property of feature-additivity, not all feature-additive methods guarantee that this equality is obtained. For example, in order to learn the weights, LIME \citep{lime} learns a linear regression locally on a neighbourhood of each instance $\textbf{x}$, where the neighbourhood is defined by setting each combination of features in the instance $\textbf{x}$ to a baseline value. Formally, for each instance $\bfx$, a \textit{simplified input} is defined as $\textbf{z'} \in \{0,1\}^{|\bfx|}$ such that if $z'_i=1$
%
%Not all feature-additive methods guarantee that learned weights satisfy the equality, even if it is their goal. Hence, equality in Equation \ref{eq:additiv} is also referred, by \citet{shap}, as the property of \textit{local accuracy}.

%The feature-additivity property makes the interpretation of the weights easy for users: if feature $x_i$ is eliminated (replaced by a baseline feature or deleted), then the prediction would decrease by $w_i(f, \textbf{x})$. 
A large number of explanatory methods rely on the feature-additivity property \citep{lime, shap, lrp, lrp-rec, deeplift, saliency, integrated-grads}. For example, LIME \citep{lime} learns a linear regression model locally on a neighbourhood of each instance $\textbf{x}$, where the neighbourhood is defined by all instances formed by setting all possible combinations of features in the instance $\textbf{x}$ to a baseline value. %For example, the neighbourhood of a 

\paragraph{Shapley values.}
\citet{shap} aim to unify the feature-additive explanatory methods by showing that the only set of feature-additive importance weights that verify three properties---\textit{local accuracy}, \textit{missingness}, and \textit{consistency}---are given by the Shapley values from cooperative game theory \citep{shapley1}. 

\begin{enumerate}\label{shapley-conds}
    \item \textit{Local accuracy} (also referred to as completeness in \citet{integrated-grads}) requires that the equality in Equation \ref{eq:additiv} holds. Note that not all feature-additive methods guarantee that the provided importance weights satisfy this equation, even if it is their goal. For example, when LIME learns a linear regression on the neighbourhood of an instance $\bfx$, this linear regression model might not be able to reproduce the same prediction on the instance $\bfx$ that it aims to explain.

    \item \textit{Missingness} requires features that are not present in an instance $\bfx$ to be given zero importance weight for the prediction $m(\bfx)$. For example, for the prediction of model $m$ on sentence $\bfx$, only the tokens present in the sentence $\bfx$ should be part of the explanation and hence can be given a non-zero importance weight, while any other token in the vocabulary that is not present in the sentence $\bfx$ has $0$ importance weight for the prediction $m(\bfx)$.
    
    \item \textit{Consistency} requires that if for two models $m$ and $m'$ the marginal contribution of a feature $x_i$ of an instance $\bfx$ is higher for model $m$ than for model $m'$ for each subset of features of $\bfx$, then the importance weight of $x_i$ for the instance~$\bfx$ should be higher for the model $m$ than for the model $m'$. Formally, if for all subsets of features $\textbf{x}_\textbf{s} \subseteq \bfx \setminus \{x_i\}$ we have that
    \begin{equation}\label{eq-consist}
    m(\textbf{x}_\textbf{s} \cup \{x_i\}) - m(\textbf{x}_\textbf{s}) \ge m'(\textbf{x}_\textbf{s} \cup \{x_i\}) - m'(\textbf{x}_\textbf{s}),
    \end{equation} 
    then $w_i(m, \textbf{x}) \ge w_i(m', \textbf{x})$.
\end{enumerate}

\citet{shap} show that the Shapley values are the only solution which satisfy all three above properties, and can be computed in closed form as follows: %for a model $f$ and an instance $\textbf{x}$ with (a variable number of) features $\textbf{x} = (x_1, x_2, ... , x_n)$, the weight $w^\textbf{x}_i(f)$ of feature $x_i$ should be:
\begin{equation}\label{eq:shapley}
\begin{gathered}
w_i(m, \textbf{x}) = \sum_{\textbf{x}_\textbf{s} \subseteq \textbf{x} \setminus \{x_i\}} \frac{|\textbf{x}_\textbf{s}|! (|\textbf{x}|-|\textbf{x}_\textbf{s}|-1)!}{|\textbf{x}|!} [m(\textbf{x}_\textbf{s} \cup \{x_i\}) - m(\textbf{x}_\textbf{s})]\,, 
\end{gathered}
\end{equation} where the sum enumerates over all subsets $\textbf{x}_\textbf{s}$ of features in $\textbf{x}$ that do not contain feature $x_i$, and $|\cdot|$ denotes the number of features of its argument. For tasks where the order of features matters, the features are kept in the same order as they appeared in $\bfx$. %To compute the model's prediction when certain features are excluded, i.e., the terms $f(\textbf{x'} \cup \{x_i\})$ and $f(\textbf{x'})$, the missing features are replaced with the above-mentioned reference value. The problem of what makes a good reference value has briefly been investigated by 

%\item \textit{feature-ranking}: the set of input features are ranked in order of their importance for the prediction, but the magnitude of the importance is not specified \citep{l2x, anchors}. Trivially, importance weights implies feature-ranking, by ordering the features in terms of the absolute value of their weights.

Since computing the Shapley values in closed form would require exponential time in the number of features, \citet{shap} provide both model-agnostic (KernalSHAP) and model-dependant (DeepSHAP and MaxSHAP) methods for approximating the Shapley values.

%\paragraph{Non-feature-additive importance weights.}
\subsubsection{Non-Feature-Additive Weights}
The majority of the importance weights explainers are based on the feature-additivity property. % which results in explanations being easy to interpret.
%However, the assumption of independence of features is a limitation of this property, as it makes it difficult to emphasize the potential interactions between features. %For example, consider the following models: $m^1(\bfx) = x_1 OR x_2$ and $m^2(\bfx) = x_1 + x_2$. For the instance $\bfx$ with $x_1 = x_2 = 1$ for both models, the importance weights of each feature would be equal ($w_1(m^1, \bfx) = w_2(m^1, \bfx) =0.5$ while )
However, importance weights explainers that do not rely on feature-additivity are also being developed. For example, LS-Tree \citep{lstree} leverages parse trees for linguistic data to assign weights to each feature such that the weights can be used to detect and quantify interactions between the tokens in a sentence. Similarly to the Shapley methods, LS-Tree also takes inspiration from cooperative game theory. However, instead of using the Shapley values, LS-Tree is inspired by the Banzhaf values \citep{banzhaf}.

\subsection{Minimal Sufficient Subsets}\label{sufficient-subsets}
Another popular way of explaining the prediction of a model $m$ on an instance $\textbf{x}$ is to provide a minimal subset of features $\text{mss}(m, \textbf{x}) \subseteq \textbf{x}$ such that these features alone suffice for the same prediction to be reached by the model if the information from all other features is missing \citep{what-made-you-do-this}. Formally,
\begin{equation}\label{eq:mss1}
m(\text{mss}(m, \textbf{x})) = m(\textbf{x}) \text{ and } %\forall \bfx_\textbf{s} \subset \text{mss}(m, \textbf{x}) \text{, we have that } m(\bfx_\textbf{s}) \neq m(\textbf{x}).  
\end{equation}
\begin{equation}\label{eq:mss2}
\forall \bfx_\textbf{s} \subset \text{mss}(m, \textbf{x}) \text{, we have that } m(\bfx_\textbf{s}) \neq m(\textbf{x}).  
\end{equation}

% \begin{equation}\label{eq:mss}
% m(\text{mss}(m, \textbf{x})) \approx m(\textbf{x}) \text{ and } \forall s \subseteq \text{mss}(m, \textbf{x}) \text{ we have that } m(s) \not\approx m(\textbf{x}).  
% \end{equation}

To compute the prediction of a model $m$ on a subset of features $\textbf{x}_\textbf{s}$, one performs either an occlusion or a deletion of the features in $\textbf{x} \setminus \textbf{x}_\textbf{s}$, as explained above.

The minimality condition is necessary to ensure that one does not provide the trivial sufficient subset formed by all features. 

It may not always be the case that a target model relies only on a subset of features, as opposed to needing all the features in an instance. If the whole input is necessary to explain the prediction, then such type of explanation would be uninformative. Nonetheless, it can be the case that a model relies only on a subset of the input features for each instance. For example, in computer vision, arguably not all pixels are essential for a model to classify an object or for providing an answer to a question. Similarly, in sentiment analysis, certain sub-phrases can be enough to identify the sentiment. 

There can be multiple minimal sufficient subsets for one instance. Then, each minimal sufficient subset is one independent potential explanation for the model prediction on the instance.
In this case, to provide a complete view of the model, one would ideally provide the collection of all minimal sufficient subsets for a model $m$ and an instance $\bfx$, i.e., $\mathcal{S}(m, \bfx) = \{ \text{mss}_k(m, \textbf{x})\}_k$.% (hence, the explanation is a mapping from $\bfx$ to $\mathcal{P}(\mathcal{P}(\bfx))$, as mentioned at the beginning of this section). 
In Chapter \ref{chap-difficulty}, I will expose problems with providing certain minimal sufficient subsets as explanations.

Explanatory methods such as L2X \citep{l2x}, SIS \citep{what-made-you-do-this}, Anchors \citep{anchors}, and INVASE \citep{invase} aim to provide this type of explanation. For example, L2X learns a minimal sufficient subset by maximising the mutual information between the prediction of the model on only a subset of features, i.e., $m(\textbf{x}_{\textbf{s}})$, and the prediction on the full instance $m(\textbf{x})$. However, L2X assumes that the cardinality of a minimal sufficient subset is known in advanced and that it is the same for all instances, which is a major limitation in practice which may result in both the minimality and the sufficiency conditions being violated. %Nonetheless, L2X provides a ranking of all the features in the instance, and its goal is to rank the features in a minimal sufficient subset highest. 
In contrast, by using an actor-critic methodology \citep{actor-critic}, INVASE allows the cardinality of the minimal sufficient subsets to differ for each instance. Both L2X and INVASE provide only one minimal sufficient subset. On the other hand, SIS provides a set of minimal sufficient subsets that do not overlap. Thus, SIS might identify only a subset $\mathcal{S'}$ of the full collection $\mathcal{S}(m, \bfx)$. % such that the minimal sufficient subsets $\text{mss}_k \in \mathcal{S'}$ have an empty intersection, i.e., for any two minimal sufficient subsets $\text{mss}_i , \text{mss}_j \in \mathcal{S'}$ we have that $\text{mss}_i \cap \text{mss}_j = \varnothing$. 

In Chapter \ref{chap-difficulty}, I provide more insight into the strengths and limitations of the Shapley-based explanations and the minimal sufficient subsets explanations. 

\section{Properties of Explanations}
%The purpose of having explanations for deep neural networks is that humans understand the decision-making processes of these models. Therefore, there are at least two main properties that explanations should satisfy. First, explanations should be easy for humans to interpret. Second, explanations should be faithfully describing the decision-making processes of the target models that they aim to explain. 
The purpose of having explanations for deep neural networks is that humans understand the decision-making processes of these models. Therefore, there are two main properties that explanations should satisfy: explanations should be easy for humans to interpret and should be faithfully describing the decision-making processes of the target models that they aim to explain.

\paragraph{Easiness to interpret.}
An explanation for a model prediction would not be useful if humans cannot easily interpret it. At the extreme, since neural networks are a composition of functions, one could explain any prediction by providing the full chain of functions applied to the instance, but this explanation would not be of any use to a human.

Each form of explanation has its own interpretation. For example, a minimal sufficient subset explanation is to be interpreted as a subset of features that suffice to lead the model to the same prediction when the information from all the rest of the features is eliminated and no other subset of this minimal sufficient subset would suffice for the model to reach the same conclusion. 

However, not all explanatory methods are easy to interpret by humans. For example, explanations in the form of Shapley values give for each feature a weighted average of the marginal contributions of the feature inside each possible subset of features of the original instance. Arguably, such quantity can be difficult to interpret by humans, who may simply resort to looking at the relative importance of features given by the magnitudes of the Shapley values and at the direction in which each feature pulls the model given by the signs of the Shapley values. I will discuss more on this topic in Chapter \ref{chap-difficulty}.

%It is also important to note that the degree of easiness to interpret may depend on each individual. For example, a patient, a doctor, and an artificial intelligence researcher may have different opinions on the easiness to interpret of a given explanation.

%At the opposite spectrum, natural language explanations are the easiest for humans to interpret since they are already in human language. 

\paragraph{Faithfulness.}\label{faithful}
Faithfulness of an explanation refers to the accuracy with which the explanation describes the decision-making process of the target model. Faithfulness of an explanation should not be confused with the property of an explanation to provide ground-truth argumentation for solving the task at hand, which is independent of a model decision-making process. I will call this latter property correctness (more details on this in Chapter \ref{chap-esnli}). For example, a natural language explanation is faithful if the target model internally made use only of the argumentation provided by the explanation (encoded in a model-intelligible way) to reach its prediction.

Explanations directly influence the perception and trust of users in target models. Hence unfaithful explanations may be dangerous because can either encourage users to trust unreliable and potential hazardous models or discourage users from trusting perfectly reliable models. 

%As a particular aspect of faithfulness, one can consider the non-ambiguity (or completeness) with which an explanation describes the decision-making process of a model. If the explanation leaves room to 

As mentioned before, verifying faithfulness of explanations is one of the two main goals of this thesis, and it will be specifically addressed in Chapters \ref{chap-difficulty}, \ref{chap-verify}, and \ref{chap-inconsist}.

%% file: difficulty-explain.tex
\chapter{Difficulties and Subjectivity of Explaining with Feature-Based Explanations}
\label{chap-difficulty}
%This chapter is based on Publication \ref{publication-verify}.

One of the main goals in this thesis is to verify whether explanations are faithfully describing the decision-making processes of the models that they aim to explain. In the first place, my focus is to introduce a framework to verify the faithfulness of the explanations provided by feature-based post-hoc explanatory methods. To do so, the first challenge is to define what means a faithful feature-based explanation. In Section \ref{2gt} of this chapter, I show that for certain models and instances, there exist more than one ground-truth feature-based explanation. Hence, the faithfulness of an explanation (or of an explainer) depends on the type of ground-truth explanation preferred in practice. %I identify two types of ground-truth feature-based explanations, which I call \textit{neighbourhood explanation} and \textit{pointwise explanation}. 

Moreover, I show that two influential classes of explainers, namely, Shapley explainers and minimal sufficient subsets explainers, target different types of ground-truth explanations. I also demonstrate that in certain cases, neither of them is enough to provide a complete view of a decision-making process. Knowing that there can be more than one ground-truth feature-based explanation, as well as knowing to which ground-truth explanation each explainer adheres to, have not, to my knowledge, been previously emphasised in the literature. However, these are critical pieces of information for both users --- to pick the class of explainers that best suits their needs and interpret the explanations correctly --- and researchers --- to tell to users the intended behaviour, strengths, and limitations of the explainers, as well as to provide fair comparisons when automatically evaluating explanatory methods. On the contrary, influential works seem to imply that there is only one ground-truth feature-based explanation that all explainers aim to find.

The second challenging aspect in verifying the faithfulness of explanations is precisely the fact that, in general, we do not know the decision-making process of a neural model. To address this challenge, in Section \ref{sel-pred} of this chapter, I investigate to what extent a particular type of feature-based self-explanatory model that is expected to provide faithful explanations actually does so. This type of model will further be employed as a testbed for verifying feature-based explanatory methods in Chapter \ref{chap-verify}.

The findings in this chapter, which are partially based on \citep{verify}, are not only important as a foundation for the verification framework that I will introduce in Chapter \ref{chap-verify}, but also by themselves because they point to fundamental aspects of explainability. 

\paragraph{Illustrating examples.} Before diving into the core of this chapter, I introduce the reader to the type of illustrating example that I will use extensively hereafter. More precisely, I will use hypothetical models for the task of textual sentiment analysis to exemplify the points made in this chapter. Such models take as input a review, i.e., a piece of text that contains an opinion on an object, and outputs a score that reflects the sentiment of the review towards the object or towards one aspect of the object. This task is a salient task in natural language processing with important real-world applications \citep{beer-annot, movie-rev}. Similarly to \citet{rcnn}, I will treat the task as a regression, with the score being a real number linearly reflecting the intensity of the sentiment. In the examples, -1 will be the most negative score, 1 the most positive score, hence 0 reflects the neutral score. We assume a difference of at least 0.1 to be significant.\footnote{In practice, this can happen, for example, if the scores are continuous transformations of star ratings that accompany the review. If one can choose to give half-star ratings of up to $5$ stars, then a difference of $0.1$ in a sentiment score corresponds to different star ratings.} 
An important type of sentiment analysis task is that of identifying the sentiment towards one aspect in a multi-aspect review, i.e., a review that contains opinions on multiple aspects of the object. For example, the BeerAdvocate dataset \citep{beer-annot} contains human-written reviews that cover four aspects of a beer, namely, appearance, palate, taste, and smell. In this chapter, I will use several examples of hypothetical models for both overall and multi-aspect sentiment analysis as illustrating examples.

\section{Multiple Types of Ground-Truth Feature-Based Explanations}
\label{2gt}

In this section, I show that there are cases of models and instances for which there exist more than one ground-truth feature-based explanation. I also show that two prevalent classes of explainers each promote different ground-truth explanations without explicitly mentioning it. This has not, to my knowledge, been emphasised in the literature so far. On the contrary, current influential works seem to imply that there is only one ground-truth feature-based explanation for a prediction of a model. Nonetheless, it is critical for both users and researchers alike to be aware of these differences.

\begin{figure*}[t]
    \centering
    \centerline{\scalebox{.73}{\begin{tabular}{ll|ll}
        \multicolumn{4}{l}{\hspace*{30ex}$m$: \textsc{if} ``very good'' \textsc{in} \textsc{input}: \textsc{return 0.9;} }\\
         \multicolumn{4}{l}{\hspace*{30.7ex}\hspace{0.4cm} \textsc{else if} ``nice'' \textsc{in input}: \textsc{return 0.7;}} \\
 \multicolumn{4}{l}{\hspace*{30.7ex}\hspace{0.4cm} \textsc{else if} ``good'' \textsc{in input}: \textsc{return 0.6;}}\\
\multicolumn{4}{l}{\hspace*{30.7ex}\hspace{0.4cm} \textsc{else return 0.}} \\  
\multicolumn{4}{l}{}\\
 \multicolumn{2}{l|}{$\textbf{x}^{(1)}$: ``The movie was good, it was actually nice.''} & \multicolumn{2}{l}{$\textbf{x}^{(2)}$: ``The movie was nice, in fact, it was very good.''}\\
 $m(\textbf{x}^{(1)}) = 0.7$ & & $m(\textbf{x}^{(2)})=0.9$ &\\ 
 & & & \\
 \underline{Shapley values} & \underline{Minimal sufficient subsets} &  \underline{Shapley values} & \underline{Minimal sufficient subsets}\\
 ``nice'': 0.4 & \{``nice''\} & ``good'': 0.417 & \{``good'', ``very''\}\\
  ``good'': 0.3 & & ``nice'': 0.367\\ 
 rest of tokens: 0 & & ``very'': 0.116\\
 & & rest of tokens: 0\\ 
    \end{tabular}}}
    \caption[Examples of cases with at least two ground-truth feature-based explanations]{Examples of cases with at least two ground-truth feature-based explanations given by Shapley explainers and minimal sufficient subsets explainers, respectively. The Shapley values are computed from Equation \ref{eq:shapley}. The assumption is that the scores are linearly reflecting the intensity of the sentiment and that a difference of 0.1 is significant.}
    \label{fig:2gt}
\end{figure*}

Hereafter, I will label as a Shapley explainer any explainer that explicitly aims to provide Shapley values as explanations. As we saw in Chapter \ref{chap-taxo}, \citet{shap} argued that all feature-additive explainers should aim for Shapley values as explanations. However, given the findings in this chapter, I want to allow feature-additive explainers to adhere to different views. 

To best illustrate the existence of more than one ground-truth feature-based explanation for the prediction of a model on an instance, in Figure~\ref{fig:2gt}, I present an example of a hypothetical sentiment analysis regression model $m$. 
This model makes its predictions as follows: the mere presence of the substring ``very good'' in the input text leads to a very positive score of 0.9. In the absence of ``very good'', if ``nice'' is present in the input text, then the model provides a positive sentiment of 0.7. If neither ``very good'' nor ``nice'' are present but ``good'' is present in the input text, then the model provides a score of 0.6, and finally, if none of these positive-indicator tokens are present, then the model defaults to the neutral score of 0. This is, therefore, a trivial model for which we know its decision-making process.
Applying this model to the instance $\textbf{x}^{(1)}$: ``The movie was good, it was actually nice.'', the model predicts $m(\textbf{x}^{(1)}) = 0.7$ because ``nice'' is present in the input instance. Hence, one can argue that ``nice'' is the only important feature for this prediction of 0.7, while ``good'' would have been the only important feature for a different prediction (of 0.6). However, one may argue that ``good'' also has to be flagged as important for this prediction for the following reason: if ``nice'' is to be eliminated, then the model would rely on ``good'' to provide a score as high as 0.6 instead of the much lower default of $0$. Therefore, we see that there are two equally valid perspectives on what a ground-truth feature-based explanation should be, even for this trivial model. Therefore, it is subjective which perspective is preferred in practice. % Nonetheless, stating that both ``nice'' and ``good'' are important for the prediction might imply that it was crucial to have both features present together in order to have the same prediction, which is not true for this model. %Thus, we see that even for a simple model for which we know the ground-truth inner workings, trying to map its decision-making process to an explanation formed only by its input features is restrictive and can lead to multiple equally valid ground-truth explanations.

The difficulty faced when trying to explain model $m$ with feature-based explanations is even more pronounced on the instance $\textbf{x}^{(2)}$: ``The movie was nice, in fact, it was very good.''. The model predicts $m(\textbf{x}^{(2)}) = 0.9$ because ``very good'' is a substring of the input instance. Hence, the features ``very'' and ``good'' form one ground-truth explanation for this prediction. However, for this instance, if ``good'' is eliminated, the model relies on ``nice'' (and not on ``very'') to provide a score as high as 0.7, while if both ``good'' and ``nice''are eliminated, then the score drops all the way to 0. Hence, from this perspective, ``nice'' can be seen as more important than ``very'', and the explanation that provides ``good'', ``nice'', and ``very'' in this order of importance is also a ground-truth feature-based explanation. 

\paragraph{Shapley values vs.\ minimal sufficient subsets.}
The two types of ground-truth explanations described above are separately advocated by the two influential classes of Shapley and minimal sufficient subsets explainers. On one hand, Shapley explainers \citep{shap} tell us that ``nice'' is the most important feature for this model on this instance $\textbf{x}^{(1)}$, with a weight of 0.4, but also that ``good'' has a significant contribution of 0.3 (see Equation \ref{eq:shapley}). On the other hand, minimal sufficient subsets explainers find that only the feature ``nice'' is important (see Equations \ref{eq:mss1} and \ref{eq:mss2}). Similarly, for the prediction of model $m$ on instance $\bfx^{(2)}$, the Shapley explanation tells us that ``good'' and ``nice'' are the two most important features with very close importance weights, and that ``very'' is about three times less important than ``nice''. On the other hand, the minimal sufficient subsets perspective tells us that the two most important features are instead ``good'' and ``very'', while ``nice'' would not be mentioned as important at all. 

The difference between the two types of ground-truth explanations stems from the fact that the explanation aligned with the Shapley values aims to provide \textit{average importance weights}\footnote{More precisely, a weighted average of the marginal contributions of the feature inside each possible subset of features of the original instance.} for the features on a \textit{neighbourhood} of the instance, while the explanation aligned with the minimal sufficient subsets aims to provide the \textit{pointwise} features used by the model on the instance \textit{in isolation}. The Shapley values come from cooperative game theory \citep{shapley1}, where they were introduced to promote \textit{fairness} in distributing a total gain of a coalition among its players. This is why the Shapley values take into account the \textit{capabilities} of players to perform in any sub-coalition, via the consistency condition in Equation \ref{eq-consist}. On the other hand, the minimal sufficient subsets perspective rewards only the players that are absolutely necessary inside the full coalition. In the example in Figure \ref{fig:2gt}, for the model $m$, ``nice'' is not necessary in the presence of ``very'' and ``good'' appearing next to each other, even if ``nice'' alone would \textit{win} a significant amount of $0.7$.

I will thereafter refer to these two types of ground-truth explanations as \textit{neighbourhood explanation} and \textit{pointwise explanation}, respectively. These names are meant to reflect the high-level fundamental difference between the two ground-truth explanations, such as considering ``nice'' as more important than ``very'' versus considering ``nice'' as not important at all for explaining the prediction of $m$ on $\bfx^{(2)}$, as opposed to low-level differences, such as the exact values of the Shapley weights or the fact that minimal sufficient subsets perspective provide subsets instead of weights, for which I will explicitly refer to as Shapley explanation (or Shapley explainer) and minimal sufficient subsets explanations (or minimal sufficient subsets explainer) to avoid confusion.

%More precisely, an explanation is considered to be a neighbourhood type of explanation if the importance of each feature in an instance reflects the feature's importance in the neighbourhood of the original instance. It can eventually have different 

\paragraph{Not always distinct.} Even though the two types of ground-truth explanations are distinct for certain models and instances, such as the ones mentioned above, in many cases they can coincide. For example, for the same model $m$ but applied to instances that contain only one of the three key sub-phrases ``very good'', ``nice'', and ``good'', such as ``The movie was good.'', both the neighbourhood explanation and the pointwise explanation point towards the same important features.

\paragraph{Not emphasised in the literature.}
To my knowledge, the fundamental difference between the two types of ground-truth feature-based explanations illustrated above was only briefly alluded by \citet{integrated-grads} in their example of explaining the function $\min(x_1, x_2)$ on the instance $x_1 = 1, x_2 = 3$. Their method, called integrated gradients, attributes the whole importance weight (of $1 = \min(1, 3)$) to the \textit{critical} feature $x_1$ (and hence $0$ importance to $x_2$), while Shapley-Shubik \citep{Shubik} attributes $0.5 = \frac{1}{2}\min(1, 3)$ weight to each feature. They also mention that preferring one explanation over the other is a subjective matter. 

On the other hand, the influential work of \citet{shap} implies, only with a graph of results (their Figure 4), that in their user study on Amazon Mechanical Turk, \textit{all} participants provided as explanation for the function $\max(x_1, x_2, x_3)$\footnote{Devised as a story of three people making money based on the maximum score that any of them achieved.} on the input $x_1 = 5, x_2 = 4, x_3 = 0$ that $x_1$ has an importance weight of $3$, $x_2$ of $2$, and $x_3$ of $0$. The authors do not mention the number or characteristics of the participants in their user study, nor do they provide the precise guidelines that they gave to the participants. More extensive and well-documented user studies are necessary to conclude the usefulness of one ground-truth explanation over the other. %Moreover, they state that the fact that DeepLIFT \citep{deeplift} is attributing the whole importance to the feature that achieves the maximum is an error that needs to be fixed. We also highlight that the integrated gradients method \citep{integrated-grads} is based on feature-additivity (their Proposition 1). While \citet{shap} argues that all feature-additive methods should satisfy the three properties described in \ref{shapley-conds}, and hence they should rely on the Shapley values as importance weights, in this thesis we refer specifically to explainers that adhere to the Shapley values type of explanation as Shapley explainers. %, allowing the feature-additive explainers such as DeepLIFT and integrated gradients to have a differnt 

%Similarly, minimal sufficient subsets methods also argue for their ground-truth explanation to be the  

Another pattern present in current works that may hinder the acknowledgement of both the existence of more than one ground-truth explanation and the fundamental difference between the Shapley and minimal sufficient subsets explainers is the fact the explainers from the different classes are directly compared on their ability to identify ground-truth features used by models for their predictions. For example, both \citet{l2x} and \citet{invase} compare L2X, a minimal sufficient subsets explainer, with Shapley methods \citep{shap} on identifying the ground-truth features used by a model trained on synthetic datasets where the ground-truth features are known, called 2-dimensional XOR, Orange Skin, and Switch Feature. While these particular synthetic datasets happen not to violate either of the two ground-truth explanations described above, this information is not mentioned at the time of comparison. Such comparisons risk inducing the idea that there is always only one ground-truth feature-based explanation that all feature-based explainers should aim to find.

%%Thus, these findings encourage researchers to directly state the intended behaviour of the explanations that their explanatory methods aim to provide (e.g., neighborhood or pointwise explanations) in order for (1) users to be able to pick the type of explanations that best suits them and interpret them accordingly, and (2) researchers to be provide fair automatic verifications and comparisons among different explanatory methods.

\begin{figure*}[t]
    \centering
    \centerline{\scalebox{0.55}{\begin{tabular}{ll|ll|ll}
\multicolumn{6}{l}{\textsc{\hspace{1.3cm}1. aspect\_indicators} = \{``taste'', ``smell'', ``appearance''\} (\textsc{and any variation, such as} ``Tastes'')} \\
\multicolumn{6}{l}{\textsc{\hspace{1.3cm}2. sentiment\_indicators} = \{``amazing'' $\rightarrow$ 1, ``good'' $\rightarrow$ 0.6, ``refreshing'' $\rightarrow$ 0.6, ``bad'' $\rightarrow$ $-$0.6, ``peculiar''$\rightarrow$ $-$0.3, ``horrible'' $\rightarrow$ $-$1\}}\\
\multicolumn{6}{l}{\textsc{\hspace{1.3cm}3. A sentiment indicator is associated to its closest aspect (occluded tokens are counted).}}\\
\multicolumn{6}{l}{\textsc{\hspace{1.3cm}4. An occluded token is considered to be neutral.}}\\
\multicolumn{6}{l}{\textsc{\hspace{1.3cm}5. If more sentiment indicators are associated to an aspect, then}}\\ 
\multicolumn{6}{l}{\hspace{1.7cm}(i) \textsc{if all are of the same sign: the score for that aspect is the score of the strongest sentiment.}}\\ %: $S_{\text{asp}}= max()$.
%\multicolumn{6}{l}{\hspace{2.8cm}\textsc{(i) if all are of the same sign, the score for that aspect is the score of strongest: $S_{\text{asp}}= max()$.}}\\
\multicolumn{6}{l}{\hspace{1.7cm}(ii) \textsc{if there are both positive and negative sentiments associated to the aspect: the score for that aspect is}}\\
\multicolumn{6}{l}{\hspace{2.3cm}\textsc{the thresholded sum of scores (max(min(sum\_scores, 1), $-$1)).}}\\ %\footnote{Such that scores are always between -1 and 1.}
\multicolumn{6}{c}{}\\
\multicolumn{2}{l|}{$m^{\text{O}}$: s = \textsc{sum of scores of aspects}} & \multicolumn{2}{l|}{$m^{\text{S}}$: \textsc{return score of smell}} & \multicolumn{2}{l}{$m^{\text{T}}$: \textsc{return score of taste}} \\
%\multicolumn{2}{l|}{\hspace{1.2cm} \textsc{aspect}} & \multicolumn{2}{l|}{} & \multicolumn{2}{l}{} \\
\multicolumn{2}{l|}{\hspace{0.6cm} \textsc{return max(min}(s, 1), $-$1)} & \multicolumn{2}{l|}{} & \multicolumn{2}{l}{} \\
\multicolumn{2}{l|}{} & \multicolumn{2}{l|}{$\bfx^{\text{S1}}$: ``Tastes horrible, peculiar smell.''} & \multicolumn{2}{l}{$\bfx^{\text{T1}}$: ``Tastes good, refreshing.''}\\
\multicolumn{2}{l|}{$\bfx^{\text{O}}$: ``The beer has an amazing appearance,} & \multicolumn{2}{l|}{$m^{\text{S}}(\bfx^{\text{S1}})=-0.3$} & \multicolumn{2}{l}{$m^{\text{T}}(\bfx^{\text{T1}})=0.6$}\\
\multicolumn{2}{l|}{\hspace{0.8cm}a good smell, a bad taste.''} & \multicolumn{2}{l|}{} & \multicolumn{2}{l}{}\\
\multicolumn{2}{l|}{$m^{\text{O}}(\bfx^{\text{O}})=1$} & \underline{Shapley explanation}               & \underline{MSS explanation} & \underline{Shapley explanation}               & \underline{MSSs explanations}\\
\multicolumn{2}{l|}{} & 1. ''smell``: $-$0.29 & \{``peculiar'', ``smell''\} & 1. ``Tastes'': 0.4  & \{``Tastes'', ``good''\},\\
\underline{Shapley explanation}               & \underline{MSS explanation}      &      2. ''Tastes``: 0.26          &         &        2. ``good'': 0.1,   & \{``Tastes'', ``refreshing''\}  \\
1. ``amazing'': 0.52 & \{``amazing'', ``appearance''\} & 3. ''horrible``: $-$0.14 &  & \hspace{0.43cm}``refreshing'': 0.1  &   \\
2. ``good'': 0.40 &  & 4. ''peculiar``: $-$0.13  & &  &\\
3. ``bad'': $-$0.23 &  &  &  &  & \\
4. ``smell'': 0.15 & &&  & & \\
5. ``appearance'': 0.12 &  & \multicolumn{2}{l|}{$\bfx^{\text{S2}}$: ``Tastes amazing, peculiar smell.''}  &  \multicolumn{2}{l}{$\bfx^{\text{T2}}$: ``Tastes amazing. The smell is also amazing.''} \\
6. ``taste'': 0.03 &  &  $m^{\text{S}}(\bfx^{\text{S2}})=-0.3$ & &  $m^{\text{T}}(\bfx^{\text{T2}})=0.6$ & \\
\multicolumn{2}{l|}{} & \multicolumn{2}{l|}{}  & \multicolumn{2}{l}{} \\
\multicolumn{2}{l|}{} &  \underline{Shapley explanation}               & \underline{MSS explanation}  &\underline{Shapley explanation}               & \underline{MSSs explanations} \\
&  & 1. ``peculiar'': $-$0.27 & \{``peculiar'', ``smell''\} &  1. ``Tastes'': 0.58  & \{``Tastes'', ``$\text{amazing}^1$''\}, \\
& & 2. ``smell'': $-$0.10 &  & 2. ``$\text{amazing}^1$'': 0.42   & \{``Tastes'', ``$\text{amazing}^2$''\}     \\
&  & 3. ``amazing'': 0.05 & &  3. ``$\text{amazing}^2$'': 0.08  &  \\
& & 4. ``Tastes'': 0.02 & & 4. ``smell'': $-$0.08 & \\
\end{tabular}}}
\caption[Examples illustrating the strengths and limitations of Shapley and minimal sufficient subsets explanations]{Examples illustrating the strengths and limitations of the two types of feature-based explanations, as presented in Section~\ref{sec:s-lim}. The five rules are common to all three models. The Shapley values were computed via Equation~\ref{eq:shapley}, and written in decreasing order of their importance (absolute value); the non-mentioned features received 0 weight. MSS stands for minimal sufficient subset. In the last example, the superscript of ``amazing'' differentiates between its two occurrences.}
\label{fig:stren_lim}
\end{figure*}

\subsection{Strengths and Limitations}\label{sec:s-lim}
This section reveals strengths and limitations of the two types of feature-based explanations presented above.

\paragraph{Redundant features.}
By looking only at the Shapley explanation for $m(\bfx^{1})$ in Figure~\ref{fig:2gt}, one cannot know whether (1)~the model requires \textit{both} features ``nice'' and ``good'' to make its prediction of $0.7$ (which is not the case for $m$), or (2)~one of these features is redundant in the presence of the other (which is the case for $m$). In contrast, minimal sufficient subset explanations do not contain redundant features (as they would violate Equation~\ref{eq:mss2}), and hence, the minimal sufficient subset explanation for $m(\bfx^{1})$ is able to distinguish between the two scenarios.

\paragraph{Feature cancellations: genuine vs.\ artefacts.}%\label{cancel}
In certain cases, there exist features that cancel each other out. 
Consider the model $m^{\text{O}}$ in Figure~\ref{fig:stren_lim}, which predicts the overall sentiment on a beer from a multi-aspect review by adding up the scores that it associates to each aspect in the review. 
On the instance $\bfx^{\text{O}}$, $m^{\text{O}}$ predicts $1$ by taking into account all three aspects. However, the minimal sufficient subset explanation is \{``amazing'', ``appearance''\}%, due to Equation~\ref{eq:mss2}. % 
---it does not contain the features ``bad'', ``taste'', ``good'', and ``smell'', due to Equation~\ref{eq:mss2}. 
%We refer to such a cancellation as a \textit{genuine cancellation}. %
Arguably, users may want to see the features from such a \textit{genuine cancellation} in the explanation. %, i.e., ``bad'', ``taste'', ``good'', and ``smell''. 
Note that these features are flagged as important by the Shapley explanation, which, nonetheless, does not clearly indicate the perfect cancellation between ``good smell'' and ``bad taste''. Moreover, the Shapley explanation gives the impression that ``smell'' and ``appearance'' are much more important (0.15 and 0.12) than ``taste'' (0.03), when, by design, $m^{\text{O}}$ equally takes into account all aspects. Hence, neither the Shapley nor the minimal sufficient subset explanation is well reflecting the decision-making process for $m^{\text{O}}(\bfx^{\text{O}})$.

Artefacts may occur when eliminating features from an instance, distorting the importance of certain features. Model $m^{\text{S}}$ in Figure~\ref{fig:stren_lim} illustrates such an example.
When $m^{\text{S}}$ is applied to $\bfx^{\text{S1}}$, it predicts $-$0.3, and the minimal sufficient subset explanation is \{``peculiar'', ``smell''\}, which, arguably, best reflects the decision-making process for $m^{\text{S}}(\bfx^{\text{S1}})$.  
However, in the Shapley explanation ,``Tastes'' appears to be twice more important than ``peculiar'', and ``horrible'' appears as important as ``peculiar'', even though ``peculiar'' is the actual sentiment indicator for smell. 
Furthermore, note how the Shapley importance weights dramatically change when only the sentiment on taste is changed in instance $\bfx^{\text{S2}}$, even though $m^{\text{S}}$ does not rely on the sentiment on taste to predict the sentiment on smell.  %Hence, the Shapley explanations for these predictions are clearly confusing.
%We call such a scenario an \textit{artefact-caused cancellation}.

\paragraph{Multiple minimal sufficient subsets: genuine vs.\ artefacts.}
In certain cases, there can exist multiple minimal sufficient subsets explanations for one prediction.
For example, for the model $m^{\text{T}}$ and instance $\bfx^{\text{T1}}$ in Figure~\ref{fig:stren_lim}, either of the features ``good'' and ``refreshing'' leads to the score of 0.6. Ideally, minimal sufficient subsets explainers provide all the genuine minimal sufficient subsets, e.g., both \{``Tastes'', ``good''\} and \{``Tastes'', ``refreshing''\}. However, many minimal sufficient subsets explainers are designed to retrieve only one minimal sufficient subset \citep{l2x, invase}. An exception is the SIS explainer \citep{what-made-you-do-this}, which retrieves a set of disjoint minimal sufficient subsets, which might also not be exhaustive (e.g., SIS would not retrieve the second minimal sufficient subset explanation for $m^{\text{T}}(\bfx^{\text{T1}})$, because ``Tastes'' is already taken by the first minimal sufficient subset).
On the other hand, the Shapley explanation gives the same importance to both ``good'' and ``refreshing''. However, with this explanation alone, one would not be able to know whether both ``good'' and ``refreshing'' are necessary to be present or if each individually suffices for the prediction of $0.6$.

Artefacts occurring when eliminating features can also lead certain subsets of features to appear as minimal sufficient subsets. 
For example, for $m^{\text{T}}$ and $\bfx^{\text{T2}}$ in Figure~\ref{fig:stren_lim}, either of the two occurrences of ``amazing'' forms an minimal sufficient subset, but the second one is not reflecting the decision-making process of the model. The Shapley explanation makes the distinction in this case.

\bigskip
In this section, we saw that for certain cases, there can be more than one ground-truth feature-based explanation for a prediction of a model. I identified two such types of explanations and called them neighbourhood and pointwise explanations. We also saw that the Shapley explainers aim to provide neighbourhood explanations, while the minimal sufficient subsets explainers aim to provide pointwise explanations. Both types of explanations and explainers give valuable insights into the decision-making process of a target model, and have their strengths and limitations. Hence, one type of explanation may be preferred over the other in different real-world use-cases, and the choice of ground-truth is therefore best left in the hands of the users. Therefore, users need to be informed of the strengths, limitations, and differences among various types of explanations and explainers, in order to pick and use them accordingly.

Finally, using both neighbourhood and pointwise explanations together may bring a complete picture of the decision-making process of a model. %, as it would be the case for our example of model $m$ in Figure \ref{fig:2gt} on instance $\bfx^{(2)}$. 
The investigation on combining the two types of explanations can therefore be an interesting future work. Additionally, future work may include a comprehensive user study to decide the extent to which users could benefit from each of these types of explanations, as well as from their combination. 

Most importantly, the work described in this section encourages researchers to pay particular attention to the assumptions behind the explanations that their explanatory methods aim to provide, and to state them directly in their works. For example, a good practice could be that each work introducing an explainer or a verification framework for explainers has a section \textit{Specifications} where the authors present the intended behaviour, strengths, and limitations of their method(s).

\section{Trusting Selector-Predictor Models}
\label{sel-pred}
A potential way of verifying the faithfulness of post-hoc explainers is to use feature-based self-explanatory models as testbed, since these models are supposed to provide the features that are relevant for their predictions. In this section, I investigate a particular type of feature-based self-explanatory model, which I call a \textit{selector-predictor} model. A selector-predictor model is composed of two sequential modules: a \textit{selector} followed by a \textit{predictor}. The selector makes a \textit{hard} selection of a subset of the input features, and \textit{only} the selected features are passed along to the predictor, which outputs the final answer. There is no restriction on the types of neural networks used for modelling the selector and the predictor. Crucially, there is also no need for supervision on the selected features, as long as the selector and the predictor are trained jointly with supervision only on the final answer. Therefore, such a feature-based self-explanatory model is very appealing. Models following this high-level architecture have been introduced, for example, by \citet{rcnn}, \citet{invase}, and \citet{tommi-neurips-rcnn}. 

%The feature-based explanations provided by a selector-predictor model are therefore presumably formed by the selected features. 
I investigate the type of selector-predictor self-explanatory model, since it seemed to conveniently provide, for any instance, the set of relevant features (the selected ones) and irrelevant features (presumably the non-selected ones). Therefore, it appeared that one could test if explainers correctly identify the irrelevant features as less important than the relevant ones. However, we will see below that this type of self-explanatory model may not always faithfully explain itself.
%the suitability of this type of model for being the target model on which we can test the faithfulness of explainers in describing the inner workings 

Formally, let $\D$ be a dataset for which any instance $\bfx$ has a potentially variable number $n$ of features $\bfx = (x_1, x_2, \ldots, x_n)$. For example, $\bfx$ can be a sentence, and the feature $x_i$ the $i$-th token in the sentence.
The selector, which we call $sel(\cdot)$, takes the input $\bfx$ and returns a subset of features $\sel \subseteq \textbf{x}$. The predictor, which we call $pred(\cdot)$, takes as input \textit{only} the features $\sel := sel(\bfx)$ and makes its prediction based exclusively on these features. It thus holds that
$$
m(\textbf{x}) = pred(sel(\textbf{x})) = pred(\sel).
$$
We also call $\nsel$ the non-selected features, i.e., $\nsel = \bfx \setminus \sel$.

For example, for a model tackling a sentiment analysis task, the selector may learn to select only the token ``amazing'' if this token is in the input text, and the predictor may learn that when it receives the input $\sel$ = \{``amazing''\}, it has to predict the most positive sentiment score.

\paragraph{Selector-predictor models aim for minimal sufficient subsets explanations.}
To encourage the selector to do a meaningful job and not simply return all features (unless necessary), one can, for example, use a regularizer that penalises the model proportionally to the number of selected features, as was done in the work of \citet{rcnn}. Thus, such a selector-predictor model aims for a minimal sufficient subset explanation, as defined by Equations \ref{eq:mss1} and \ref{eq:mss2}. For example, a selector-predictor model that learns to mimic the decision-making process of the model $m$ from Figure \ref{fig:2gt}, would likely not select the feature ``nice'' if ``very good'' is a substring of the input. This is because selecting ``nice'' in addition to ``very'' and ``good'' would not change the prediction, but it would increase the penalisation on the number of selected features. 

\paragraph{Minimality is not guaranteed.}
While a regularizer can incentives the minimality on the number of selected features, it is not guaranteed that a selector-predictor model would never select features that are not strictly necessary for the prediction, as this depends on how well the model learns to do so during training time. Moreover, it is an open question whether checking if the set of selected features is minimal %finding a minimal set of features in an instance 
can be done with less than $O(2^{|\sel|})$ queries to the model, since any subset of the selected features may be a minimal sufficient subset.

From the perspective of a self-explanatory model, not complying with the minimality condition may sometimes be crucial. For example, if the model learned to rely on a spurious correlation, such as learning that the word ``bottle'' alone implies a positive sentiment, and if the selected features include ground-truth sentiment-indicator words, then the explanation formed by the selected features would hide from the users the spurious correlation learned by the model. On the other hand, as mentioned in Section \ref{2gt}, minimality implies considering features that cancel each other out as irrelevant, and this may not always be desired. Hence, the fact that a selector-predictor model may not always fulfil the minimality condition might sometimes be useful in practice. 

From a verification perspective, if certain selected features are irrelevant for the prediction, then it is a mistake to penalise an explainer for not presenting those features as more important than other equally irrelevant features. 

%Showing to the user features that were not essential for the prediction as important may not always be a problem. In fact, for certain cases, it may even be desired. %to show to the users certain features that were not absolutely necessary for the prediction. 
%For example, \citet{rcnn} tackled the task of sentiment analysis and explicitly added a second regularizer that penalizes their selector-predictor model for selecting disconnected tokens in order to incentivize the model to select contiguous sub-phrases. This regularizer likely comes in conflict with the regularizer used for selecting as few tokens as possible, yet this a design choice. However

\begin{figure}[h]
\centerline{\scalebox{.97}{\begin{tabular}{l}
%$m^{\text{ns}}$:\\
\textsc{if} ``very good'' \textsc{in} \textsc{input}:
  \textsc{select} \{``very''\}  \& \textsc{return 1}; \\
\textsc{else if} ``not good'' \textsc{in} \textsc{input}: \textsc{select} \{``not''\} \& \textsc{return -0.8}; \\
\textsc{else if} ``good'' \textsc{in} \textsc{input}: \textsc{select} \{``good''\} \& \textsc{return} 0.8; \\
\textsc{else select} $\emptyset$ \& \textsc{return 0}.
\end{tabular}}}
\caption[Example of a selector-predictor model that does not always select sufficient subsets of features]{Example of a selector-predictor model that does not always select sufficient subsets of features. For instances containing ``very good'' or ``not good'', ``good'' is a crucial feature for the prediction, yet the model does not select it. However, the model would not present this undesired behaviour for any instance that does not contain ``very good'' or ``not good''.}
\label{fig:handshake}
\end{figure}

\paragraph{Sufficiency is not guaranteed.}
Another concerning issue of selector-predictor models is the fact that selected features are not guaranteed to form a sufficient subset, i.e., the model can rely on features it does not select. More precisely, $m$ may learn an internal emergent communication protocol \citep{communication} between its selector and predictor such that non-selected features are crucially influencing the prediction via a hidden encoding that the selector and predictor agree upon during training. An example of a model that does not select sufficient subsets is given in Figure~\ref{fig:handshake}. For instances containing the substrings ``very good'' or ``not good'', such as $\textbf{x}^{(1)}$: ``The movie is very good.'' and $\textbf{x}^{(2)}$: ``The movie is not good.'', the feature ``good'' is influencing the prediction, but the model does not select it. However, the same model is not exhibiting the undesired behaviour for any instance that does not contain ``very good'' or ``not good'' as substrings. In practice, spurious correlations in datasets may lead to even more misleading selections of features done by a selector-predictor model. Unlike checking the minimality condition, fortunately, it is far less time-consuming to check whether a selector-predictor model $m$ selects a sufficient subset on an instance $\bfx$. By definition, we have that
\begin{equation}
\sel \text{ is a sufficient subset} \iff m(\sel) = m(\bfx).   
\end{equation}
%$m(\sel) \neq m(\bfx) \iff m$ exhibits a handshake on $\bfx$.
Hence, for using a selector-predictor model either as a self-explanatory model or as a testbed for verifying explainers, it is crucial to check the sufficiency condition beforehand.

\paragraph{Inconsistency in selecting features that cancel each other out.} 
We saw that a selector-predictor is not guaranteed to select a minimal set of sufficient features. Hence, on each instance, such a model may select certain groups of features that cancel each other out, while it may not select others. This inconsistency can have advantages and disadvantages. On one hand, such a model has a chance to select the features that form genuine cancellations and not select the features that form artefact-caused cancellations. Neither minimal sufficient subsets explainers nor Shapley explainers are capable of doing so. On the other hand, the opposite may also happen, that is, a selector-predictor model may select features that form artefact-caused cancellations and not select features that form genuine cancellations. Hence, users need to be aware of this risk. It is an open question whether one could check in less than an exponential number of queries to the model whether there are non-selected features that cancel each other out.

\section{Conclusions and Open Questions}
In this chapter, I unveiled certain difficulties of explaining models with feature-based explanations. First, I showed that there can be more than one ground-truth feature-based explanation for the prediction of a model on an instance. I also showed that two prevalent classes of post-hoc feature-based explanatory methods aim to provide different ground-truth explanations, and I uncovered some of their strengths and limitations. 

Second, I investigated a type of feature-based self-explanatory model, called selector-predictor, and showed its strengths and limitations in providing its own explanations.

The findings in this chapter encourage researchers to directly state the specifications of their explainers and verification frameworks. They also pave the way to further investigations, such as: (1) which type of feature-based ground-truth explanations or which combination of them best suit users in different circumstances, and (2) whether selector-predictor models can be regularised during training to enforce the sufficiency condition on the selected features.

%% file: verify.tex
\chapter{Verifying Feature-Based Post-Hoc Explanatory Methods}
\label{chap-verify}
%This chapter is based on Publication \ref{publication-verify}.

In this chapter, which is based on \citep{verify}, I introduce a framework for verifying the faithfulness with which  %instance-wise 
feature-based post-hoc explanatory methods describe the decision-making processes of the models that they aim to explain. 

% More precisely, \vf{} focuses on finding false positives, i.e., if an explainer ranks less important features higher than features that are more important for the predictions of the target model. The framework has the novelty of sanity checking explainers on non-trivial neural networks trained on real-world datasets and for which we know parts of their decision-making process, as explained in the previous chapter.

% \vf{} is generic and can be instantiated on various tasks and domains. I instantiate it on the task of sentiment detection and provide three sanity tests, which can be used off-the-shelf for testisting exiting and future explantory methods.\footnote{The tests are available at \url{https://github.com/OanaMariaCamburu/CanITrustTheExplainer}.}
% Moreover, I present the performance of three explanatory methods on these tests. Finally, I discuss ways in which the current limitations of this verification framework can further be addressed and open the path towards more robust and flexible verification frameworks that can be adapted to users needs.

\section{Motivation}\label{sec:intro}
As we saw in Chapter \ref{chap-taxo}, a large number of %instance-wise
feature-based post-hoc explanatory methods have recently been developed with the goal of shedding light on the decision-making processes learned by neural models~\citep{lime, shap, lrp, deeplift, integrated-grads, saliency, anchors, l2x, lstree}. While current explainers manage to point out catastrophic biases of certain models, such as relying on the background of an image to discriminate between \textit{Wolf} and \textit{Husky}~\citep{lime}, it is an open question how to verify if these methods are faithfully describing the decision-making process of a model that has a less obvious bias. This is a difficult task precisely because, generally, the decision-making process of deep neural networks is not known. Consequently, the automatic verification of explanatory methods usually relies either on oversimplistic scenarios (an interpretable yet trivial target model such as a linear regression model, or a non-trivial target model but trained on synthetic datasets), or %when the target model is a complex neural network trained on a real-world dataset, a prevalent way to verify the explainers is to assume that the target models behave reasonably, i.e., that they do not rely on irrelevant correlations. 
on the assumption that accurate target models must behave reasonably, i.e., that they do not rely on spurious correlations in the datasets. For example, in their verification test based on morphosyntactic agreement, \citet{nina} assume that a model which predicts whether a verb should be singular or plural given the tokens before the verb must be doing so by focusing on a noun that the model must have identified as the subject. Such assumptions may be poor since recent works show that neural networks can learn to heavily rely on surprising spurious correlations even when they provide correct answers \citep{artifacts, rightforwrong, breaking}. Therefore, it is not reliable to verify the faithfulness of an explainer only based on whether the explanations appear coherent for the task at hand. %Assessing the faithfulness of an explainer to a complex, real-world target model is a challenging task, precisely because the ground-truth decision-making process of such models is not known.

In order to overcome these issues, I propose to use the type of selector-predictor model introduced in Chapter \ref{chap-difficulty}, since it has the potential to provide a set of relevant and irrelevant features for each of its predictions. Also, this type of model is a non-trivial neural model that can be trained on real-world datasets. 
%%%In this chapter, I introduce \vf, an automatic verification framework that employs selector-predictor models as target models. This framework can verify explainers on non-trivial neural networks trained on real-world datasets and for which we know part of their decision-making processes. 
%More precisely, similarly to the BAM framework \citep{bam}, \vf{} tests for false positives, i.e., it tests if explainers rank less relevant features higher than more relevant features. 
However, as we saw in the previous chapter, selector-predictor models can have certain limitations that one needs to take into account when using them as testbeds for verifying explainers. Therefore, I introduce checks to account for some of these limitations, as well as further improvements that can be done to further alleviate the remaining limitations.

I instantiate the \vf{} framework on the selector-predictor model introduced by \citet{rcnn} and trained on the BeerAdvocate dataset \citep{beer-annot} for the task of multi-aspect sentiment analysis. I thereby produce three sanity tests, one for each aspect in BeerAdvocate. Further, I test three popular explainers, namely, LIME~\citep{lime}, KernalSHAP~\citep{shap}, and L2X~\citep{l2x}, and provide their performances on each of these three sanity tests to raise awareness of the potential unfaithful explanations that these explainers can produce. %The fact that pointwise explainers do not pass our relatively simple tests is concerning, as we are testing for arguably easy pointwise-errors. %Even more surprising, the two neighborhood explainers we tested, LIME \citep{lime} and KernalSHAP \citep{shap}, performed better on our tests than L2X \citep{l2x}, a pointwise explainer. 

The three sanity tests can be used off-the-shelf\footnote{The tests are available at \url{https://github.com/OanaMariaCamburu/CanITrustTheExplainer}.} for testing existing and future feature-based post-hoc explanatory methods. Moreover, \vf{} is generic and can be instantiated on other tasks, domains, and neural architectures for the selector and the predictor networks, thus allowing the generation of a large number of sanity tests.

\section{Related Work}\label{sec:rel_work}
Despite the increasing number of proposed explanatory methods, it is still an open question how to thoroughly validate their faithfulness to the target models that they aim to explain. There are four types of verification commonly performed: 
\begin{enumerate}
    \item \textbf{Interpretable yet simple target models.} Typically, explainers are tested on linear regression and decision trees (e.g., in \citet{lime}) or support vector representations (e.g., in \citet{maple}). While this evaluation accurately assesses the faithfulness of the explainer to the target model, these very simple models may not be representative of the large and intricate neural networks used in practice.
    
    \item \textbf{Synthetic setups.} Another popular setup is to create synthetic tasks where the set of important features is controlled \citep{l2x, invase, shap}. For example, \citet{l2x} performed evaluations on four synthetic tasks. %: 2-dim XOR, Orange Skin, nonlinear additive model, and switch feature. 
    While there is no limit on the complexity of the target models trained in these setups, their synthetic nature may still prompt the target models to learn simpler functions than the ones needed for real-world applications. This, in turn, may ease the job for the explainers.
    
    \item \textbf{Assuming a reasonable behaviour of the target model.} In this setup, one identifies certain intuitive heuristics that a high-performing target model is assumed to follow. For example, \citet{l2x} assume that a neural network that performs well on sentiment analysis must rely on tokens that convey the predicted sentiment. Similarly, \citet{nina} assumes that a neural network that performs well on deciding whether a verb is singular or plural must rely on a noun of the same number that the model must have identified as the subject of the verb. Thus, under this evaluation procedure, an explainer is assessed based on whether it points out the features that are in ground-truth correlation with the prediction of the model. However, neural networks may rely on surprising spurious correlations even when they obtain a high test accuracy~\citep{artifacts, lime, rightforwrong, breaking}. Hence, this type of verification is not reliable for assessing the faithfulness of the explainers to the target model.
    
    \item \textbf{Improved human understanding of a model.} A non-automatic way to evaluate explainers is to check how their explanations help humans predict the behaviour of target models. In this evaluation, humans are presented with a series of a predictions of a model and explanations from different explainers, and are asked to infer the predictions that the model would make on a separate set of examples. One concludes that an explainer $e_1$ is better than an explainer $e_2$ if humans are better at predicting the output of the model after seeing explanations from $e_1$ than after seeing explanations from $e_2$~\citep{anchors}. While this type of evaluation is arguably the most effective since it simulates the real-world usage of explanatory methods, it is nonetheless expensive and requires considerable human effort if it is to be applied to complex real-world neural network models. %Moreover, if $e_1$ is not faithful 
    Hence, it is desirable to have complementary automatic sanity tests that can be used before human evaluations. 
    
\end{enumerate}

In contrast to the above, \vf{} generates automatic sanity tests in which the target model is a non-trivial neural network trained on real-world datasets and for which we know part of its decision-making process.

\vf{} is similar in goal to the framework introduced by \citet{sanity}. However, their framework tests for the basic requirement that an explainer should provide different explanations for a target model trained on the real data than for the same target model trained on randomised data, or for target models that were not trained at all. \vf{} is more challenging and requires a stronger fidelity of the explainer to the target model. Similarly to the concurrent work of \citet{bam}, \vf{} checks for false positives, i.e., whether the explainers are ranking less relevant features higher than more relevant ones. However, their framework uses semi-synthetic datasets, since they paste objects randomly over scene images, while \vf{} can be applied to any real-world dataset.

\section{Verification Framework}
As in the previous chapter, let $\D$ be a dataset for which any instance $\bfx$ has a potentially variable number of features $\bfx = (x_1, x_2, \ldots, x_n)$. Let $m(\textbf{x}) = \text{pred}(\text{sel}(\textbf{x}))$ be a selector-predictor model, and denote by $\sel := \text{sel}(\bfx)$ the selected features and by $\nsel := \bfx \setminus \sel$ the non-selected features. The goal is to find a set of instances $\G \subseteq \D$ such that, for each instance $\bfx \in \G$, we know a subset of features that are relevant to the prediction of $m$ on $\bfx$ and a subset of features that are irrelevant to this prediction. With a dataset $\G$ with such guarantees on the relevance of the features in each instance, one can verify whether an explainer is wrongly attributing more importance to irrelevant features rather than to relevant ones.

In the previous chapter, I showed that selector-predictor models aim to provide a minimal sufficient subset as an explanation, via the selected features. Hence, such models aim for the pointwise type of ground-truth explanations. Consequently, this verification framework is tailored for this kind of ground-truth explanations. 
However, we also saw that neither the minimality nor the sufficiency of the selected features is guaranteed. Therefore, it is possible that the sanity tests produced by \vf{} do not penalise neighbourhood explanations if redundant features are selected (such as ``nice'' in the instance $\bfx^{(2)}$ in Figure \ref{2gt}).

Below I show how one can use such a model as a testbed for explainers, together with the strengths and limitations that come along.

\paragraph{Finding a set of relevant features.}
It is an open question how to check whether the subset of selected features is minimal without an exponential number of queries to the model (on each subset of the selected features). Not being able to guarantee minimality may bring the following issues:
\begin{enumerate}[({I}1)]
    \item\label{i1} certain selected features may be irrelevant to the prediction,
    \item\label{i2} certain groups of features that cancel each other out may be selected while others may not be selected.
\end{enumerate}

Issue (I\ref{i1}) can lead to wrongly penalising an explainer for not placing an irrelevant feature higher than another equally irrelevant feature. To avoid this, I identify the subset of selected features whose individual absence significantly changes the prediction. More precisely, I identify the subset $\scr \subseteq \sel$ as
\begin{equation}\label{eq:clearly-rel}
    \scr = \{ x_i \in \sel : \text{ abs}({\text{pred}}(\sel \setminus \{x_i\}) - m(\bfx) ) \ge \tau \},
\end{equation}
where $\tau$ is the significance threshold, and $\text{abs}(\cdot)$ is the absolute value function. The name $\scr$ stands for Selected Independently Relevant features, emphasising the fact that the relevance of these features is determined by whether their absence alone influences the prediction. 
It is important to note that simply because a selected feature alone did not make a significant change in prediction does not mean that this feature is not relevant, as it may be essential in combination with other features. Hence, $\scr$ is not to be regarded as the only relevant set of features. Yet, considering only this subset of relevant features was enough to obtain conclusive results, as we will see in Section \ref{sec:exp_analysis}.

On the other hand, in Chapter \ref{chap-difficulty}, we saw that the elimination of a feature can influence the prediction as part of an artefact-caused cancellation. Hence, this can be the case for features in $\scr$.
Moreover, given the issue (I\ref{i2}), one may end up with cases for which the explainers are (arguably wrongly) penalised for ranking features from genuine cancellations as more important than features from artefact-caused cancellations, which happens in case the latter are not selected. This is an important limitation of the proposed verification framework. Hopefully, future work will provide a solution to this corner case. In Section \ref{sec:improvements}, I provide a way to decrease the probability of this corner case happening. 

For any instance $\bfx$, the check in Equation \ref{eq:clearly-rel} takes linear time in the number of selected features $|\sel|$ (which is usually significantly less than the total number of features $|\bfx|$). However, this check needs to be done only once in order to obtain an off-the-shelf sanity test that can be applied to any number of explainers.

\paragraph{Sufficiency of the selected features.}
As we saw in Chapter \ref{chap-difficulty}, the sufficiency of the selected features can be checked relatively easily by probing whether $m(\sel) = m(\bfx)$. To avoid any noise that can appear in practice from checking this equality, for this verification framework, I check the stricter condition of
\begin{equation}
\label{ssx}
\text{\textbf{S}}^m_{\sel} = \sel,
\end{equation}
which implies $m(\sel) = m(\bfx)$.

\underline{Proof}: If $\text{\textbf{S}}^m_{\sel} = \sel$ then, by applying the $\text{pred}()$ module we have that $\text{pred}(\text{\textbf{S}}^m_{\sel}) = \text{pred}(\sel)$.

Since by construction $m(\textbf{x}) = \text{pred}(\sel)$, the above equality translates to $m(\sel) = m(\bfx).$  $\square$

\medskip
More instances from $\D$ may be preserved if one instead checks directly for $m(\sel) = m(\bfx)$ with an appropriate tolerance threshold for the equality. It is an open question whether (and why) a selector-predictor model would change its selection of features yet provide the same final prediction.

\paragraph{Final dataset $\G$.}
To obtain the desired dataset $\G$, we first prune $\D$ to a dataset where for each instance, the selected features $\sel$ form a sufficient subset via Equation \ref{ssx}. From this pruned dataset, we further remove the instances for which $\scr = \emptyset$. Hence, we obtain the dataset $\G$ such that for each instance $\bfx \in \G$ we have (1)~the selected features are a sufficient subset, and (2)~there is a subset of these features $\scr \subseteq \sel$ that are independently relevant features. 

%% PROBLEM
\vf{} tests if an explainer ranks irrelevant features higher than relevant ones, under two important assumptions: (i)~that any feature that is not part of a sufficient subset is considered irrelevant (with the warning that groups of feature that cancel each other out may not be part of the sufficient subset), and (ii)~that any selected feature that is independently relevant is considered relevant (with the warning that features from artefact-caused cancellations may be considered relevant).

\paragraph{Metrics.}
Let $f^{e,m}_1(\bfx), f^{e,m}_2(\bfx), \ldots , f^{e,m}_{|\bfx|}(\bfx)$ be the features in $\bfx$ sorted in decreasing order of importance as returned by the explainer $e$. For example, feature $f^{e,m}_1(\bfx)$ is the feature considered by the explainer $e$ as the most relevant for the prediction of $m$ on $\bfx$. %Given the two guarantees (G\ref{g1}) and (G\ref{g2}) for the model $m$ on dataset $\G$, we test if explainers rank all the features in $\text{\textbf{N}}^m_x$ lower than any feature in $\scr$. 
To quantify the extent to which an explainer ranks irrelevant features higher than relevant ones, one can use the following error metrics:\footnote{The limitations of these errors will be discussed in Section \ref{specs}.}

\begin{enumerate}[({M}1)]
    \item \label{m1} \textbf{Percentage of instances for which the most relevant feature provided by the explainer is among the irrelevant features:}
    \begin{equation}
    \text{$\text{PCT}_{\text{first}}$}= \frac{1}{|\G|}\sum\limits_{\bfx \in \G} \mathbb{1}_{\{f^{e,m}_1(\bfx) \in \nsel\}}\,,
    \end{equation}
    where $\mathbb{1}$ is the indicator function.
    
    \item \label{m2} \textbf{Percentage of instances for which at least one irrelevant feature is ranked higher than an independently relevant feature:} 
    \begin{equation}
    \text{$\text{PCT}_{\text{misrnk}}$} = \frac{1}{|\G|}
    \sum\limits_{\bfx \in \G}
    \mathbb{1}_{ \{ \exists i < j \mid f^{e,m}_i(\bfx) \in \nsel \text{ and } f^{e,m}_j(\bfx) \in \scr \}\,. }
    \end{equation}
    
    \item \label{m3} \textbf{Average number of irrelevant features ranked higher than any independently relevant feature:}
    \begin{equation}
    \text{$\text{AVG}_{\text{misrnk}}$} = \frac{1}{|\G|}\sum\limits_{\bfx \in \G} \sum\limits_{i<r'} \mathbb{1}_{ \{ f^{e,m}_i(\bfx) \in \nsel \} \,,}
    \end{equation}
    where $r' = \text{argmax}_j \{f^{e,m}_j(\bfx) \in \scr \}$ is the lowest rank of an independently relevant feature. 

\end{enumerate}

Metric $\text{PCT}_{\text{first}}$ shows the percentage of instances for which the explainer tells us that the most relevant feature is one that was irrelevant for the prediction. Metric $\text{PCT}_{\text{misrnk}}$ shows the percentage of instances for which there is at least one error in the ranking. Finally, metric $\text{AVG}_{\text{misrnk}}$ gives the average number of irrelevant features per instance that are ranked higher than any independently relevant feature.

%\paragraph{Create your own sanity tests.}
%Consider a generic task and a dataset $\D$. A new sanity test can be created by performing the following steps:
\vf{} can be summarised in the following steps, which one can use to create new sanity tests based on other tasks, datasets, or architectures for the selector and predictor modules.
\begin{enumerate}
    \item Choose a task and a dataset $\D$.
    \item Choose a selector-predictor model $m$.
    \item Train $m$ on $\D$.
    \item\label{s-prune} Using the trained and fixed model $m$, prune the dataset $\D$ to $\G \subseteq \D$ by first eliminating the instances $\bfx$ that do not satisfy Equation \ref{ssx}, and then by further eliminating the instances for which $\scr = \emptyset$ according to Equation \ref{eq:clearly-rel}.
    %such that for every instance $\bfx \in \D'$ we are guaranteed that (i) all $x_i \in \nsel$ are irrelevant for the prediction $m(\bfx)$, and (ii) there exists $\scr \subseteq \sel$, 
    \item Using the metrics defined above, test an explainer on $m$ applied to instances in $\G$.
\end{enumerate}

\begin{table*}[t]
\centering
\caption{Error rates of the explainers on the sanity tests generated by \vf. For an explainer to pass this sanity tests, its errors should be all $0$. Best results (lowest errors) are in bold, although the tests are not comprehensive, so one cannot fully compare the explainers based on their performance on these tests. $\text{AVG}_{\text{misrnk}}$ reports an average with the standard deviation in parentheses.\label{tab:results} }
%\vskip 0.15in
\begin{small}
\resizebox{\textwidth}{!}{
\begin{tabular}{ l  c  c  c  c  c  c  c  c c c c }
\multicolumn{1}{c}{}& \multicolumn{3}{c}{{\large{APPEARANCE}}}&\multicolumn{3}{c}{{\large{AROMA}}}&\multicolumn{3}{c}{{\large{PALATE}}}  \\ 
	\hline \\
	\multicolumn{1}{c}{Model} & $\text{PCT}_{\text{first}}$ & $\text{PCT}_{\text{misrnk}}$ & 
	$\text{AVG}_{\text{misrnk}}$ & $\text{PCT}_{\text{first}}$ & $\text{PCT}_{\text{misrnk}}$ & $\text{AVG}_{\text{misrnk}}$ & $\text{PCT}_{\text{first}}$ & $\text{PCT}_{\text{misrnk}}$ & $\text{AVG}_{\text{misrnk}}$ \\ 
	\hline \\
	
 	 LIME & \textbf{4.24} & 24.39 & 7.02 (24.12) & 14.79 & 32.08 & 12.74 (33.54) & 2.92 & 13.93 & \textbf{3.48} (17.38) \\

 	 KernalSHAP & 4.74 & \textbf{16.81} & \textbf{1.16} (7.75) & \textbf{4.24} & \textbf{13.53} & \textbf{0.83} (7.10) & \textbf{2.65} & \textbf{9.20} & 9.25 (9.70) \\

	 L2X & 6.58 & 28.85 & 3.54 (12.66) & 12.95 &	31.61 &	4.41 (16.25) & 12.77 & 29.83 & 3.70 (13.05) \\
 
\end{tabular}}
\end{small}
%\vskip -0.1in
%\end{adjustbox}
\end{table*}

\section{Experiments}\label{sec:exp_analysis}

To prove the effectiveness of \vf, I instantiate it on the RCNN model introduced by \citet{rcnn} and apply it to the task of sentiment analysis in natural language.

\paragraph{The RCNN.} The RCNN is a selector-predictor type of model, for which:
\begin{enumerate}
    \item the selector (called generator in their paper) takes as input a piece of text $\textbf{x}$ and, for each feature (token in our case) $x_i \in \bfx$, outputs $p_i \in [0,1]$. Then, $p_i$ is used as the parameter of a Bernoulli distribution modelling the probability that the token $x_i$ is selected, and
    \item the predictor (called encoder in their paper) takes as input $\sel = \{x_i \sim \text{Bernoulli}(p_i)\}$ (at test time $\sel = \{x_i | p_i \ge 0.5 \}$) and returns the prediction as a real number in $[0,1]$.
\end{enumerate}

Both the selector and the predictor are recurrent convolutional neural networks \citep{lei15} (hence, the name of RCNN). The selector has the additional property of being bidirectional, so that the decision to select a token is based on the entire context around the token. There is no direct supervision on the subset selection, and the selector and predictor are trained jointly, with supervision only on the final prediction. To train the RCNN, two regularizers are deployed: one to encourage the selection of fewer tokens, and a second one to encourage the selection of a subphrase, rather than disconnected tokens. The latter is employed because in the dataset on which this model is applied, the relevant tokens usually form a subphrase.
At training time, to circumvent the non-differentiability introduced by the intermediate sampling, the gradients for the selector are estimated using the REINFORCE procedure~\citep{reinforce}. 

\paragraph{Dataset.} 
In this experiments, I use the BeerAdvocate corpus,\footnote{\url{http://people.csail.mit.edu/taolei/beer/}}
on which the RCNN was initially evaluated~\citep{rcnn}.
BeerAdvocate consists of a total of ${\sim}100$K human-generated multi-aspect beer reviews, where the three considered aspects are: appearance, aroma, and palate. Even though it appeared to be only one dataset with each review containing information about all three aspects, in the provided link to the corpus there are separate datasets for each aspect, which appeared to be slightly different (for example, the training set for the appearance aspect has $10$K more instances than the training set for the other two aspects).
The reviews are accompanied by fractional ratings between $0$ and $5$ for each aspect independently, which I rescale between $0$ and $1$, similarly to \citet{rcnn}. 
Also following the procedure in \citet{rcnn}, I train three separate RCNN models, one for each aspect independently, with the code from the original paper. I did not do any hyperparameter search but used the default settings in their code.\footnote{\url{https://github.com/taolei87/rcnn}} %The training was executed on a single GPU ({\sc TITAN XP} with 12GB of memory).
%We hence, train three RCNNs on three datasets. 

%For each aspect $a$, we check whether (i) all $x_i \in \nsel$ have zero-contribution, and (ii) if there exists $\scr \in \sel$. If one of the two conditions is not met, then the point is discarded and not deployed to evaluate the explainers. 
For each aspect $a$ and the trained model $\text{RCNN}_a$, I obtain a pruned version %$\mathcal{G}^{\text{RCNN}_a}$ according to step \ref{s-prune} of our framework. For simplicity, I refer to the pruned datasets $\mathcal{G}^{\text{RCNN}_a}$ as $\ga$. 
$\ga$ according to step \ref{s-prune}.\footnote{Note that each $\ga$ depends on the trained $\text{RCNN}_a$ model. For the same RCNN architecture trained by starting from another initialisation, or with different hyperparameters, one may obtain a different $\ga$.} 
To obtain the independently relevant tokens, I choose a threshold of $\tau = 0.1$ in Equation \ref{eq:clearly-rel}. Since the scores are in $[0, 1]$ and the ground-truth ratings correspond to \{$0$, $0.1$, $0.2$, $\ldots$ , $1$\}, a change in prediction of $0.1$ is significant for this dataset. %, and hence, this threshold ensures that we only penalize the explainers when we are sure they have misranked an independently relevant feature.

\begin{table*}[t]
\caption{Statistics of the datasets $\ga$ for each aspect. $|\ga|$ is the number of instances after pruning dataset $\D$, $|\bfx|$ is the average length of the reviews in $\ga$. $|\sela|$, $|\scra|$, and $|\nsela|$ are the number of selected tokens, independently relevant selected tokens, and non-selected tokens, respectively. In parentheses are the standard deviations. The column $\%(\text{\textbf{S}}^a_{\sela} \neq \sela)$ provides the percentage of instances eliminated due to the possibility that the model might not have selected a sufficient subset of tokens. Finally, $\%(\scra = \emptyset)$ shows the percentage of instances (out of the remaining ones after the first pruning) further eliminated due to the absence of selected tokens that alone change the prediction by at least $0.1$.}
\label{data-stats}
%\vskip 0.15in
\begin{small}
\begin{center}
\resizebox{\textwidth}{!}{{\begin{tabular}{l c c c c c c c}%{@{}l c c c c c c c@{}}
\large{Aspect ($a$)} & \large{$|\ga|$} & \large{$|\bfx|$} & \large{$|\sela|$} & \large{$|\scra|$} & \large{$|\nsela|$} & \large{$\%(\text{\textbf{S}}^a_{\sela} \neq \sela)$} & \large{$\%(\scra = \emptyset)$} \\
\midrule
APPEARANCE & 20508 & 145 (79) & 16.9 (8.4) & 1.33 (0.70) & 121 (56) &  15.9 & 73.2 \\
AROMA & 7621 & 139 (74) & 11.15 (6.48) & 1.16 (0.49) & 123 (57) & 72.0 & 58.0 \\
PALATE & 16494 & 153 (76) & 9.14 (5.38) & 1.21 (0.55) & 137 (59) &  39.2 & 66.5 \\
\end{tabular}}}
\end{center}
\end{small}
%\vskip -0.1in
\end{table*}

Table \ref{data-stats} shows the statistics of the $\ga$ datasets. We see that we detect only one or two independently relevant tokens per instance, showing that the threshold of $0.1$ is likely very strict. However, it is better to be more conservative in order to ensure that the sanity tests do not wrongly penalise explainers. We also see that the percentages of instances eliminated in order to ensure the sufficiency condition (Equation \ref{ssx}) can be quite large, up to $72\%$ for the aroma aspect. Therefore, as future work, it would be interesting to check whether the model indeed provides significantly different predictions for such a large proportion of instances, or whether it just makes slightly different selections that still lead to the same final prediction. Nonetheless, for the scope of this work, we are left with a large number of instances for each sanity check.

\paragraph{Verifying explainers.} I test three popular explainers: LIME \citep{lime}, KernalSHAP \citep{shap}, and L2X \citep{l2x}. I use the code of each explainer as provided in the original repositories,\footnote{ \url{https://github.com/marcotcr/lime/tree/master/lime}, \\ \url{https://github.com/slundberg/shap},\\ \url{https://github.com/Jianbo-Lab/L2X/tree/master/imdb-token}.} with their default settings for text explanations, with the exceptions that: (1)~for L2X, I set the dimension of the word embeddings to $200$ (the same as in the RCNN) and I increase the maximum number of training epochs from $5$ to $30$, and (2)~for LIME and KernalSHAP, I increase the number of samples per instance to $10$K, since the length of the input text was relatively large (${\sim}240$ tokens per instance). Due to time constraints, I run each explainer only once, yet the number of instances on which the explainers are tested is rather large (${\sim}44.5$K in total over the three aspects).

%\paragraph{Type of ground-truth explanation.}
As mentioned in Chapter~\ref{chap-difficulty}, LIME and KernalSHAP adhere to the neighbourhood type of ground-truth explanation, hence, this verification is not directly targeting these explainers. However, we see in Table \ref{tab:results} that, in practice, LIME and KernalSHAP outperformed L2X on the majority of the metrics, even though L2X is a minimal sufficient subsets explainer. There are a couple of hypotheses that could explain why this is the case. First, it might be the case that the RCNN models learn to select most of the redundant features, or that there are no redundant features in the dataset. Hence, it is possible that the sanity tests are equally valid for both the pointwise and the neighbourhood type of ground-truth explanations. In Section \ref{sec:improvements}, I provide a further check that can be done to examine this. Second, a major limitation of L2X is the requirement to know in advance the number of sufficient features per instance. Indeed, L2X learns a distribution over the set of features by maximising the mutual information between the prediction of the target model on subsets of $K$ features and the prediction of the target model on the full instance, where $K$ is assumed to be known and the same for all instances. In practice, one usually does not know how many features per instance a model relies on. To test L2X in real-world circumstances, I set $K$ to the average number of tokens highlighted by human annotators on the subset manually annotated in \citet{beer-annot}. I obtained an average $K$ of $23$, $18$, and $13$ for the three aspects, respectively.

In Table~\ref{tab:results}, we see that, on metric $\text{PCT}_{\text{first}}$, all explainers are prone to stating that the most relevant token is an irrelevant token, as much as $14.79\%$ of the time for LIME and $12.95\%$ of the time for L2X in the aroma aspect. %We consider this the most dramatic form of failure since those tokens were proven to not influence at all the prediction. 
The results on metric $\text{PCT}_{\text{misrnk}}$ show that all explainers tend to rank at least one irrelevant token higher than an independently relevant one, i.e., there is at least one error in the predicted ranking. Finally, the results on metric $\text{AVG}_{\text{misrnk}}$ show that, on average, KernalSHAP only places one irrelevant token ahead of any independently relevant token for the first two aspects but as much as 9 tokens for the third aspect, while L2X places around 3-4 irrelevant tokens ahead of an independently relevant token for all three aspects.

\begin{figure*}[h]
        \centering
        \includegraphics[width=0.9999\linewidth,trim={1cm 17cm 1cm 12cm},clip]{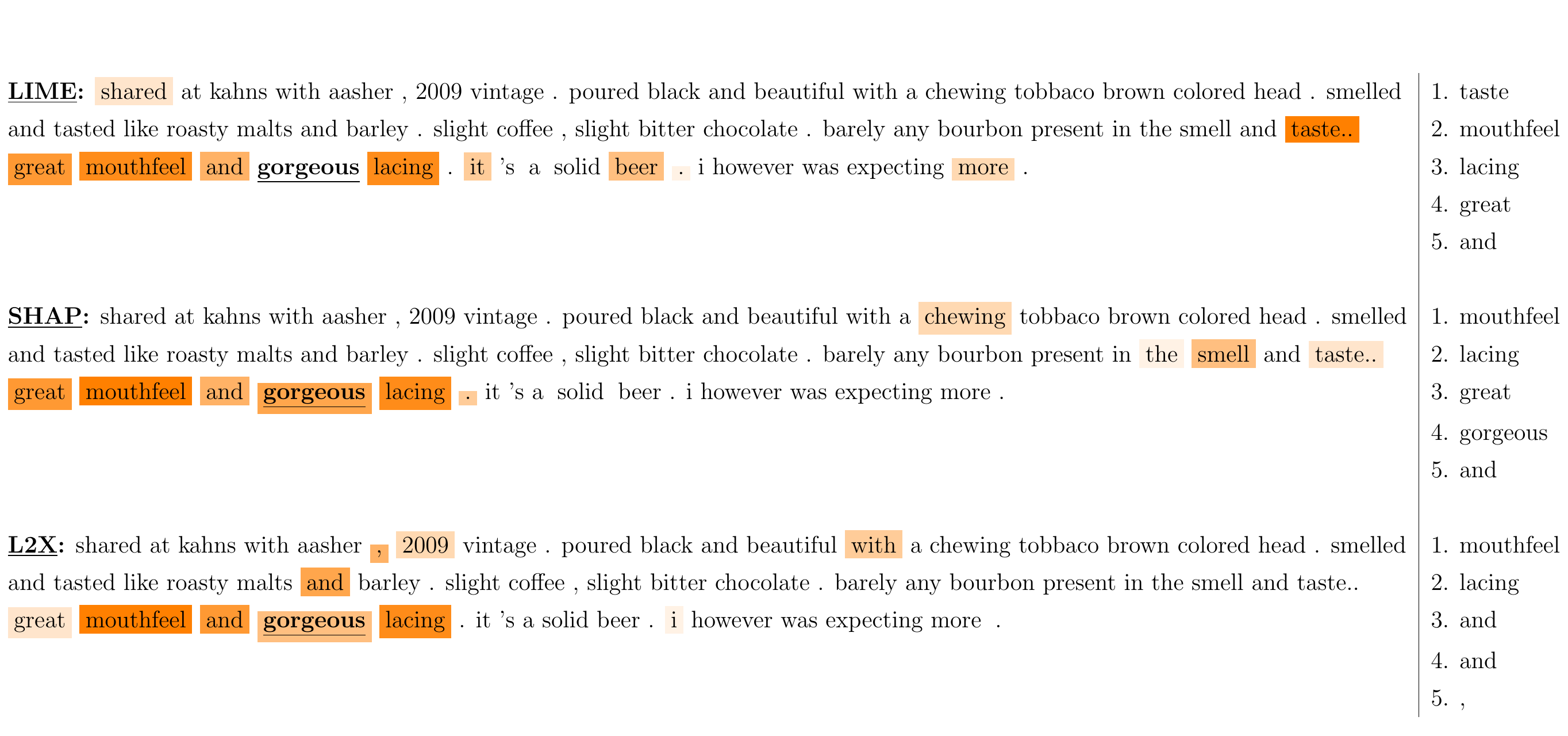}
        \caption[Explainers rankings on an instance from the palate sanity test]{Explainers rankings on an instance from the palate sanity test. The top 5 features ranked by each explainer are on the right-hand side. Additionally, the heatmap corresponds to the ranking provided by each explainer, where the intensity of the colour decreases linearly with the rank of the token.\footnote{While for LIME and KernalSHAP, one could use the actual weights for the heatmap, for consistency with L2X, and since the explainers are only verified on their rankings, I keep the linearly-decreasing heatmap.} For visibility reasons, only the first $K=10$ ranked tokens are shown.
        The tokens selected by the model, $\sela$, are in bold (in this case only one), and the detected independently relevant tokens, $\scra$, are additionally underlined (also one in this case). The non-bold tokens should not be ranked higher than any bold and underlined token. KernalSHAP is simply referred as SHAP.
        \label{fig:qualitative} }
\end{figure*}

\paragraph{Qualitative analysis.}%\label{p:qualit} 
In Figure~\ref{fig:qualitative}, I present an example from the palate sanity test, i.e., the sanity test created with the RCNN trained for the palate aspect. Examples from the aroma and appearance sanity tests are in Appendix~\ref{more-examples-sanity}. % Tokens in $\sela$ are in bold, and the independently relevant tokens from $\scra$ are additionally underlined. The 5 top-ranked tokens by each explainer are also mentioned on the right-hand side. 

Notice that all explainers are prone to rank tokens that were irrelevant for the prediction of the model higher than tokens on which the model actually relied on for its prediction. % as well as to attribute importance to very random features such as punctuations. %, with LIME and KernalSHAP even ranking the tokens ``mouthfeel'' and ``lacing'' belonging to $\nsel$ as first two most important features. % and L2X ranking it as second.
For example, ``gorgeous'', the only token used by the model on the given instance, was given almost zero weight by LIME, and did not even make it into the top five tokens given by L2X. Instead, L2X gives ``mouthfeel'', ``lacing'', ``and'', and even ``,'' as most important tokens. Note that such an explanation would likely give humans a less good impression and diminish their trust in the model, even though the model actually relied on the informative token ``gorgeous''.

\section{Specifications}
\label{specs}

As mentioned in Chapter \ref{chap-difficulty}, it is essential for works introducing explainers or verification frameworks for explainers to state the intended behaviour, strengths, and limitations of their methods. Hence, in this section, I provide the specifications for \vf.

\subsection{Intended Behaviour}

\paragraph{Type of ground-truth explanation.} \vf{} is tailored for the pointwise type of ground-truth explanations. Hence, it \textit{might} penalise an explainer for ranking a feature that is redundant in the presence of other features higher than a strictly necessary feature. For example, \vf{} might penalise an explainer that ranks ``nice'' higher than ``very'' for the model $m$ on the instance $\bfx^{(2)}$ in Figure \ref{2gt}. However, if the selector-predictor model also selects redundant features (in our example, if it selects ``nice'' in addition to ``very'' and ``good''), then the framework does not penalise an explainer regardless of the order in which it ranks these three features. In Section \ref{sec:improvements}, I provide a way of checking whether for a selector-predictor model and an instance, there exist redundant features that are not selected. This check will allow the extension of the framework to verifying explainers in the case when neighbourhood explanations are desired. Meanwhile, the current framework and results should be used only when pointwise ground-truth explanations are desired. Consequently, what is called \textit{error} in this chapter refers to an error under the pointwise perspective, which, as we saw, may not be an error under the neighbourhood perspective.

\paragraph{A sanity check, not a complete evaluation.} The sanity tests provided by this framework cannot be used as a complete evaluation for concluding the faithfulness of explainers in full generality. This is because we do not know the full behaviour of the model and thus cannot provide a ranking of all features.
Furthermore, similarly to the BAM framework \citep{bam}, the \vf{} framework does not guarantee to identify \textit{all} false positives. This is because certain features from $\sel$ might have been relevant for the prediction even if they do not pass the independently relevant check from Equation \ref{eq:clearly-rel}. Thus, \vf{} generates necessary but not sufficient sanity tests.

\paragraph{Ranking.} The current metrics assume that the explanatory methods give a ranking of the features in terms of their relevance to the prediction of the target model. However, the metrics can be adapted to methods that provide only subsets of important features.

\subsection{Strengths}

\paragraph{Non-trivial neural model, real-world dataset, and partially known decision-making process.} 
As mentioned before, an important challenge in verifying explanatory methods is how to automatically sanity check these methods on neural models similar to the ones deployed in the real world and without speculating the decision-making processes of these networks. To my knowledge, the verification framework introduced in this chapter is the first to take a step in this direction. While the current version of the framework still contains certain unknowns with respect to the decision-making process of the testbed model, the directions of further improvements provided in Section \ref{sec:improvements} aim to alleviate most of these unknowns.

\paragraph{Artefact-caused minimal sufficient subsets.}
Another important strength of this verification framework is that it provides (potentially a superset of) the exact minimal sufficient subset on which the target model relied for its prediction on an instance. This allows us to penalise explainers for providing artefact-based minimal sufficient subsets, which, as we saw in Chapter \ref{chap-difficulty}, would not be faithful explanations and can consequently distort the perception and trust of users in the model. This would not be possible to be checked simply by probing whether a subset provided by an explainer is sufficient.

%\paragraph{Artefact-caused cancellations.}
% While artefact-caused cancellations may arguably be interesting to be presented in an explanations, it is also debatable whether they should be ranked higher than the features that actually lead to the prediction. 
% In addition, our framework also has a chance to penalize 

\subsection{Limitations}
The current version of the framework has two major limitations that I describe below. While these cases might not happen often in practice, it is important to acknowledge them. In Section \ref{sec:improvements}, I provide checks that can alleviate these limitations.

\paragraph{Genuine cancellations.} If the selector-predictor model does not select groups of features that form a genuine cancellation, then the framework would penalise an explainer for placing these features higher than an independently relevant selected feature. If additionally, this independently relevant selected feature is part of an artefact-caused cancellation, then this would arguably be a wrong penalisation. However, for this to happen, two conditions need to be met: the model would have to (1)~select features from artefact-caused cancellations, and (2)~not select features from genuine cancellations. Hence, it is reasonable to believe that this would not happen often in practice.

\paragraph{Genuine minimal sufficient subsets.} If for an instance there are several genuine minimal sufficient subsets and if the selector-predictor model does not select all of them, then our framework would penalise an explainer for ranking features from a non-selected yet equally genuine minimal sufficient subset higher than features from selected genuine minimal sufficient subsets. This would arguably be an incorrect penalisation for the explainer. It is unclear how often this scenario happens in practice. %Contrary to the scenario causing the previous limitation, the scenario causing this limitation may happen more often given the 

\section{Directions and Guidelines of Improvement}
\label{sec:improvements}

In this section, I provide three directions for extending and improving the \vf{} framework, with concrete guidelines. %by using, for example, the following additional checks.

\paragraph{Check for non-selected redundant features.}
To adapt \vf{} to the cases where the neighbourhood type of explanation is desired, one can further eliminate certain instances that contain non-selected redundant features based on the following observation: if the selector-predictor model applied to the instance formed only by the non-selected features gives a prediction that is significantly different from the prediction of the model on the baseline input, then there is a redundant feature among the originally non-selected features. Formally, if for a selector-predictor $m$ and an instance $\bfx$, we have that $m(\nsel) \neq m(\textbf{b})$, where $\textbf{b}$ is the baseline input for the task at hand, then there must be a redundant feature among $\nsel$, and this feature would be among the newly selected features $\text{sel}(\nsel)$.  

However, $m(\nsel) = m(\textbf{b})$ does not guarantee the absence of redundant features among $\nsel$. This is because there might be redundant features that require the presence of certain originally selected features in order to \textit{be activated}. 
For example, assume that ``nice'' is redundant in the presence of ``very good'' and that the model also requires the presence of the feature ``tastes'' in order to activate these sentiment-providing features. Then, on an instance such as ``The beer tastes nice, very good.'' the model might select ``tastes'', ``very'', and ``good'', and the check above would not identify ``nice'' as a redundant feature because ``tastes'' is missing from $\nsel$. 

%Note that this check being done in addition to the sufficiency check of $m(\sel) \neq m(\bfx)$ can not result in flagging artefact-caused redundancies. This is because $m(\sel) \neq m(\bfx)$ guarantees that the originally selected features are sufficient

%Note that even if time would not be a problem to check 
Future work may focus on how to guarantee the absence of redundant features instead of simply decreasing the probability of their existence.

% \begin{equation}
%     \text{abs}(m(\nsel) - m(\textbf{b})) > \tau \text{, then there }
% \end{equation}

\paragraph{Check for other genuine minimal sufficient subsets.}
Similarly to the check above, if $m(\nsel) = m(\bfx)$, then there is another genuine minimal sufficient subset among the non-selected features. However, this check does not account for genuine equivalent subsets that intersect with the originally selected features.
As before, this check only reduces the probability of wrongly penalising explainers. Future work may find ways to provide a guarantee. 

\paragraph{Check for non-selected features that cancel each other out.} 
To decrease the probability of having groups of features that cancel each other out among the non-selected features, one can check if the elimination of each non-selected feature alone changes the prediction of the model. More precisely, if there exists a non-selected feature $x_i \in \nsel$ such that $m(\bfx \setminus \{x_i\}) \neq m(\bfx)$, then there exists a group of features that cancel each other out that the model did not select. However, this check is again not enough to guarantee the absence of such a group of features among the non-selected features. This is because there can exist groups of features that cancel each other out but for which we need to eliminate more than one feature at a time to observe a change in prediction. For example, if the cancellation follows a rule such as [``amazing'' or ``awesome'' cancel out with either ``horrible'' or ``awful''], then for an instance that contains all the four features, eliminating either of them in isolation does not change the prediction. %However, checking for all possible 
%$(x_i \text{ OR } b) \text{ AND } (c \text{ OR } d)$ and all features are true, then the cancellation would not be detected by eliminating each feature at a time. 

\paragraph{Further improvements.}
Doing all of the above eliminations might lead to a small number of instances in the final dataset $\G$. In this case, one may salvage more instances in the following ways. First, one can check the sufficiency condition using the equality in prediction $m(\sel) \neq m(\bfx)$ rather than the stronger condition in Equation \ref{ssx}. While, in theory, there is no guarantee that any instance at all would be salvaged in this way, in practice, I noticed that the difference between the two predictions was often very small even when the newly selected features were different than the originally selected ones. Second, instead of eliminating instances for which these conditions do not hold, one can adjust the sets of relevant and irrelevant features according to the newly selected and non-selected features. Such investigation is left as future work.

\section{Conclusions and Open Questions} % \label{sec:concl}
In this chapter, I introduced a framework capable of generating automatic sanity tests to verify whether post-hoc explainers are ranking features that are irrelevant for the prediction of a model on an instance higher than features that are relevant for the prediction of the model on that instance. I instantiated the framework on a task of sentiment analysis and produced three sanity tests, on which I tested three explainers, namely, LIME, KernalSHAP, and L2X. 

This framework unveiled certain problems of current explanatory methods, such as the fact that they may point to an irrelevant feature as the most important feature for a prediction. Identifying such problems is a first step towards improving these methods in the future. 

 is generic and can be instantiated on other tasks and domains, while the three sanity tests can be used off-the-shelf to test other existing and future explainers. %While the there are limitations in the current version of the framework, there is 

Future work includes performing the improvements described in Section \ref{sec:improvements}, instantiating \vf{} on other tasks and modalities, as well as testing more explanatory methods.

%By showing both theoretical and empirical examples on which these methods fail to explain a target model faithfully, I~aim to inspire the development of more robust explanatory methods.

%Finally, this chapter emphasises the importance of a deep and rigorous understanding of the decision-making process of a target model before using it as a testbed for verifying explanatory methods.

%% file: esnli.tex
\chapter{Neural Networks that Generate Natural Language Explanations}
\label{chap-esnli}

In this chapter, which is based on \citep{esnli}, I explore self-explanatory neural models. More precisely, I investigate
whether neural models improve their behaviour %and performance 
if they are additionally given natural language explanations for the ground-truth labels at training time, and whether these models can generate such explanations for their predictions at test time.

\section{Motivation}
As mentioned in Chapter \ref{chap-intro}, models trained simply to obtain a high accuracy on held-out sets can often learn to rely on shallow input statistics, resulting in brittle models. % susceptible to adversarial attacks.
For example, \citet{lime} present a document classifier that distinguishes between {\it Christianity} and {\it Atheism} with a test accuracy of $94\%$. However, on close inspection, the model spuriously separates classes based on words contained in the headers, such as ``Posting'', ``Host'', and ``Re''. 
Spurious correlations in both training and test sets allow for such undesired models to obtain high accuracies. Much more complex hidden correlations may be present in any arbitrarily large and human-annotated dataset \citep{artifacts, dasgupta2018evaluating, artifacts-CNN, biases3, biases4}. Such correlations may be difficult to spot, and even when one identifies them, it is an open question how to mitigate them \citep{mitigate-artefacts}. 

In this chapter, I investigate a direction that has the potential to both steer neural models away from relying on spurious correlations and provide explanations for the predictions of these models. This direction is that of enhancing neural models with the capability to learn from natural language explanations during training time and to generate such explanations at test time. For humans, it has been shown that explanations play a key role in structuring conceptual representations for categorisation and generalisation \citep{expl-cog1, expl-cog2}. Humans also benefit tremendously from reading explanations before acting in an environment for the first time \citep{atari}. Thus, explanations may also be used to set a model in a better initial position to further learn the correct functionality. Meanwhile, at test time, generating correct argumentation in addition to obtaining a high accuracy has the potential to endow a model with a higher level of transparency and trust.   

%In this work, we introduce a new dataset and models for exploiting and generating explanations for the task of recognizing textual entailment. 
Incorporating external knowledge into a neural model was shown to result in more robust models \citep{breaking-nli}. % show that models achieving high accuracies on SNLI, such as \citep{snli-1, snli-2, esim}, show dramatically reduced performance on this simpler dataset, while the model of \citet{kim} is more robust due to incorporating external knowledge. 
Free-form natural language explanations are a form of external knowledge that has the following advantages over formal language. First, it is easy for humans to provide free-form language, eliminating the additional effort of learning to produce formal language, thus making it simpler to collect such datasets. Secondly, natural language explanations might potentially be mined from existing large-scale free-form text. Finally, natural language is readily comprehensible to an end-user who needs to assert the reliability of a model. 
%Thirdly, the formal languages chosen by researchers may differ from work to work and therefore models constructed over one formal language might not be trivially transferred to another. Meanwhile free-form explanations are generic and applicable to diverse areas of research, such as natural language processing, computer vision, or policy learning. 

Despite the potential for natural language explanations to improve both learning and transparency, there is a scarcity of such datasets in the community, as discussed in Section \ref{nle}. 
To address this deficiency, I collected a large corpus of ${\sim}570$K human-annotated explanations for the SNLI dataset~\citep{snli}. I chose SNLI because it constitutes an influential corpus for natural language understanding that requires deep assimilation of fine-grained nuances of commonsense knowledge.
%A plethora of models have been developed on this dataset, including previous state-of-the-art in universal sentence representations \citep{infersent}, which demonstrates the power of this task and dataset.
I call this explanation-augmented dataset e-SNLI, which I release publicly\footnote{The dataset can be found at \url{https://github.com/OanaMariaCamburu/e-SNLI}.} to advance research in the direction of training with and generation of free-form natural language explanations. 
 
%To demonstrate the efficacy of the e-SNLI dataset, 
%I show that it is much more difficult for neural models to produce correct natural language explanations based on spurious correlations than it is to produce correct labels. Further, I develop models that predict a label and generate an explanation for their prediction. I also investigate how the presence of natural language explanations at training time can guide neural models into learning better universal sentence representations \citep{infersent} and into having better capabilities to solve out-of-domain instances.

Secondly, I show that it is much more difficult for a neural model to produce correct natural language explanations based on spurious correlations than it is for it to produce correct labels based on such correlations. 

Thirdly, I develop models that predict a label and generate an explanation for their prediction, and I investigate the correctness of the generated explanations. 

Finally, I investigate whether training a neural model with natural language explanations can result in better universal sentence representations produced by this model and in better performance on out-of-domain datasets.

\paragraph{Remark.} In this chapter, I use the concept of correct explanation to refer to the correct argumentation for the ground-truth label on an instance. 
This should not be confused with the concept of faithful explanation, which refers to the accuracy with which an explanation describes the decision-making process of a model, as described in Section \ref{faithful}. 
The capability of a neural model to generate correct explanations is an important aspect of the development of such models. 
For example, correct argumentation may sometimes be needed in practice, alongside the correct final answer.
Hence, in this chapter, I inspect the correctness of the explanations generated by the introduced neural models. In the next chapter, I will take a step towards verifying the faithfulness of these explanations.% is given in Chapter \ref{chap-inconsist}.

\section{Background}
The task of recognising textual entailment is a critical natural language understanding task. Given a pair of sentences, called the premise and hypothesis, the task consists of classifying their relation as either (a)~{\it entailment}, if the premise entails the hypothesis, (b)~{\it contradiction}, if the hypothesis contradicts the premise, or (c)~{\it neutral}, if neither entailment nor contradiction hold. For example, the premise ``\emph{Two doctors perform surgery on patient.}'' and the hypothesis ``\emph{Two doctors are performing surgery on a man.}'' constitute a neutral pair (because one cannot infer from the premise if the patient is a man).
The SNLI dataset \citep{snli}, containing ${\sim}570$K instances of human-generated triples (premise, hypothesis, label), has driven the development of a large number of neural models \citep{RocktaschelGHKB15, snli-1, snli-2, esim, snli-4, kim, infersent}. %The follow-up dataset of MultiGenre natural language inference (MultiNLI)~\citep{multinli}, in the same format as SNLI and comparable in size, includes a more diverse range of genres of written and spoken English, as well as test sets for cross-genre transfer evaluation.

Moreover, the power of SNLI transcends the task of natural language inference. \citet{infersent} showed that training a model to produce universal sentence representations on SNLI can be both more efficient and more accurate than certain training approaches on orders of magnitude larger but unsupervised datasets \citep{skip, fastsent}. 
%Their results further improved when training on the union of SNLI and MultiNLI, becoming the previous state-of-the-art in universal sentence representations. 
I take this approach one step further and, in Section \ref{expthree}, I investigate whether the additional layer of natural language explanations can bring further improvement. 

%\subsubsection{Artifacts in SNLI and natural language inference Models} 
Recently, an increasing amount of analysis has been carried out on the spurious correlations in the SNLI dataset and on how different models rely on these correlations \citep{dasgupta2018evaluating, artifacts, breaking-nli}. In particular, \citet{artifacts} show that specific words in the hypothesis tend to be strong indicators of the label, e.g., ``friends'' and ``old'' appear very often in neutral hypotheses, ``animal'' and ``outdoors'' appear most of the time in entailment hypotheses, while ``nobody'' and ``sleeping'' appear mostly in contradiction hypothesis. %cast doubt on whether models trained on SNLI are learning to understand language, or are largely fixating on spurious correlations, also called artefacts. For example, they show that specific words in the hypothesis tend to be strong indicators of the label, e.g., {``friends'', ``old''} appear very often in neutral hypotheses, {``animal'', ``outdoors''} appear most of the time in entailment hypotheses, while {``nobody'', ``sleeping''} appear mostly in contradiction hypothesis. %Sentence lengths seem to also be correlated with labels, with neutral hypotheses being generally significantly longer than entailed ones. 
They also show that a premise-agnostic model, i.e., a model that only takes as input the hypothesis and predicts the label, obtains $67\%$ test accuracy. In Section \ref{expzero}, I show that it is much more difficult to rely on spurious correlations to generate explanations than to generate labels.

\section{The e-SNLI Dataset}

\begin{figure}
  \centering
  \resizebox{0.99\linewidth}{!}{
  \begin{tabular}{l}
    \toprule

    \textsc{Premise}: An adult dressed in black \hl{holds a stick}. \\ 
    \textsc{Hypothesis}: An adult is walking away, \hl{empty-handed}. \\ 
    \textsc{Label}: contradiction\\ 
    \textsc{Explanation}: Holds a stick implies using hands so it is not empty-handed.\\

    \midrule
    
    \textsc{Premise}: A child in a yellow plastic safety swing is laughing as a dark-haired woman \\
    in pink and coral pants stands behind her.\\ 
    \textsc{Hypothesis}:	A young \hl{mother} is playing with her \hl{daughter} in a swing.\\ 
    \textsc{Label}: neutral\\ 
    \textsc{Explanation}: Child does not imply daughter and woman does not imply mother.\\

    \midrule
    
    \textsc{Premise}: A \hl{man} in an orange vest \hl{leans over a pickup truck}. \\ 
    \textsc{Hypothesis}: A man is \hl{touching} a truck.\\ 
    \textsc{Label}: entailment\\ 
    \textsc{Explanation}: Man leans over a pickup truck implies that he is touching it. \\

    \bottomrule
\end{tabular}}
  \caption[Examples from the e-SNLI dataset]{Examples from the e-SNLI dataset. Annotators were given the premise, the hypothesis, and the label. They highlighted the words that they considered essential for the label and provided the explanations.%\vspace*{-5ex}
}
  \label{examples_table}
\end{figure}

In this section, I present the methodology that I used to collect e-SNLI on Amazon Mechanical Turk. The main question that this dataset had to answer was: {\it Why is a pair of sentences in a relation of entailment, neutrality, or contradiction?} I encouraged the annotators to focus on the salient elements that induce the given relation, and not on the parts that are repeated identically in the premise and hypothesis. I also asked them to explain all the parts of the hypothesis that do not appear in the premise. %For neutral and contradictory pairs, while I encouraged stating all the elements that contribute to the relation, I considered an explanation correct, if at least one element is stated. Finally, 
Additionally, I asked the annotators to provide self-contained explanations, as opposed to sentences that would make sense only after reading the premise and hypothesis. For example, an explanation of the form ``Anyone can knit, not just women.'' would be preferred to the explanation ``It cannot be inferred they are women.''
%Although we accept both explanations as correct, the former could bring more benefit to the learning process as well as being more useful for future research directions, such as explanation-retrieval from a larger set of commonsense free-form facts mined from large scale text corpora. 

The main challenge of collecting this dataset is that, in crowd-sourcing, it is difficult to control the quality of free-form annotations. As a solution, I preemptively blocked the submission of obviously incorrect explanations by doing in-browser checks. For example, the annotators were not allowed to submit unless each explanation contained at least three tokens and it was not a copy of the premise or of the hypothesis. A second way of guiding the annotators to provide correct explanations was by asking them to proceed in two steps. First, they were required to highlight the words from the premise or hypothesis that they considered essential for the given relation. Secondly, annotators had to formulate each explanation using the words that they highlighted. In-browser checks were performed to ensure that a minimal number of words were highlighted and that at least half of the highlighted words were used in the explanation, so that the explanation is on-topic.
To account for the particularities of each relation, the minimal number of words required to be highlighted depended on the relation.
For entailment pairs, annotators were required to highlight at least one word from the premise. The annotators were also encouraged (but not required) to highlight words from the hypothesis.
For contradiction pairs, they were required to highlight at least one word in both the premise and in the hypothesis. 
For neutral pairs, they were required to highlight at least one word in the hypothesis, and they were not allowed to highlight words from the premise. This specific constraint was introduced to prevent workers from confusing the premise with the hypothesis. Neutral pairs were often confusing, since annotators were easily prone to focus on the details from the premise that could not be found in the hypothesis instead of the other way around. This was not the case for the contradiction pairs, since a contradiction relation between entities is usually symmetrical, nor for the entailment pairs, for which the relation between entities is intuitive (e.g., a human would naturally say that ``A dog is an animal.'' instead of ``An animal is a dog.''). Hence, requiring annotators to highlight words with restrictions specific for each relation was especially useful to place the annotators into the correct mindset, and additionally provided a way to filter incorrect explanations. Moreover, the highlighted words may also provide a valuable future resource, for example, for providing supervision for attention models or for evaluating them~\citep{RocktaschelGHKB15, snli-2}.
Finally, an in-browser check also verified that the annotators used other words than the ones highlighted, since a correct explanation would need to articulate a link between these words.

I collected one explanation for each instance in the training set and three explanations for each instance in the validation and test sets. Table \ref{examples_table} shows examples of collected explanations. There were 6325 workers with an average of $86 \pm 403$ explanations per worker.

Note that an explanation that could also be generated automatically, such as ``Just because [entire premise] doesn't mean [entire hypothesis]'' (for neutral pairs), ``[entire premise] implies [entire hypothesis]'' (for entailment pairs), or ``It can either be [entire premise] or [entire hypothesis]'' (for contradiction pairs), would be uninformative. Therefore, I assembled a list of possible such templates, given in Appendix \ref{uninformative-filtering}, which I used for filtering the dataset of such uninformative explanations. More precisely, I filtered an explanation if its edit distance to one of the templates was less than $10$ characters. I ran this template detection on the entire dataset and reannotated the detected explanations ($11\%$ in total).

\paragraph{Quality of collected explanations.}\label{partial-score}
In order to measure the quality of the collected explanations, I selected a random sample of $1000$ examples and manually graded their correctness between $0$ (incorrect) and $1$ (correct). For entailment, an explanation received a potential partial score of $\frac{k}{n}$ if exactly $k$ out of $n$ required arguments were mentioned. For neutral and contradiction pairs, one correct argument is enough to conclude a correct explanation. %However, if more arguments are given and at least one is incorrect,

The annotation resulted in a total error rate of $9.6\%$, with $19.5\%$ on entailment, $7.3\%$ on neutral, and $9.4\%$ on contradiction pairs. %The annotations were very strict, and any error in formu
The higher error rate on the entailment pairs is firstly due to partial explanations, as annotators had an incentive to provide shorter inputs, so they often only mentioned one argument. A second reason is that many of the entailment pairs have the hypothesis as almost a subset of the premise, prompting the annotators to paraphrase the premise or hypothesis. Strictly speaking, this would not be an error given that in those cases, there was nothing salient to explain.
However, it was considered an error in this annotation because one could still formulate an argumentation-like explanation rather than simply paraphrasing the input. Future work may investigate automatic ways to detect the incorrect explanations, so that one can reannotate them.

\section{Experiments}
In this section, I perform a series of experiments to investigate the capabilities of neural models to learn from the natural language explanations in e-SNLI as well as to generate such explanations at test time. First, I show that a model that relies on spurious correlations in SNLI to provide correct labels is not able to also provide correct explanations based on these correlations. % (this model is referred to as \expzero). 
Further experiments elucidate whether models trained on e-SNLI are able to: (i)~predict a label and generate an explanation for the predicted label, % (\expone), 
(ii)~generate an explanation then predict the label given only the generated explanation, % (\exptwo), 
(iii)~learn better universal sentence representations, and 
(iv)~perform better on out-of-domain natural language inference datasets.

Throughout the experiments, the models largely follow the architecture of the model called \infersent{} from \citet{infersent}.\footnote{I build on top of their code, which is available at \url{https://github.com/facebookresearch/InferSent}.}  
%I fixed the issues raised in \url{https://github.com/facebookresearch/InferSent/issues/51} and reproduced results.} 
More precisely, \infersent{} separately encodes the premise and the hypothesis using two bidirectional long short-term memory recurrent units (BiLSTM) with internal sizes of $2048$ \citep{lstm, bidir}. Max-pooling over timesteps is used for obtaining the vector representations of each sentence (of dimension $2 \cdot 2048 = 4096$, due to bidirectionality). Let $\textbf{u}$ be the vector representation of the premise, and $\textbf{v}$ the vector representation of the hypothesis. The final vector of features is $\textbf{f} = [\textbf{u}, \textbf{v}, |\textbf{u} - \textbf{v}|, \textbf{u} \odot \textbf{v}]$, which is passed to a multilayer perceptron (MLP) with three layers of dimension $512$ each and without non-linearities\footnote{The default in the original \infersent{} code.}, which predicts the label. This is a typical high-level architecture often employed on SNLI \citep{snli}.

%I fixed all the hyperparameters for these modules the best ones found in \citet{infersent}.

To generate explanations, I use a one-layer LSTM module, whose internal size is a hyperparameter with values among \{$512$, $1024$, $2048$, $4096$\}. I also experimented with gated recurrent units (GRUs) \citep{gru} for the decoder, however, the LSTMs units performed best in all experiments.
Recurrent dropout \citep{dropout} was also applied to the explanations decoder, with a rate fixed to $0.5$. 
%For model \expone, another hyperparameter appears, which is the coefficient corresponding to the ratio between the label loss and the explanation loss (with values in $\{0, 0.1, 0.2, \ldots , 1\}$).

To reduce the size of the output vocabulary for generating explanations, tokens that appear less than $15$ times in the training set of explanations are replaced with {\tt <UNK>}, resulting in an output vocabulary of ${\sim}12$K tokens. 

The preprocessing and optimisation were kept the same as in \citet{infersent}. In particular, to input sentences into a model, the already trained GloVe word embeddings were used and fixed throughout the experiments \citep{glove}. The premises and hypotheses were cut at a maximum of $84$ tokens, while and the explanations at a maximum of $40$ tokens (these limits were decided based on the statistics of dataset).

All models were trained with stochastic gradient descent (SGD) with the learning rate starting at $0.1$ and decayed by a factor of $0.99$ every epoch. I have experimented with the Adam optimizer \citep{adam}, however SGD performed best in all experiments. The batch size was fixed at $64$.

For the majority of the experiments, I use five seeds for the random number generator and provide the average performance with the standard deviation in parentheses. If no standard deviation is reported, the results are from only one seed.

\subsection{Premise Agnostic} %\expzero: Generate an Explanation Given only the Hypothesis}
\label{expzero}

\citet{artifacts} show that a neural model that \textit{only} has access to the hypotheses can predict the correct label $67\%$ of the times, by relying on spurious correlations in SNLI. Therefore, it is interesting to evaluate to what extent a model can also rely on spurious correlations to generate correct explanations.
%\newline

\paragraph{Model.} The model \expzero{} is formed of one BiLSTM encoder with max-pooling, which provides a vector representation for the hypothesis, and one LSTM decoder that takes this representation as its initial state as well as additional input at every timestep, and generates an explanation. 

For a fair comparison, I also separately train a model with the same architecture for the encoder of the hypothesis, but followed by a 3-layer MLP for predicting the label instead of generating an explanations. I call this model \expzerol. No hyperparameter search is performed for this model.

\paragraph{Model selection.} The hyperparameters for \expzero{} are the internal size of the decoder (as mentioned above, with values of $512$, $1024$, $2048$ and $4096$).
The model selection is performed via perplexity on the validation set, which strictly decreased when the decoder size was increased. However, for practical reasons, the decoder internal size was not increased beyond $4096$, which is therefore the chosen value for this hyperparameter for this model.

\paragraph{Results.} %The (average over 5 seeds) word perplexity of \expzero{} on the test set was $10.77 (0.03)$ and the BLEU score was $17.43 (0.45)$. However, since for evaluating generated natural language 
I manually inspected the explanations generated by \expzero{} for the first $100$ test instances, and obtained a correctness score of only $6.83$.\footnote{Partial scoring is explained in Section \ref{partial-score}.} On the other hand, \expzerol{} obtained $66$ correct labels for the same instances. This comforts the intuition that it is much more difficult (approx.\ 10 times for this architecture) to rely on spurious correlations to generate correct explanations than to rely on these correlations to predict correct labels. %This brings evidence that, when a model generates correct natural language explanations

\subsection{Predict then Explain} %\expone: Jointly Predict a Label and Generate an Explanation for the Predicted Label}
\label{expone}

In this experiment, I investigate the capability of a neural model to generate natural language explanations for its predictions. 
% that aims to justify the label predictions of the model. 

\begin{figure}[t]
        \centering
        \includegraphics[width=0.8\linewidth]{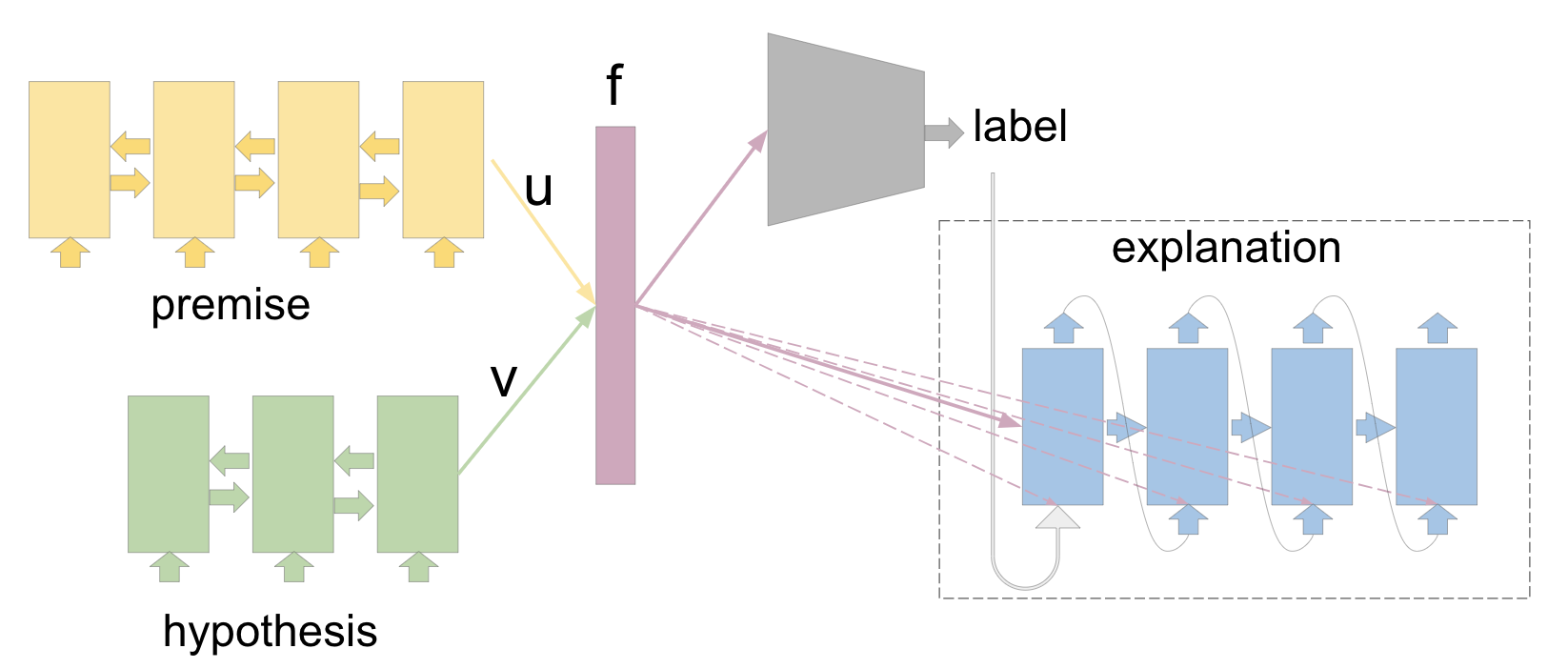}
        \caption[Architecture of the \expone{} model]{Architecture of the \expone{} model.}
        \label{fig:model}
\end{figure}

%the model introduced by \citet{infersent}, called BiLSTM-Max, since it follows a generic high-level architecture often employed on SNLI \citep{snli}. More precisely, BiLSTM-Max is formed of two 2048-BiLSTM networks with max-pooling that separately encode the premise, in a vector called $\textbf{u}$, and the hypothesis, in a vector called $\textbf{v}$. The final vector of features is $\textbf{f} = [\textbf{u}, \textbf{v}, |\textbf{u} - \textbf{v}|, \textbf{u} \odot \textbf{v}]$, which is passed to an MLP classifier that predicts the label. This is a typical high-level architecture often employed on SNLI \citep{snli, infersent}. In addition, a one-layer LSTM decoder for generating explanations is added by connecting the vector $\textbf{f}$ both as an initial state and concatenated to the word embedding at each timestep.

\paragraph{Model.} I enhance the \infersent{} model with an explanation generator by simply connecting the feature vector $\textbf{f}$ to a one-layer LSTM decoder, both as an initial state and concatenated to the word embedding input at every timestep. I call this model \expone{}. To condition the explanation on the label, the label is inputted at the first timestep of decoder (as the word ``entailment'', ``contradiction'', or ``neutral''). At training time, the gold label is provided, while at test time, the label predicted by the model is used. This architecture is depicted in Figure \ref{fig:model}. %This model is referred to as \expone.

\paragraph{Loss function.} Negative log-likelihood is used for both the classification loss and the explanation loss. Note that the explanation loss is much larger in magnitude than the classification loss, due to the summation of negative log-likelihoods over a large number of the words in the explanations. To account for this difference during training, a weighting coefficient $\alpha \in [0,1]$ is used, such that the overall loss is 
\begin{equation} 
  L_\text{total} = \alpha L_\text{label} + (1 - \alpha) L_\text{explanation}.
\end{equation}

\paragraph{Model selection.}
The hyperparameters in this experiment are the decoder internal size, $\alpha$ (for which I chose values from $0.1$ to $0.9$ with a step of $0.1$) and the learning rate (again, with values among $512$, $1024$, $2048$, and $4096$). 
To investigate how well a model can generate explanations without sacrificing prediction accuracy, in this experiment, only the label accuracy is used as the criterion for model selection. As future work, one can inspect different trade-offs between label and explanation performance. %In fact, in out next experiment we provide the other extreme of validating only for perplexity of the explanation and eliminating the classifier.
The best validation accuracy, of $84.37\%$, was obtained for $\alpha=0.6$ and the decoder internal size of $512$. %while \infersent{} with no explanations produced $84.30\%$ validation accuracy. 

\paragraph{Results.}
The test accuracy of \infersent{} over five seeds is $84.01\%$ $ (0.25)$. The \expone{} model obtains essentially the same test accuracy, of $83.96\%$  $ (0.26)$, which shows that one can get additional explanations without sacrificing label accuracy. 

For the generated explanations, the test perplexity is $10.58$ $ (0.40)$ and the BLEU (bilingual evaluation understudy) score \citep{BLEU} is $22.40$ $ (0.70)$. Since e-SNLI has three explanations for each instance in the validation and test sets, an inter-annotator BLEU score can be obtained, for example, by computing the BLEU score of the third explanation with respect to the first two explanations. In this way, I obtained an inter-annotator BLEU score of $22.51$. For consistency, I use the same two explanations as only references when computing the BLEU score for the generated explanations. 

Given the low inter-annotator score and the fact that the BLEU score of the generated explanations almost matches the inter-annotator BLEU score, we can see that this metric is not reliable for assessing the quality and correctness of the generated explanations. Therefore, I~manually annotated the first 100 instances in the test set (following the same partial scoring as in Section \ref{partial-score}). Since the explanation is conditioned on the predicted label, it is not expected that the model generates correct explanations when it predicts incorrect labels. %In fact, this would be undesirable if one wants faithful explanations. 
Therefore, an appropriate correctness measure for the explanations is the percentage of correct explanations among the subset for which the predicted label is correct. For the first 100 instances in the test set, the percentage of correct explanations for \expone{} is only $34.68\%$ (out of the 80 instances for which the model predicts correct labels). While this percentage is quite low, one should keep in mind that the selection criteria for \expone{} was only the label accuracy. In the next experiment, I show how selecting (and training) only for generating explanations results in a higher percentage of correct explanations when the predicted label is also correct..

\begin{figure}[t]
        \centering
        \includegraphics[width=0.8\linewidth]{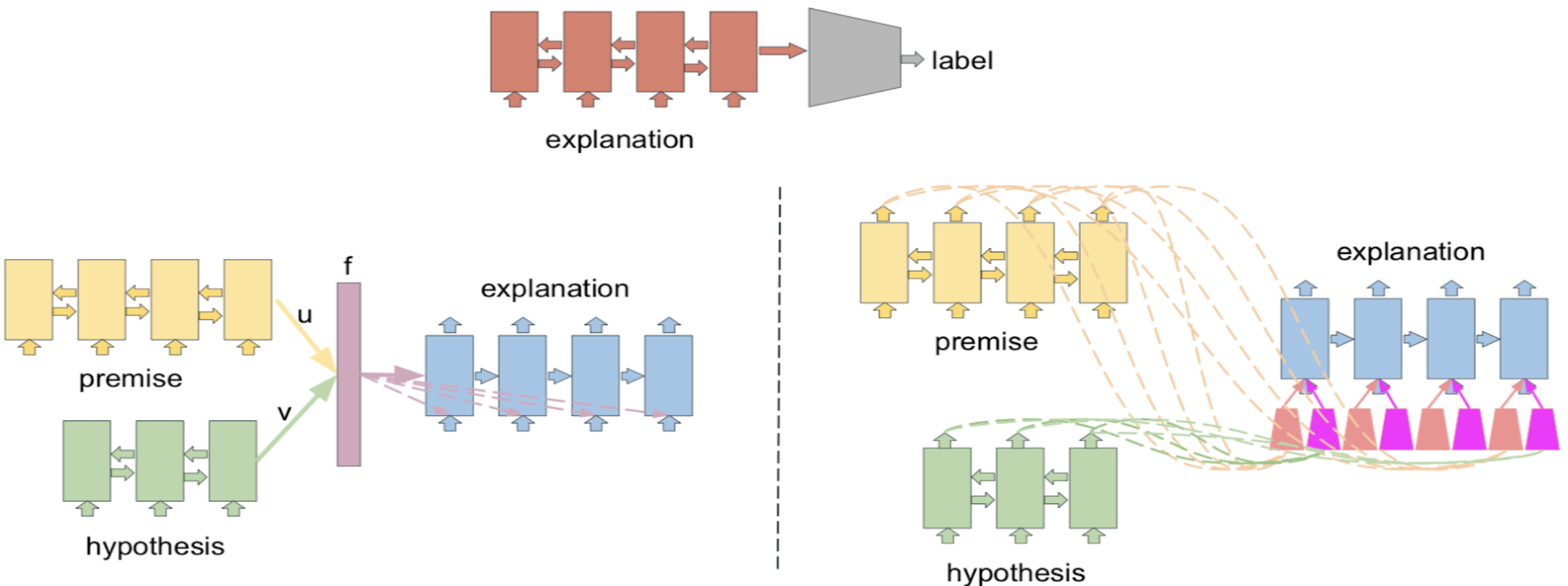}
        \caption[Architectures of the two versions of the \exptwo model]{Architectures of the two versions of \exptwo. At the top is the \jtl{} model that predicts a label from an explanation alone. On the left is the \exptwoseqtoseq{} model. On the right is the \exptwoattention{} model.}
        \label{fig:model2}
\end{figure}

\subsection{Explain then Predict} %\exptwo: First Generate an Explanation then Predict a Label}
\label{exptwo}

In the previous experiment, \expone{} generated an explanation conditioned on its predicted label. %because we wanted to see how the typical architecture used on SNLI can be adapted to justify its decisions in natural language.
However, this type of model might not provide explanations that are faithful to the decision-making process of the model. For instance, the explanation generator does not have access to the inner workings of the classifier (see Figure \ref{fig:model}).
%an arguably more natural approach for solving inference is to think of the explanation first and based on the explanation to decide a label. 
To increase the probability of obtaining faithful explanations, one can reverse the order between the label prediction and the explanation generation. Thus, I propose a model that first generates an explanation given a pair of (premise, hypothesis), and then predicts a label given only the generated explanation. I call this type of model \exptwo, and I will provide two versions of it below. This is a sensible decomposition for the e-SNLI dataset, due to the following key observation: for e-SNLI, one can easily detect for which label an explanation has been provided. This might not be the case in general, as the same explanation can correctly argue for different labels, depending on the premise and hypothesis. For example, ``A woman is a person.'' would be a correct explanation both for the entailment pair (``A woman is in the park.'', ``A person is in the park.'') and for the contradiction pair (``A woman is in the park.'', ``There is no person in the park.''). However, there are multiple ways of formulating an explanation. In our example, for the contradiction pair, one could also explain that ``There cannot be no person in the park if a woman is in the park.'', which read alone would allow one to infer that the pair was a contradiction. The latter kind of explanation is dominant in the e-SNLI dataset.
%To support this observation, 
% Moreoever, we looked at 100 random explanations where the classifier predicted the wrong label and noticed that xxx of them were actually incorrect explanations written by the workers and not 

\paragraph{Model.} For the part of the model \exptwo{} that predicts an explanation given a premise and a hypothesis, I propose two architectures. First, \exptwoseqtoseq{} is a simple sequence-to-sequence model. Essentially, it is the same as \expone{} but without the MLP classifier (and no label prepended to the explanation). Second, \exptwoattention{} is an attention version of this \exptwoseqtoseq. Attention mechanisms in neural networks brought consistent improvements over the non-attention counter-parts in various areas, such as computer vision \citep{showattend}, speech \citep{listenattend}, and natural language processing \citep{best-attention, attention}. Similarly to \exptwoseqtoseq, the encoders are two 2048-BiLSTM units and the decoder is a one-layer LSTM. The difference is that \exptwoattention{} uses two separate but identical attention heads to attend over the tokens from the premise and hypothesis while generating the explanation. 
Finally, I train a neural model, called \jtl, that given only an explanation predicts a label. \jtl{} consists of a 2048-BiLSTM with max-pulling encoder and a 3-layer MLP classifier, and obtains a test accuracy of $96.83\%$. This high test accuracy proves that it is reasonable to decompose this model into two separate modules. 
An illustration of the two versions of this model is depicted in Figure \ref{fig:model2}. %For details of the attention modules, see Appendix \ref{appendix-attention}.

\paragraph{Model selection.} The only hyperparameter in this experiment is the internal size for the decoder. The model selection criterion is the perplexity on the validation set of e-SNLI. The best perplexity for both \exptwoseqtoseq{} and \exptwoattention{} was obtained for an internal size of $1024$.

\paragraph{Results.} With the described setup, the label accuracy drops from $83.96\%$ $ (0.26)$ in \expone{} to $81.59\%$  $ (0.45)$ in \exptwoseqtoseq{} and $81.71\%$ $ (0.36)$ in \exptwoattention. 
However, manual annotation of the generated explanations for the first 100 instances in the test set, resulted in significantly higher percentages of correct explanations: $49.8\%$ for \exptwoseqtoseq{} and $64.27\%$ for \exptwoattention. Note that performing model selection only on the perplexity on the generated explanation, and using attention mechanisms significantly increases the quality of the explanations. %For the reference, the perplexity and BLEU score are $8.95 (0.03)$ and $24.14 (0.58)$ for \exptwoseqtoseq{}, and $6.1 (0)$ and $27.58 (0.47)$  for \exptwoattention. 
This experiment shows that, despite a small decrease in label accuracy, one can increase the correctness of the explanations generated for the instances for which a model predicts the correct label. 

The results of all models are summarised in Table \ref{summary}.

\begin{table}[]
  \caption[The performance of \expone{} and \exptwo.]{The performance of \expone, \exptwo, and of the \infersent{} baseline. The averages are over five seeds and the standard deviations are in parentheses. Expl@100 is the percentage of correct explanations in the subset of instances for which the model predicted the correct label in the first 100 instances from the e-SNLI test set. The correctness of the explanations was obtained after manual annotation. In bold are the best results.}
  \smallskip 
  \label{summary}
  \centering
  \resizebox{1.0\textwidth}{!}{
  \begin{tabular}{lllll}
    \toprule
    Model     & Label Accuracy & Perplexity & BLEU & Expl@100 \\
    \midrule
    \infersent{} & \textbf{84.01} (0.25) & - & - & - \\
    \expone{} & 83.96 (0.26) & 10.58 (0.40)  & 22.40 (0.70) & 34.68  \\
    \exptwoseqtoseq & 81.59 (0.45) & 8.95 (0.03) & 24.14 (0.58) & 49.80 \\
    \exptwoattention & 81.71 (0.36) & \textbf{6.10} (0.00)  & \textbf{27.58} (0.47) & \textbf{64.27} \\
    \bottomrule
  \end{tabular}
  }
\end{table}

\paragraph{Qualitative analysis.} 
In Table \ref{generated-expls}, one can see examples of explanations generated by (a) \expone, (b) \exptwoseqtoseq, and (c) \exptwoattention. At the end of each explanation, in brackets is the score that I manually allocated, as explained in Section \ref{partial-score}. Notice that the explanations are mainly on topic for all the three models, with minor exceptions, such as the mention of ``camouflage'' in (1c). Also notice that even when incorrect, the generated explanations are sometimes frustratingly close to being correct. For example, explanation (2b) is only one word (out of its 20 words) away from being correct. 

It is also interesting to inspect the explanations provided when the predicted label is incorrect. For example, in (1a), we see that the network omitted the information of ``facing the camera'' from the premise and therefore classified the pair as neutral, which is backed up by an otherwise correct explanation in itself. We also see that model \exptwoseqtoseq{} correctly classifies this pair as entailment. However, it only generates one out of the three necessary arguments that lead to entailment, and it also picks arguably the easiest argument. On the other hand, the attention model label the instance (1c) as neutral and argues for the subtle fact that ``standing'' and ``facing a camera'' are not enough to conclude ``posing for a picture'', while humans labelled this instance as entailment. However, the explanations also mentions ``camouflage'', which is not necessarily the same as hoodies. Therefore, I labelled this explanations as incorrect (score of 0 in the brackets).

While more research is needed to know whether the generated explanations provide reliable clues about the decision-making process of the model, it is reassuring to see that the explanations are plausible.

%While we agree with the SNLI annotation of this pair as entailment, and we score this explanation as wrong, off the record, we give some credit to this explanation, since there can be (less likely) reasons why people could stand and face a camera, other than taking a picture, for example, it could be a security camera.

%also misclassifies the pair as neutral but with an explanation that is also wrong in itself while pointing to the correct evidence. 

\begin{table}[]
\caption{Examples of predicted labels and generated explanations from (a)~\expone{}, (b)~\exptwoseqtoseq{}, and (c)~\exptwoattention. In brackets is the score of correctness, with partial scoring as described in Section \ref{partial-score}. }

\smallskip 
\label{generated-expls}
\resizebox{\textwidth}{!}{%
\begin{tabular}{lll}
\toprule
\iffalse
\multicolumn{3}{c}{\begin{tabular}[c]{@{}c@{}} (1) \textsc{Premise}:  A couple walk hand in hand down a street. \\ \textsc{Hypothesis}:  A couple is walking together. \\ \textsc{Gold label}:  entailment\end{tabular}} \\  

\\

\multicolumn{1}{l|}{\begin{tabular}[c]{@{}l@{}}(a) \textsc{Predicted label}:  entailment \\ \textsc{Explanation}:  A couple walking \\ down a street is a rephrasing \\ of walking together. [0] \end{tabular}}                                & \multicolumn{1}{l|}{\begin{tabular}[c]{@{}l@{}}(b) \textsc{Predicted label}:  entailment \\ \textsc{Explanation}:  A couple walking \\ down a street is a couple. [0] \end{tabular}}                                                                       & \begin{tabular}[c]{@{}l@{}}(c) \textsc{Predicted label}:  entailment \\ Generated explanation: A hand \\ walk hand in hand is a rephrasing \\ of walking together. [1] \end{tabular}  \\

\fi

\multicolumn{3}{c}{\begin{tabular}[c]{@{}c@{}} (1) \textsc{Premise}:  3 young man in hoods standing in the middle of a quiet street facing the camera. \\ \textsc{Hypothesis}:  Three hood wearing people pose for a picture. \\ \textsc{Gold label}:  entailment\end{tabular}} \\  

\\

\multicolumn{1}{l|}{\begin{tabular}[c]{@{}l@{}}(a) \textsc{Predicted label}:  neutral \\ \textsc{Explanation}:  Just because the men are in the \\ middle of a street doesn't mean \\ they are posing for a picture. [0] \end{tabular}}                                & \multicolumn{1}{l|}{\begin{tabular}[c]{@{}l@{}}(b) \textsc{Predicted label}:  entailment \\ \textsc{Explanation}:  three young men \\ are people. [0.33] \end{tabular}}                                                                       & \begin{tabular}[c]{@{}l@{}}(c) \textsc{Predicted label}:  neutral \\ \textsc{Explanation}:  Just because three young man \\ in camouflage standing in the middle of \\ a quiet street facing the camera \\ does not mean they pose for a picture. [0] \end{tabular}  \\

\\ \midrule

\multicolumn{3}{c}{\begin{tabular}[c]{@{}c@{}}(2) \textsc{Premise}:  Three firefighter come out of subway station.\\ \textsc{Hypothesis}:  Three firefighters putting out a fire inside of a subway station. \\ \textsc{Gold label}:  neutral\end{tabular}} \\                             

\\

\multicolumn{1}{l|}{\begin{tabular}[c]{@{}l@{}}(a) \textsc{Predicted label}:  contradiction \\ \textsc{Explanation}:  The firefighters \\ can not be putting out a fire station and \\ putting out a fire at the same time. [0] \end{tabular}} & \multicolumn{1}{l|}{\begin{tabular}[c]{@{}l@{}}(b) \textsc{Predicted label}:  neutral\\ \textsc{Explanation}:  The fact that three\\ firemen are putting out of a subway\\ station doesn't imply that they\\  are putting out a fire. [0]\end{tabular}} & \begin{tabular}[c]{@{}l@{}}(c) \textsc{Predicted label}:  neutral\\ \textsc{Explanation}:  The firefighters \\ may not be putting out a fire \\ inside of the subway station. [1]\end{tabular} \\

\\  \midrule

\multicolumn{3}{c}{\begin{tabular}[c]{@{}c@{}} (3) \textsc{Premise}:  A blond-haired doctor and her African American assistant looking threw new medical manuals.\\ \textsc{Hypothesis}:  A man is eating pb and j.\\ \textsc{Gold label}:  contradiction\end{tabular}}   \\                                                                                \\

\multicolumn{1}{l|}{\begin{tabular}[c]{@{}l@{}}(a) \textsc{Predicted label}:  contradiction\\ \textsc{Explanation}:  A man is not a woman. [1] \end{tabular}}                                                                                   & \multicolumn{1}{l|}{\begin{tabular}[c]{@{}l@{}}(b) \textsc{Predicted label}:  contradiction\\ \textsc{Explanation}:  One can not be looking \\ and eating simultaneously. [0] \end{tabular}}                                                               & \begin{tabular}[c]{@{}l@{}}(c) \textsc{Predicted label}:  contradiction\\ \textsc{Explanation}:  A person can not be looking \\ at a medical and a book at the same time. [0]
\end{tabular} 
\\
\bottomrule
\end{tabular}%
}
\end{table}

\subsection{Universal Sentence Representations}%\expthree: Universal Sentence Representations}
\label{expthree}
Obtaining universal sentence representations is the task of training an encoder to provide semantically meaningful fixed-length representations of phrases/sentences \citep{skip, fastsent, infersent}. 
These representations are further used as features in other downstream tasks, particularly for tasks where labelled training data are scarce. 
In computer vision, pretrained ImageNet-based encoders provide standard image feature extractors for other downstream tasks \citep{imagenet}. 
However, in natural language processing, there is still no consensus on general-purpose sentence encoders. It is an open question on which task and dataset should such an encoder be trained. Traditional approaches make use of very large unsupervised datasets, taking weeks to train \citep{skip}. Recently, \citet{infersent} showed that training only on natural language inference is both more accurate and more time-efficient than training on orders of magnitude larger but unsupervised datasets. Their results encourage the idea that more supervision can be more beneficial than larger but unsupervised datasets. Therefore, it is interesting to investigate whether an additional layer of supervision in the form of natural language explanations can further improve the learning of universal sentence representations.

\paragraph{Model.} I use the \expone{} model trained in Section \ref{expone}. To ensure that a potential improvement of \expone{} over \infersent{} comes from the explanations and not simply from the addition of a language decoder, I introduce the \inferSentAutoencoder{} model as an additional baseline. \inferSentAutoencoder{} follows the same architecture as \expone, but instead of decoding explanations, it decodes the premise and hypothesis from their corresponding vector representations using a shared LSTM decoder. For \inferSentAutoencoder, I use the same hypeparameters as for \expone.

\paragraph{Evaluation metrics.}
Typically, sentence representations are evaluated by using them as fixed features on top of which shallow classifiers are trained to perform a series of downstream tasks. \citet{infersent} provide an excellent tool for evaluating sentence representations on 10 diverse tasks: movie reviews ({MR}), product reviews ({CR}), subjectivity/objectivity ({SUBJ}), opinion polarity ({MPQA}), question-type ({TREC}), sentiment analysis ({SST}), semantic textual similarity ({STS}), paraphrase detection ({MRPC}), entailment ({SICK-E}), and semantic relatedness ({SICK-R}). MRPC is evaluated with accuracy/F1-score. STS14 is evaluated with the Pearson/Spearman correlations. SICK-R is evaluated with the Pearson correlation. For all the rest of the tasks, accuracy is used for performance evaluation.
A detailed description of each of these tasks and of the evaluation tool can be found in their paper.%, which we use for comparing the quality of the sentence embeddings obtained by additionally providing our explanations on top of the label supervision. 

\paragraph{Results.} Table \ref{senteval} shows the results (as averages over 5 models trained with different seeds, with the standard deviations in parentheses) of \expone, \infersent, and \inferSentAutoencoder{} on the 10 downstream tasks mentioned above. To test whether the differences in performance of \inferSentAutoencoder{} and \expone{} relative to the \infersent{} baseline are significant, I performed a Welch’s t-test.\footnote{Using the implementation in  \texttt{scipy.stats.ttest\_ind} with \texttt{equal\_var=False}.} The results that appeared significant under the significance level of $0.05$ are marked with `*'.

\begin{table}
  \caption{Performance of \expone{} and of the baselines, \infersent{} and \inferSentAutoencoder, on the 10 downstream tasks. Results are the averages of 5 runs with different seeds, with the standard deviations shown in parentheses. The best result for every task is indicated in bold. `*' indicates a significant difference at level $0.05$ with respect to the \infersent{} baseline.}
  
  \smallskip 
  \label{senteval}
  \centering
  \begin{adjustbox}{width=1\textwidth}
  \begin{tabular}{lllllllllll}
    \toprule
    %\multicolumn{2}{c}{Part}                   \\
    %\cmidrule(r){1-2}
    Model     & MR & CR & SUBJ & MPQA & SST2 & TREC & MRPC & SICK-E & SICK-R & STS14 \\
    \midrule
    \infersent{} & \textbf{78.18}  &	81.28 &	\textbf{92.46} &	88.46 &	\textbf{82.12} &	89.32 &	74.82 /	82.74 &	\textbf{85.96} &	0.887 &	0.65 / 0.63 \\
& (0.25) & (0.15) &	(0.15) &	(0.21) &	(0.22) &	(0.50) &	(0.66 /	0.27) &	(0.32) &	(0.002) &	(0.00 / 0.00) \\
    \inferSentAutoencoder &  75.94* &	79.26* &	91.72* &	88.16 &	80.90* &	\textbf{90.52*} &	\textbf{76.20*} /	82.48 &	85.58 &	0.880* &	0.50* / 0.50* \\
& (0.18) &	(0.36) &	(0.28) &	(0.26) &	(0.48) &	(0.52) &	(0.93 /	1.23) &	(0.33) &	(0.00) &	(0.02 / 0.02)\\
    \expone{} & 77.76	&  \textbf{81.3} &	92.14* &	\textbf{88.78*} &	81.84 &	90.00 &	75.56 /	\textbf{83.24*} &	85.92 &	\textbf{0.890*} &	\textbf{0.68} / \textbf{0.65*} \\
& (0.44) &	(0.16) &	(0.21) &	(0.22) &	(0.40) &	(0.51) &	(0.62 / 0.24) &	(0.52) &	(0.000) &	(0.01 / 0.01)\\
    \bottomrule
  \end{tabular}
  \end{adjustbox}
\end{table}

We see that \inferSentAutoencoder{} performs significantly worse than \infersent{} on six tasks and significantly outperforms it on only two tasks. This indicates that adding a language generator can actually hurt performance. Instead, \expone{} significantly outperforms \infersent{} on four tasks, while it is significantly outperformed only on one task. Therefore, one can conclude that training with explanations helps the model to learn overall better sentence representations.

\subsection{Performance on Out-of-Domain Datasets}%\expfour : Transfer without Fine-tuning to Out-of-domain NLI} 
\label{exp4}
%Improving the capabilities of a model to perform
It is known that models trained on one distribution of instances do not perform well on instances from another distribution.
%Transfer without fine-tuning to out-of-domain natural language inference datasets is known to be a challenging task. 
For example, \citet{snli} obtained an accuracy of only $46.7\%$ when training a model on SNLI and evaluating it on SICK-E \citep{sick}. Therefore, it is interesting to investigate whether explanations can help models to improve their performance on out-of-domain datasets, in both label prediction and explanation generation.
I investigate this using the SICK-E~\citep{sick} and MultiNLI~\citep{multinli} datasets for natural language inference. %SICK-E is a simple 
%MultiNLI includes a diverse range of genres of written and spoken English, as well as test sets for cross-genre transfer.

\paragraph{Model.} I use the \expone{} model trained in Section \ref{expone}. As baselines, I again use the already trained \infersent{} and \inferSentAutoencoder{} models. 

\paragraph{Results.} Table \ref{artifacts} shows the performance of \expone, \infersent, and \inferSentAutoencoder{} when evaluated without fine-tuning on SICK-E and MultiNLI. We see that the improvements obtained with \expone{} are very small. Therefore, we cannot conclude on the usefulness of explanations in improving the performance of this model on out-of-domain datasets. 

To also evaluate the capability of \expone{} to generate correct natural language explanations, I manually annotated the explanations generated by this model for the first 100 instances in test sets of SICK-E and MultiNLI. The percentage of correct explanations in the subset where the label was predicted correctly is $30.64\%$ for SICK-E and only $1.92\%$ for MultiNLI.
During annotation, I noticed that the explanations in SICK-E, even when wrong, were generally on-topic and valid statements, while the ones in MultiNLI were generally nonsense or off-topic. This is not surprising, since SICK-E is less complex and more similar to SNLI than MultiNLI is. 

%Therefore, transfer learning for generating explanations in out-of-domain natural language inference would constitute challenging future work.

\begin{table}%{0.5\textwidth}
%\begin{figure}[t]
  \caption{The performance without fine-tuning of \expone{} and the two baselines, \infersent{} and \inferSentAutoencoder, on SICK-E and MultiNLI. Results are average accuracies over 5 seeds, with standard deviations in parentheses. expl@100 shows the correctness of generated explanations via manual annotation. }
  \label{artifacts}
  \centering
  \resizebox{0.7\linewidth}{!}{
  \begin{tabular}{lll}
    \toprule
    Model     & SICK-E acc/expl@100 & MultiNLI acc/expl@100 \\
    \midrule
    \infersent & 53.27	(1.65) / - &	57.00 (0.41) / - \\
    
    \inferSentAutoencoder{} &  52.90 (1.77) / - & 55.38 (0.90) / - \\

    \expone{} & \textbf{53.54} (1.43) / 30.64	 & \textbf{57.16} (0.51) / 1.92  \\
    
    \bottomrule
  \end{tabular}
  }
%\end{figure}  
\end{table}

\section{Conclusions and Open Questions}
In this chapter, I introduced e-SNLI, a large dataset of ${\sim}570$K human-written natural language explanations for the influential task of natural language inference. I showed that it is much more difficult for a neural model that generates correct labels based on spurious correlations in SNLI to also generate correct explanations based on these correlations. This brings empirical evidence that models that generate correct explanations are more reliable than models that only predict correct labels. 

I implemented various models that generate natural language explanations for their label predictions and I quantified the capability of these models to generate correct explanations when they predict the correct label. 
I also investigated the usefulness of providing these explanations at training time for obtaining better universal sentence representations and for improving the performance of a model on out-of-domain datasets.

Thus, the work described in this chapter paves the way for future development of robust neural models that can learn from natural language explanations at training time as well as generate correct natural language explanations at test time. Moreover, the e-SNLI dataset can also be exploited for other goals. For example, similar to the evaluation performed for visual question answering in \citet{humanVQA}, the highlighted tokens in e-SNLI may provide a source of supervision and evaluation for attention models~\citep{RocktaschelGHKB15, snli-2}. 

It also remains an open question how to automatically evaluate the quality of the generated explanations, since we have seen that automatic measures, such as the BLEU score, are not suitable. 

It is also left as open question how to verify if the generated explanations are faithfully describing the decision-making process of the model.

%% file: inconsistencies.tex
\chapter{Verifying Neural Networks that Generate Natural Language Explanations}
\label{chap-inconsist}

In this chapter, which is based on \citep{inconsistencies}, I introduce a simple yet effective adversarial framework to verify if neural models can generate inconsistent natural language explanations.

%In Chapter \ref{chap-taxo}, we saw that a growing number of works propose enhancing neural models with the capability to generate natural language explanations that support their predictions. In the same direction, in Chapter \ref{chap-esnli}, I also train neural models to generate such explanations and quantify the correctness of the argumentation provided by the generated explanations. However, even when a generated explanation provides the correct argumentation, it is not guaranteed to faithfully describe the decision-making process of the model. As we saw in Section \ref{faithful}, faithfulness is an important property of explanations. In this chapter, which is based on Publication~\ref{publication-inconsist}, I introduce a simple yet effective adversarial framework to verify if neural models can generate inconsistent natural language explanations. I apply the framework to the \exptwoattention{} model introduced in the previous chapter, and I show that this model is capable of generating a significant number of inconsistent explanations. Moreover, as part of the framework, I address the problem of adversarial attacks with \emph{full} target sequences, a scenario that was not previously addressed in sequence-to-sequence attacks, and which can be useful for other tasks. %, such as dialog systems.

\section{Motivation}
In order to explain the predictions produced by black-box neural models, a growing number of works propose extending these models with natural language explanation generation modules, thus obtaining models that explain themselves in human language~\citep{DBLP:conf/eccv/HendricksARDSD16,zeynep,cars,world-tree,math,explainyourself}.

In this chapter, I first draw attention to the fact that such models, while appealing, are nonetheless prone to generating inconsistent explanations. Two explanations are considered to be inconsistent if they provide contradictory arguments about the instances and predictions that they aim to explain. For example, consider a visual question answering (VQA) task~\citep{zeynep} and two instances where the image is the same but the questions are different, say ``Is there an animal in the image?'' and ``Can you see a Husky in the image?''. If for the first instance a model predicts ``Yes.'' and generates the explanation ``Because there is a dog in the image.'', while for the second instance the \emph{same} model predicts ``No.'' and generates the explanation ``Because there is no dog in the image.'', then the model is producing a pair of inconsistent explanations. %, since it says both that there is and that there is not a dog in that image.
%
%We point out that even if the predictions are correct --- in our example, if there is a dog in the image but it is not a Husky --- 

Inconsistent explanations reveal at least one of the following undesired behaviours:
\begin{inparaenum}[(i)]%[\itshape (i\upshape)]
\item at least one of the explanations is not faithfully describing the decision-making process of the model, or
\item the model relied on a faulty decision-making process for at least one of the instances.
\end{inparaenum}
For a pair of inconsistent explanations, further investigation is needed to conclude which of these two behaviours is the actual one (and might vary for each instance). Indeed, a pair of inconsistent explanations does not necessarily imply at least one unfaithful explanation. In the previous example, if the image contains a dog, it is possible that the model identifies the dog when it processes the image together with the first question, and that the model does not identify the dog when it processes the image together with the second question, hence both explanations would faithfully reflect the decision-making process of the model even if they are inconsistent. Similarly, a pair of inconsistent explanations does not necessarily imply that the model relies on a faulty decision-making process, because the explanations may not faithfully describe the decision-making process of the model. 

Before investigating the problem of identifying which of the two undesired behaviours is true for a given pair of inconsistent explanations, it is sensible to first examine whether a model is capable of generating inconsistent explanations. To this goal, I introduce the \incf{} framework. %for checking if models are robust against generating inconsistent natural language explanations. More precisely, 
For a given model $m$ and an instance $\bfx$, \incf{} aims to generate at least one input $\bfr$ that causes $m$ to generate an explanation that is inconsistent with the explanation generated by $m$ for $\bfx$. Thus, \incf{} falls under the category of \emph{adversarial methods}, i.e., methods searching for inputs that cause a model to produce undesired answers~\citep{DBLP:conf/pkdd/BiggioCMNSLGR13, adv-img1}. 

As part of \incf, I address the problem of adversarial attacks with exact target sequences, a scenario that has not been previously addressed in sequence-to-sequence attacks, and which can be useful for other areas, such as dialog systems.
Finally, I apply the framework to \exptwoattention{} presented in the previous chapter, and show that this model can generate a significant number of inconsistent explanations.

\section{Related Work}\label{related}
\paragraph{Verifying models that generate natural language explanations.}
Works on verifying models that generate natural language explanations are very scarce. \citet{grounding} identify the risk of generating natural language explanations that mention attributes from a strong class prior without any evidence being present in the input. 
In this chapter, I bring awareness to another risk, which is of generating inconsistent explanations.
%
%Similarly, \citet{grounding} identify the risk of mentioning attributes from a strong class prior without any evidence being present in the input. 
%
%In NLP, the problem of explaining the behaviour of a model is either via black-box analysis, or via modifications of the model (\emph{white-box} analysis).
%
%Black-box methods generate explanations by analysing the model behaviour in different regions of the input space~\citet{DBLP:conf/kdd/Ribeiro0G16,DBLP:conf/naacl/Nguyen18,DBLP:conf/aaai/Ribeiro0G18}, or via token sensitivity analysis~\citet{DBLP:journals/corr/LiMJ16a,DBLP:conf/emnlp/FengWGIRB18}.
%
%White-box analysis methods augment models to generate explanations jointly with predictions~\citet{esnli,DBLP:conf/emnlp/LeiBJ16}, the magnitude of inner activations~\citet{DBLP:journals/corr/abs-1812-08718}, gradients~\citet{DBLP:journals/corr/LiMJ16a}, and attention weights~\citet{DBLP:journals/corr/BahdanauCB14}.
%
%However, at the time of this writing, no method in the literature focuses on the problem of assessing the quality of produced explanations, and on identifying inputs yielding to inconsistent natural language explanations.
%
%
\paragraph{Generating adversarial examples.}
Generating adversarial examples is an active research area in natural language processing~\citep{asurvey,DBLP:journals/corr/abs-1902-07285}. For instance, \citet{DBLP:conf/emnlp/JiaL17} analyse the robustness of extractive question answering models on examples obtained by adding adversarially generated distracting text.
%
%Other works analyse sensitivity to small random character perturbations~\citet{DBLP:journals/corr/abs-1711-02173,DBLP:conf/cvpr/HosseiniXP17}, paraphrasing~\citet{DBLP:journals/corr/abs-1804-06059}, and simple transformations requiring lexical and world knowledge~\citet{breaking-nli}.
%
However, most works build on the requirement that the adversarial input should be a small perturbation of an original input~\citep{DBLP:journals/corr/abs-1711-02173,DBLP:conf/cvpr/HosseiniXP17,keywords}, or should be preserving the semantics of the original input~\citep{DBLP:journals/corr/abs-1804-06059}. %, but leading to a different prediction.
%While this is necessary for testing the robustness of a model,
Our setup does not have this requirement, and any pair of task-realistic inputs that causes the model to produce inconsistent explanations suffices.
%
%Also, existing adversarial models do not always require the adversarial input to be grammatically correct, and often they can change words or characters to completely random ones~\citep{keywords}.
%
%This is acceptable for certain cases, such as summarisation of long pieces of text, where changing a few words would likely not change the main flow of the text.
%
%
%%%In cases like ours, where the inputs are short sentences, and the model is tested for fine-grained reasoning, it is desirable that the adversarial examples are gra%mmatically correct.
%
Most importantly, to my knowledge, no previous adversarial attack for sequence-to-sequence models generates \emph{exact target sequences}, i.e., given a sequence, find an input that causes the model to generate the exact given sequence.
Closest to this goal, \citet{DBLP:conf/iclr/ZhaoDS18} propose an adversarial framework for removing or adding tokens in the target sequence for the task of machine translation. 
Similarly, \citet{keywords} require the presence of pre-defined tokens anywhere in the target sequence. They only test with up to three required tokens, and their success rate dramatically drops from $99\%$ for one token to $37\%$ for three tokens for the task of automatic summarisation. Hence, their method would likely not generalise to exact target sequences.

Finally, \citet{DBLP:conf/conll/Minervini018} attempted to find inputs where a model trained on SNLI \citep{snli} violates a set of logical constraints. This scenario may, in theory, lead to finding inputs that cause the generation of inconsistent explanations. 
However, their method needs to enumerate and evaluate a potentially very large set of perturbations of the inputs, \eg{} removing sub-trees or replacing tokens with their synonyms, thus being computational expensive.
%Besides the computational overhead, it also may easily generate ungrammatical inputs. 
Moreover, their scenario does not address the question of automatically producing undesired (inconsistent) sequences.

\section{Framework}
Consider a model $m$ that, for each instance $\bfx$, generates a natural language explanation for its prediction on the instance. We refer to the explanation generated by $m$ for $\bfx$ as $\bfe_{m}(\bfx)$.
I propose a framework that, for an instance $\bfx$, aims to generate new instances for which the model produces explanations that are inconsistent with $\bfe_{m}(\bfx)$. 

\incf{} consists of the following high-level steps. Given an instance $\bfx$, 
\begin{inparaenum}[(A)]%[\itshape (A\upshape)]
\item create a list of explanations that are inconsistent with $\bfe_{m}(\bfx)$, and \label{A}
\item given an inconsistent explanation from the list created in \ref{A}, find an input that causes $m$ to generate this precise inconsistent explanation. \label{B}
\end{inparaenum}

\paragraph{Setup.} The setup we are facing has three desired properties that make it different from commonly researched adversarial settings in natural language processing:
\begin{itemize}

\item At step (\ref{B}), the model has to generate a exact target sequence, since the goal is to generate the \emph{exact} explanation that was identified at step (\ref{A}) as inconsistent with the explanation $\bfe_{m}(\bfx)$.

\item Adversarial inputs do not have to be a paraphrase or a small perturbation of the original input, since the objective is to generate inconsistent explanations rather than incorrect predictions --- these can happen as a byproduct.
\item Adversarial inputs have to be realistic to the task at hand. % and relevant --- in previous works, this requirement never appears jointly with the aforementioned two properties.
\end{itemize}

To my knowledge, this work is the first to tackle this setup, especially due to the challenging requirement of generating a exact target sequence, as described in \Cref{related}. % for comparison with existing works.

\paragraph{Context-dependent inconsistencies.} In certain tasks, instances consist of a context (such as an image or a paragraph), and some assessment to be made about the context (such as a question or a hypothesis). Since explanations may refer (sometimes implicitly) to the context, the assessment of whether two explanations are inconsistent may also depend on it. For example, in VQA, the inconsistency of the two explanations ``Because there is a dog in the image.'' and ``Because there is no dog in the image.'' depends on the image. However, if the image is the same, the two explanations are inconsistent regardless of which questions were asked on that image. 

For the above reasons, given an instance $\bfx$, I differentiate between parts of the instance that will remain fixed in \incf{} (referred to as \emph{context parts} and denoted as $\bfxd$) and parts of the instance that \incf{} will vary in order to obtain inconsistencies (referred to as \emph{variable parts} and denoted as $\bfxi$). Hence, $\bfx = (\bfxd, \bfxi)$. In the VQA example above, $\bfxd$ would be the image, and $\bfxi$ would be the question.
%
%
%In our visual question answering example, $\bfxd$ is the image and $\bfxi$ is the question.~\footnote{One may imagine datasets in which the questions are expected to bring additional information such as ``If we knew that [...], [question].'', and such cases need to be treated accordingly.}
%

\paragraph{Stand-alone inconsistencies.}
There are cases for which explanations are inconsistent regardless of the input. For example, explanations formed purely of background knowledge such as ``A woman is a person.'' and ``A woman is not a person.''\footnote{Which was generated by the model in our experiments.} are always inconsistent (and sometimes outrageous), regardless of the instances that lead to them. 
For these cases, \incf{} can treat the whole input as variable, i.e., $\bfxd = \emptyset$ and $\bfri = \bfx$.

\paragraph{Steps.}\label{steps}
\incf{} consists of the following steps:
\begin{enumerate}%[wide, labelwidth=!, labelindent=0pt]
    \item Reverse the explanation generator module of model $m$ by training a \rvj{} model to map from the generated explanation and the context part of the input to the variable part of the input, \ie{} $\rvj(\bfxd, \bfe_{m}(\bfx)) = \bfxi$.
    \item \label{s2} For each explanation $\bfe = \bfe_m(\bfx)$:
    \begin{enumerate}
        \item \label{rules} Create a list of statements that are inconsistent with $\bfe$, referred as $\mathcal{I}_{\bfe}$. 
        \item \label{b} Query \rvj~on each $\bfi \in \mathcal{I}_{\bfe}$ and the context $\bfxd$. Get the new variable part $\bfri = \rvj(\bfxd, \bfi)$ of a \emph{reverse input} $\bfr = (\bfxd, \bfri)$, which \emph{may} cause $m$ to produce inconsistent explanations.
        \item Query $m$ on each reverse input to get a \emph{reverse explanation} $\bfe_{m}(\bfr)$.
        \item \label{check_rules} Check if each reverse explanation $\bfe_{m}(\bfr)$ is indeed inconsistent with $\bfe$ by checking if $\bfe_{m}(\bfr) \in \mathcal{I}_{\bfe}$. % (with an exact string match)
    \end{enumerate}
    
\end{enumerate}
To execute step (\ref{rules}), for any task, one may define a set of logical rules to transform an explanation into an inconsistent counterpart, such as negation or replacement of task-essential tokens with task-specific antonyms. For example, in explanations for self-driving cars~\citep{cars}, one can replace ``green light'' with ``red light'', or ``the road is empty'' with ``the road is crowded'' (which are task-specific antonyms), to get inconsistent --- and hazardous --- explanations such as ``The car accelerates because there is a red light.''.
Another strategy to obtain inconsistent explanations consists of swapping explanations from mutually exclusive labels. For example, assume a recommender system predicts that movie X is a bad recommendation for user Y ``because X is a horror movie.'', implying that user Y does not like horror movies. If the same system also predicts that movie Z is a good recommendation to the same user Y ``because Z is a horror movie.'', then we have an inconsistency, as the latter would imply that user Y likes horror movies.

While this step requires a degree of specific adjustment to the task at hand, it is arguably a small price to pay to ensure that the deployed system is coherent. An interesting direction for future work would be automatise this step, for example, by training a neural network to generate task-specific inconsistencies after crowd-sourcing a dataset of inconsistent explanations for the task at hand. % as future work.
To execute step (\ref{check_rules}), \incf{} currently checks for an exact string match between a reverse explanation and any of the inconsistent explanations created at step (\ref{rules}). Alternatively, one can train a model to identify if a pair of explanations forms an inconsistency. %, which would also be an interes as future work.

Finally, while the framework itself does not directly enforce the adversarials to be realistic for the task at hand, the fact that they are generated by a model that is trained on realistic data induces the majority of the adversarials to be realistic, as we will see in the experiments below.

\begin{table*}[t]
\caption{Examples of inconsistent explanations detected by \incf{} on the model \exptwoattention.} % The reverse hypotheses (right) are realistic.} % and relevant to the premises.} 
\resizebox{0.99\textwidth}{!}{%
\begin{tabular}{p{12cm}|p{12cm}}

\toprule

\multicolumn{2}{c}{\begin{tabular}[c]{@{}c@{}} \textsc{Premise:} A guy in a red jacket is snowboarding in midair.\end{tabular}}

\\

\begin{tabular}[c]{@{}l@{}}
\textsc{Original Hypothesis:} A guy is outside in the snow. \\
\textsc{Predicted Label:} entailment \\
\textsc{Original explanation:} {\bf Snowboarding is done outside.}\end{tabular} &

\begin{tabular}[c]{@{}l@{}}
\textsc{Reverse Hypothesis:} The guy is outside. \\
\textsc{Predicted label:} contradiction \\
\textsc{Reverse explanation:} {\bf Snowboarding is not done outside.} \end{tabular}

\\

\midrule

\multicolumn{2}{c}{\begin{tabular}[c]{@{}c@{}} \textsc{Premise:} A man talks to two guards as he holds a drink.\end{tabular}} 

\\

\begin{tabular}[c]{@{}l@{}}
\textsc{Original Hypothesis:} The prisoner is talking to two guards in \\ the prison cafeteria. \\
\textsc{Predicted Label:} neutral \\
\textsc{Original explanation:} {\bf The man is not necessarily a} \\ {\bf prisoner.} \end{tabular}
& 
\begin{tabular}[c]{@{}l@{}}
\textsc{Reverse Hypothesis:} A prisoner talks to two guards. \\
\textsc{Predicted Label:} entailment \\
\textsc{Reverse explanation:} {\bf A man is a prisoner.}
\\
\end{tabular} 

\\

\midrule

\multicolumn{2}{c}{\begin{tabular}[c]{@{}c@{}}
\textsc{Premise:} Two women and a man are sitting down eating and drinking various items.\end{tabular}}

\\

\begin{tabular}[c]{@{}l@{}}
\textsc{Original Hypothesis:} Three women are shopping at the mall. \\
\textsc{Predicted label:} contradiction \\
\textsc{Original explanation:} \textbf{There are either two women and } \\ \textbf{a man or three women.} \\
\end{tabular}                                
& 
\begin{tabular}[c]{@{}l@{}}
\textsc{Reverse Hypothesis:} Three women are sitting down eating. \\
\textsc{Predicted label:} neutral \\
\textsc{Reverse explanation:} \textbf{Two women and a man are three} \\ {\bf women.} \\
\end{tabular}

\\

\midrule

\multicolumn{2}{c}{\begin{tabular}[c]{@{}c@{}} \textsc{Premise:} Biker riding through the forest.\end{tabular}}

\\

\begin{tabular}[c]{@{}l@{}}
\textsc{Original Hypothesis:} Man riding motorcycle on highway. \\
\textsc{Predicted Label:} contradiction \\
\textsc{Original explanation:} {\bf Biker and man are different.}\end{tabular} &

\begin{tabular}[c]{@{}l@{}}
\textsc{Reverse Hypothesis:} A man rides his bike through the forest. \\
\textsc{Predicted label:} entailment \\
\textsc{Reverse explanation:} {\bf A biker is a man.} \end{tabular}

\\

\midrule

\multicolumn{2}{c}{\begin{tabular}[c]{@{}c@{}} \textsc{Premise:} A hockey player in helmet.\end{tabular}} 

\\

\begin{tabular}[c]{@{}l@{}}
\textsc{Original Hypothesis:} They are playing hockey \\
\textsc{Predicted Label:} entailment \\
\textsc{Original explanation:} {\bf A hockey player in helmet is} \\ {\bf playing hockey.} \end{tabular}
& 
\begin{tabular}[c]{@{}l@{}}
\textsc{Reverse Hypothesis:} A man is playing hockey. \\
\textsc{Predicted Label:} neutral \\
\textsc{Reverse explanation:} {\bf A hockey player in helmet doesn't} \\ \bf{imply playing hockey.} \end{tabular} 

\\

\midrule

\multicolumn{2}{c}{\begin{tabular}[c]{@{}c@{}}
\textsc{Premise:} A blond woman speaks with a group of young dark-haired female students carrying pieces of paper. \end{tabular}}

\\

\begin{tabular}[c]{@{}l@{}}
\textsc{Original Hypothesis:} A blond speaks with a group of young \\ dark-haired woman students carrying pieces of paper. \\
\textsc{Predicted label:} entailment \\
\textsc{Original explanation:} \textbf{A woman is a female.}\\
\end{tabular}                                
& 
\begin{tabular}[c]{@{}l@{}}
\textsc{Reverse Hypothesis:}The students are all female. \\
\textsc{Predicted label:} neutral \\
\textsc{Reverse explanation:} \textbf{The woman is not necessarily } \\ {\bf female.} \end{tabular}

\\
\midrule

\multicolumn{2}{c}{\begin{tabular}[c]{@{}c@{}}
\textsc{Premise:} The sun breaks through the trees as a child rides a swing. \end{tabular}}

\\

\begin{tabular}[c]{@{}l@{}}
\textsc{Original Hypothesis:} A child rides a swing in the daytime. \\
\textsc{Predicted label:} entailment \\
\textsc{Original explanation:} \textbf{The sun is in the daytime.} \\
\end{tabular}                                
& 
\begin{tabular}[c]{@{}l@{}}
\textsc{Reverse Hypothesis:} The sun is in the daytime. \\
\textsc{Predicted label:} neutral \\
\textsc{Reverse explanation:} \textbf{The sun is not necessarily in the } \\ {\bf daytime.} \end{tabular}

\\

\midrule

\multicolumn{2}{c}{\begin{tabular}[c]{@{}c@{}}
\textsc{Premise:} A family walking with a soldier. \end{tabular}}

\\

\begin{tabular}[c]{@{}l@{}}
\textsc{Original Hypothesis:} A group of people strolling. \\
\textsc{Predicted label:} entailment \\
\textsc{Original explanation:} \textbf{A family is a group of people.} \\
\end{tabular}                                
& 
\begin{tabular}[c]{@{}l@{}}
\textsc{Reverse Hypothesis:} A group of people walking down a street. \\
\textsc{Predicted label:} contradiction \\
\textsc{Reverse explanation:} \textbf{A family is not a group of people.} \end{tabular}
\\

\bottomrule

\end{tabular}%
}
\label{inconsist-examples}
\end{table*}

\section{Experiments}
\label{experiments}
I test \incf{} on the model \exptwoattention{} introduced in the previous chapter.
%
%LATER: We highlight that our final goal is not a label attack, even if, for this particular model in which the label is predicted solely from the explanation, we implicitly also have a label attack with high probability.\footnote{Their Explanation-to-Label component had $96.83\%$ test accuracy.}
%
I set $\bfxd$ as the premise (as this represents the given context for the task of solving natural language inference) and $\bfxi$ as the hypothesis. However, due to the nature of SNLI for which decisions are based mostly on commonsense knowledge, the explanations are most of the time independent of the premise, such as ``A dog is an animal.''. Hence, it would be possible to also reverse the premise and not just the hypothesis, which is left as future work. %For example, \rvj~can work in two steps by first generating a reverse premise conditioned on the explanation, and then generating a reverse hypothesis conditioned on the explanation and the reverse premise. This would result in a higher search space yet potentially at the cost of a decrease of quality of instances since errors in the reverse premise may propagate into the reverse hypothesis; we leave this as future work.
For the \rvj~model, I use the same neural architecture and hyperparameters used for \exptwoattention. Thus, \rvj~takes as input a premise-explanation pair and generates a hypothesis. The trained \rvj~model is able to reconstruct \emph{exactly the same} (i.e, string matching) hypothesis with $32.78\%$ test accuracy. 

\paragraph{Creating $\mathcal{I}_{\bfe}$.}
To execute step (\ref{rules}), I employ negation and swapping explanations.
For negation, I simply remove the tokens ``not'' and ``n't'' if they are present. If these tokens appear more than once in an explanation, I create multiple inconsistencies by removing only one occurrence at a time. I do not attempt to add negation tokens, as this may result in grammatically incorrect sentences.

For swapping explanations on e-SNLI, we note that the explanations in this dataset largely follow a set of label-specific templates. This is a natural consequence of the task and the SNLI dataset and not a requirement in the collection of the e-SNLI. For example, annotators often used ``One cannot X and Y simultaneously.'' to explain a contradiction, ``Just because X, doesn't mean Y.'' for neutral, or ``X implies Y.'' for entailment.
Since any two labels are mutually exclusive, transforming an explanation from one template to a template of another label should automatically create an inconsistency.
For example, for the explanation of the contradiction ``One cannot eat and sleep simultaneously.'', one matches X to ``eat'' and Y to ``sleep'', and creates the inconsistent explanation ``Eat implies sleep.'' using the entailment template ``X implies Y.''. Thus, for each label, I created a list of the most used templates that I manually identified for the explanations e-SNLI, which can be found in Appendix \ref{assec:templates}. A running example of creating inconsistent explanations by swapping is given in Appendix \ref{run-ex}. 
If there is no negation and no template match, the instance is discarded. In our experiments, only $2.6\%$ of the e-SNLI test set were discarded for this reason. This procedure may result in grammatically or semantically incorrect inconsistent explanations. However, as we will see below, \rvj~performed well in generating correct and relevant reverse hypotheses even when its input explanations were not correct. This is not surprising, since \rvj~has been trained to output ground-truth hypotheses. %We manually annotated $100$ random reversed hypotheses generated by the \rvj~and found $81\%$ to be grammatically and semantically valid sentences which, together with the original premise, form realistic and interesting NLI instances.

The rest of the steps follow as described in (\ref{b}) - (\ref{check_rules}). More precisely, for each inconsistent explanation in $\mathcal{I}_{\bfe}$, \rvj{} returns a reverse hypothesis, which is fed back to \exptwoattention{} to get a reverse explanation. To check whether a reverse explanation is inconsistent with the initial explanation, I checked for an exact string match with the explanations from the list of inconsistent explanations from step (\ref{rules}). %Note that it is likely that a large number of inconsistent explanations are discarded due to insignificant syntactic differences. %However, when an exact match is found, i.e., a \textit{potential inconsistency}, it is very likely to be a \textit{true inconsistency}. We manually annotated a random sample of $100$ pairs of potential inconsistencies and found $85\%$ to be true inconsistencies. 

\paragraph{Results and discussion.}
\incf{} identifies a total of $1044$ pairs of inconsistent explanations starting from the e-SNLI test set, which contains of $9824$ instances. Nonetheless, there are, on average, $1.93 \pm 1.77$ distinct reverse hypotheses giving rise to the same pair of inconsistent explanation. Since the hypotheses are distinct, each of these instances is a separate valid adversarial input. However, if one is strictly interested in the number of distinct pairs of inconsistent explanations, then, after eliminating duplications, $540$ pairs of distinct inconsistencies remain.

Since the generation of natural language is always best evaluated by humans, I manually annotated $100$ random distinct pairs. I found that $82\%$ of the reverse hypotheses form realistic instances together with the premise. I also found that the majority of the unrealistic instances are due to a repetition of a token in the hypothesis. For example, ``A kid is riding a helmet with a helmet on training.'' is a generated reverse hypothesis which is very close to a perfectly valid hypothesis. %This kind of error may eventually be overcome with better training.

Given the estimation of $82\%$ of the detected inconsistencies to be caused by realistic reverse hypotheses, $\sim\!\!443$ distinct pairs of inconsistent explanations are detected by \incf. While this means that \incf{} only has a success rate of $\sim\!\!4.51\%$ on \exptwoattention, it is nonetheless alarming that this simple and under-optimised adversarial framework detects a significant number of inconsistencies on a model trained on $\sim\!\!570$K examples. In \Cref{inconsist-examples}, we see examples of detected inconsistencies. Note how the reverse hypotheses are realistic given the associated premises. %More examples can be found in Appendix \ref{more-examples-inconsist}. 

\paragraph{Manual scanning.} It is interesting to investigate to what extent one can find inconsistencies via brute-force manual scanning. I performed three such experiments, with no success. On the contrary, I noticed a good level of robustness against inconsistencies when scanning through the generic adversarial hypotheses introduced by \citet{behaviour}. %The details are in \Cref{man-scan}.

In the first experiment, I manually analysed the first $50$ instances in the test set without finding any inconsistency. However, these examples were involving different concepts, thus decreasing the likelihood of finding inconsistencies. To account for this, in the second experiment, I constructed three groups around the concepts of \emph{woman}, \emph{prisoner}, and \emph{snowboarding}, respectively, by simply selecting the explanations in the test set containing these words. 
I selected these concepts, because \incf{} detected inconsistencies about them (see Table \ref{inconsist-examples}).

For \emph{woman}, there were $1150$ instances in the test set on which \exptwoattention{} generated explanations containing this word. I looked at a random sample of $20$ explanations, among which I did not find any inconsistency.
For \emph{snowboarding}, there were $16$ instances in the test set for which \exptwoattention{} generated explanations containing this word, and no inconsistency among them.
For \emph{prisoner}, there was only one instance in the test set for which \exptwoattention{} generated an explanation containing this word, so there was no way to find out that the model is inconsistent with respect to this concept simply by scanning the test set.

I only looked at the test set for a fair comparison with \incf{} that was only applied on this set.

Even if the manual scanning were successful, it should not be regarded as a proper baseline, since it does not bring the same benefits as \incf. Indeed, manual scanning requires considerable human effort to look over a large set of explanations in order to find if any two are inconsistent. Even a group of only $50$ explanations required a non-negligible amount of time. Moreover, restricting ourselves to the instances in the original dataset would clearly be less effective than being able to generate new instances. The \incf{} framework addresses these issues and directly provides pairs of inconsistent explanations. Nonetheless, this experiment is useful for illustrating that \exptwoattention{} does not provide inconsistent explanations in a frequent manner. 

In the third experiment of manual scanning, I experimented with a few manually created hypotheses from the dataset introduced by \citet{behaviour}. These hypothesis had been shown to induce confusion at the label level. However, \exptwoattention{} showed a good level of robustness against inconsistencies on the inspected hypotheses. For example, for the neutral pair (premise: ``A bird is above water.'', hypothesis: ``A swan is above water.''), \exptwoattention{} generates the explanation ``Not all birds are a swan.'', while when interchanging ``bird'' with ``swan'', i.e., for the pair (premise: ``A swan is above water.'', hypothesis: ``A bird is above water.''), \exptwoattention~generates the explanation ``A swan is a bird.'', showing a good understanding of the relationship between the entities ``swan'' and ``bird''.
Similarly, interchanging ``child'' with ``toddler'' in (premise: ``A small child watches the outside world through a window.'', hypothesis: ``A small toddler watches the outside world through a window.'') does not confuse the model, which generates ``Not every child is a toddler.'' and ``A toddler is a small child.'', respectively. Further investigation on whether the model can be tricked on concepts where it seems to exhibit robustness, such as \emph{toddler} or \emph{swan}, is left for future work.

\section{Conclusions and Open Questions}
In this chapter, I drew attention to the fact that models generating natural language explanations are prone to producing inconsistent explanations. %, which can undermine users' trust in the model. 
This concern is general and can have a large practical impact. For example, users would likely not accept a self-driving car if its explanation module is prone to state both that ``The car accelerates because there is no one crossing the intersection.'' and that ``The car accelerates because there are people crossing the intersection.''.

I introduced a generic framework for identifying such inconsistencies and showed that \exptwoattention, the model that produced the highest parentage of correct explanations on e-SNLI in Chapter \ref{chap-esnli}, can generate a significant number of inconsistencies.

Valuable directions for future work include: (1) developing more advanced frameworks for detecting inconsistencies, (2) investigating whether inconsistencies are due to unfaithful explanations or to a faulty decision-making process of the model, and (3) developing more robust models that do not generate inconsistencies.

%% file: conclusions.tex
\chapter{General Conclusions and Perspectives}
\label{chap-concl}

In this thesis, I investigated two major directions for explaining deep neural networks: feature-based post-hoc explanatory methods and self-explanatory neural models that generate natural language explanations for their predictions. For both directions, I investigated the question of verifying the faithfulness with which the explanations describe the decision-making processes of the models that they aim to explain. In addition, for the class of self-explanatory models that generate natural language explanations, I investigated whether these models exhibit an improved behaviour due to the additional supervision in the form of natural language explanations at training time. The results are as follows.

First, I showed that, despite the apparent implicit assumption that there is only one ground-truth feature-based explanation for the prediction of a model on an instance, 
there are often more such ground-truths. Moreover, I showed that two important classes of post-hoc explanatory methods aim to provide different types of ground-truth feature-based explanations, without explicitly stating so, and I unveil certain strengths and limitations for each of these types. Furthermore, for certain cases, none of these types of explanations is enough to provide a non-ambiguous view of the decision-making process of a model.
I also showed that the selector-predictor type of neural model, which is expected to provide the relevant features for each prediction, has certain limitations in doing so, and I provided solutions that can alleviate some of these limitations.
%These findings have been presented in Chapter \ref{chap-difficulty}. 

Future work on rigorous specifications of explanatory methods would be necessary for proper usage of these methods in practice. More investigation into how explanations can provide a complete view of the decision-making process of a model would also be valuable. Moreover, robustifying selector-predictor models to provide faithful insight into their decision-making processes would also be an important direction of research.

Second, I introduced a framework for verifying the faithfulness of the explanations provided by feature-based post-hoc explanatory methods. This framework relies on the use of selector-predictor models as target models, after the limitations of this type of model are alleviated. 
This framework is automatic and verifies explainers on a realistic scenario, i.e., on non-trivial neural models that are trained on real-world datasets. 
The framework is also generic and has, therefore, the advantage to generate a potentially very large number of off-the-shelf sanity tests by being instantiated on other tasks and other architectures of the selector and the predictor modules. 
The results of testing three popular explanatory methods on this framework showed its potential to reveal unfaithful explanations provided by these methods.
I also presented ways for further improving this framework.

%Several open questions on verifying the faithfulness of feature-based post-hoc explanatory still remain. 
As future work, the introduced verification framework can further be improved, for example, using the guidelines that I introduced in Section \ref{sec:improvements}. Also, other types of feature-based self-explanatory methods might also be exploited as a testbed for verifying explainers. Moreover, recall that the introduced framework is a sanity check and not an evaluation framework for providing the performance of explainers in full generality. Complete evaluation frameworks would also be highly desired.

Third, I investigated the class of self-explanatory neural models that generate natural language explanations for their predictions. To address the scarcity of datasets of natural language explanations, I introduced a new and large dataset of ${\sim}570$K human-written natural language explanations, which I called e-SNLI. Further, I showed that it is more difficult for neural models to rely on spurious correlations to provide correct explanations than to provide correct labels. This empirically supports the intuition that models that can generate correct argumentation in addition to correct predictions are more reliable than models that just provide correct predictions. Furthermore, I showed that a series of models that generate natural language explanations at test time provide relatively low performance in generating correct explanations. However, I also showed that improvements are possible with better architectures. Finally, I showed that the presence of additional explanations at training time guide models into learning better universal sentence representations and into having better capabilities to solve out-of-domain instances, even though the improvements were minor for the tested models. %The results have been presented in Chapter \ref{chap-esnli}. 

Future work would further exploit the e-SNLI dataset to advance research in the direction of training with and generation of natural language explanations. The models introduced in this thesis, while providing encouraging results, are not yet deployable in the real world, and more work is needed to advance this promising direction. New datasets of natural language explanations would also be highly beneficial.

Fourth, as a step towards verifying the faithfulness of the generated natural language explanations, I introduced an adversarial framework to check whether neural models generate inconsistent natural language explanations. Inconsistent explanations expose either an unfaithful explanation or a faulty decision-making process of the model, either of which is undesirable. 
I applied the framework to the model that produced the highest percentage of correct explanations on e-SNLI and I showed that this model is capable of generating a significant number of inconsistent explanations. Moreover, as part of the introduced framework, I addressed the problem of adversarial attacks with full target sequences, a scenario that has not been previously investigated in sequence-to-sequence attacks, and which can be useful for other natural language tasks. %The results have been presented in Chapter \ref{chap-inconsist}. 
%, such as dialog systems.

Verifying the faithfulness of natural language explanations generated by self-explanatory models remains an open question. 
The adversarial framework introduced in this thesis checks whether models can generate inconsistent natural language explanations. However, further work is needed to find whether an inconsistency is due to unfaithful explanations or to a flaw in the decision-making process of the model. Moreover, future work on building more robust models that do not generate inconsistent explanations would also be very valuable. My hope is that, in the future, we will have robust and accurate neural models that faithfully explain themselves in human language.

% I hope that this work brings inspiration to the community, and that, in the near future, we will have post-hoc explanatory methods
% as well as robust neural models that faithfully explain themselves in human language.

%This thesis opens various directions of future investigations and improvements. First, future work on rigorous specifications of explanatory methods is necessary for proper usage of these methods in practice.

%this thesis encourages researchers to directly state the specifications of their explainers and verification frameworks, so that a more riguour and uni

%it draws attention to the fact that feature-based explanations requires a rigorous analysis of the limitations of such explanations. 
%An important direction of future research would be to investigate the impact that each type of 

%Secondly, the question of verifying the faithfulness of explanations is still a largely open question, both for feature-based post-hoc explanatory methods and for self-explanatory models with natural language explanations. For post-hoc explanatory methods, the framework introduced in Chapter \ref{chap-verify} can largely be improved, for example, by following the directions in Section \ref{sec:improvements}. Other types of feature-based self-explanatory methods might also be exploited to 

%Finally, there is large room for improvement in the direction of training neural models with natural language explanations and obtaining correct such explanations at test time.

%% file: appendix0.tex
\chapter{Examples from the Aroma and Appearance Sanity Tests}
\label{more-examples-sanity}

\begin{figure*}[t]
        \centering
        \includegraphics[width=0.9999\linewidth,trim={1cm 14cm 1cm 10cm},clip]{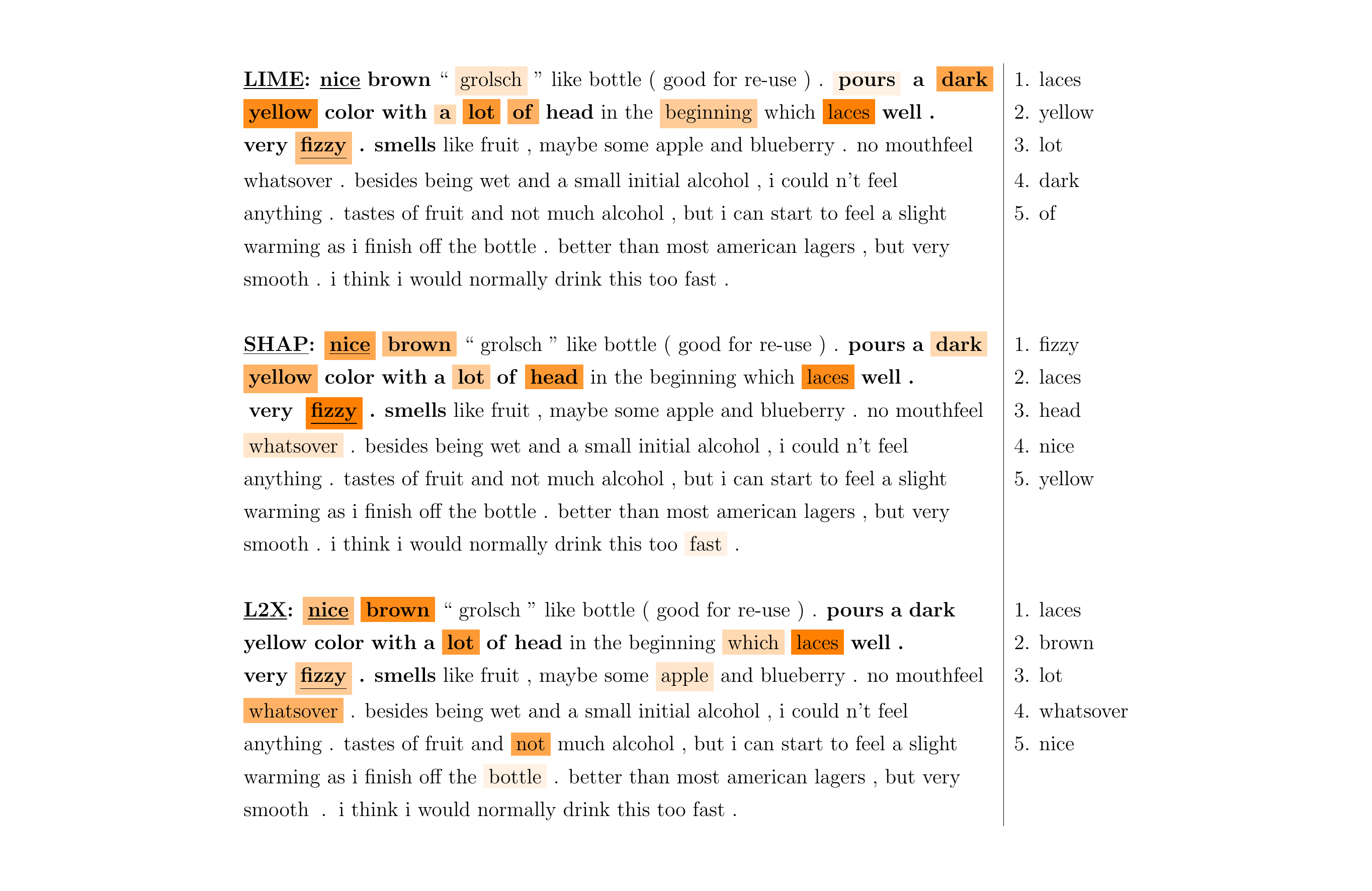}
        \caption[Explainers rankings on an instance from the appearance sanity test]{Explainers rankings (with top 5 features on the right-hand side) on an instance from the appearance aspect in our test. The selected tokens are in bold, and the independently relevant tokens are additionally underlined. The non-bold tokens should not be ranked higher than any bold and underlined token. KernalSHAP is simply referred as SHAP.
        \label{fig:qualitative_3} }
\end{figure*}

\begin{figure*}[t]
        \centering 
        \includegraphics[width=0.9999\linewidth,trim={1cm 12cm 1cm 7cm},clip]{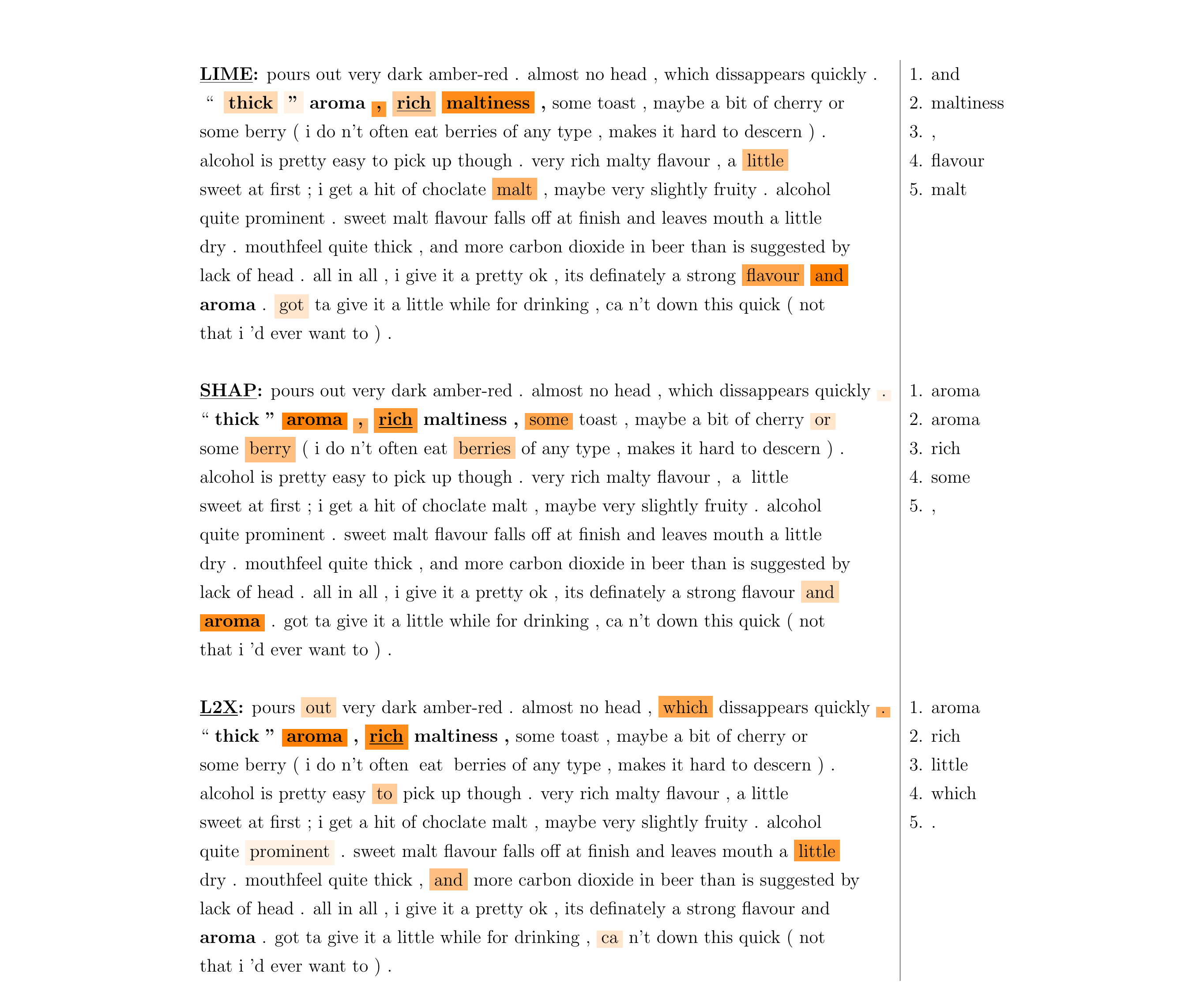}
        \caption[Explainers rankings on an instance from the aroma sanity test]{Explainers rankings (with top 5 features on the right-hand side) on an instance from the aroma aspect in our test. The selected tokens are in bold, and the independently relevant tokens are additionally underlined. The non-bold tokens should not be ranked higher than any bold and underlined token. KernalSHAP is simply referred as SHAP.
        \label{fig:qualitative_2} }
\end{figure*}

Figures \ref{fig:qualitative_3} and \ref{fig:qualitative_2} each provide an example from our sanity tests from the appearance and aroma aspects, respectively. 

In Figures \ref{fig:qualitative_3}, we see that LIME and L2X detected ``laces'' as the most important token, yet this token is an irrelevant token for the prediction. KernalSHAP however managed to detect ``fizzy'' as most important token, which appears as a better behaviour since ``fizzy'' is an independently relevant token. However, ``nice'' is also an independently relevant token, yet KernalSHAP flags it as less important than ``laces''. \vf{} penalises the explainers for these errors.

Similarly, in the example in Figure \ref{fig:qualitative_2}, we see that LIME ranks the irrelevant tokens ``and'' and ``flavour'' higher than the independently relevant token ``rich''. \vf{} penalises LIME for this. However, the fact that KernalSHAP and L2X placed ``aroma'' (with both its occurrences for SHAP) higher than ``rich'' is not penalised by \vf{} because both occurrences of ``aroma'' were selected tokens.

%% file: appendix_esnli.tex
\chapter{Templates to Filter Uninformative Explanations}
\label{uninformative-filtering}
This Appendix presents the list of templates used to filter uninformative explanations in Section \ref{partial-score}. These templates were identified by manually scanning through the dataset.

\paragraph{ }
\textbf{General templates}

$<$\textsc{premise}$>$

$<$\textsc{hypothesis}$>$ 

$<$\textsc{hypothesis}$>$ $<$\textsc{premise}$>$ 

$<$\textsc{premise}$>$ $<$\textsc{hypothesis}$>$

Sentence 1 states $<$\textsc{premise}$>$. Sentence 2 is stating $<$\textsc{hypothesis}$>$

Sentence 2 states $<$\textsc{hypothesis}$>$. Sentence 1 is stating $<$\textsc{premise}$>$

There is $<$\textsc{premise}$>$

There is $<$\textsc{hypothesis}$>$ 

\paragraph{ }
\textbf{Entailment templates} 

$<$\textsc{premise}$>$ implies $<$\textsc{hypothesis}$>$

If $<$\textsc{premise}$>$ then $<$\textsc{hypothesis}$>$

$<$\textsc{premise}$>$ would imply $<$\textsc{hypothesis}$>$

$<$\textsc{hypothesis}$>$ is a rephrasing of $<$\textsc{premise}$>$

$<$\textsc{premise}$>$ is a rephrasing of $<$\textsc{hypothesis}$>$ 

In both sentences $<$\textsc{hypothesis}$>$

$<$\textsc{premise}$>$ would be $<$\textsc{hypothesis}$>$

$<$\textsc{premise}$>$ can also be said as $<$\textsc{hypothesis}$>$

$<$\textsc{hypothesis}$>$ can also be said as $<$\textsc{premise}$>$

$<$\textsc{hypothesis}$>$ is a less specific rephrasing of $<$\textsc{premise}$>$

This clarifies that $<$\textsc{hypothesis}$>$

If $<$\textsc{premise}$>$ it means $<$\textsc{hypothesis}$>$

$<$\textsc{hypothesis}$>$ in both sentences

$<$\textsc{hypothesis}$>$ in both

$<$\textsc{hypothesis}$>$ is same as $<$\textsc{premise}$>$

$<$\textsc{premise}$>$ is same as $<$\textsc{hypothesis}$>$

$<$\textsc{premise}$>$ is a synonym of $<$\textsc{hypothesis}$>$

$<$\textsc{hypothesis}$>$ is a synonym of $<$\textsc{premise}$>$.
						
\paragraph{ }
\textbf{Neutral templates} 

Just because $<$\textsc{premise}$>$ doesn't mean $<$\textsc{hypothesis}$>$

Cannot infer the $<$\textsc{hypothesis}$>$

One cannot assume $<$\textsc{hypothesis}$>$

One cannot infer that $<$\textsc{hypothesis}$>$

Cannot assume $<$\textsc{hypothesis}$>$

$<$\textsc{premise}$>$ does not mean $<$\textsc{hypothesis}$>$

We don't know that $<$\textsc{hypothesis}$>$

The fact that $<$\textsc{premise}$>$ doesn't mean $<$\textsc{hypothesis}$>$

The fact that $<$\textsc{premise}$>$ does not imply $<$\textsc{hypothesis}$>$

The fact that $<$\textsc{premise}$>$ does not always mean $<$\textsc{hypothesis}$>$

The fact that $<$\textsc{premise}$>$ doesn't always imply$<$\textsc{hypothesis}$>$.

\paragraph{ }
\textbf{Contradiction templates}

In sentence 1 $<$\textsc{premise}$>$ while in sentence 2 $<$\textsc{hypothesis}$>$

It can either be $<$\textsc{premise}$>$ or $<$\textsc{hypothesis}$>$

It cannot be $<$\textsc{hypothesis}$>$ if $<$\textsc{premise}$>$

Either $<$\textsc{premise}$>$ or $<$\textsc{hypothesis}$>$ 

Either $<$\textsc{hypothesis}$>$ or $<$\textsc{premise}$>$ 

$<$\textsc{premise}$>$ and other $<$\textsc{hypothesis}$>$

$<$\textsc{hypothesis}$>$ and other $<$\textsc{premise}$>$

$<$\textsc{hypothesis}$>$ after $<$\textsc{premise}$>$

$<$\textsc{premise}$>$ is not the same as $<$\textsc{hypothesis}$>$

$<$\textsc{hypothesis}$>$ is not the same as $<$\textsc{premise}$>$ 

$<$\textsc{premise}$>$ is contradictory to $<$\textsc{hypothesis}$>$ 

$<$\textsc{hypothesis}$>$ is contradictory to $<$\textsc{premise}$>$ 

$<$\textsc{premise}$>$ contradicts $<$\textsc{hypothesis}$>$

$<$\textsc{hypothesis}$>$ contradicts $<$\textsc{premise}$>$

$<$\textsc{premise}$>$ cannot also be $<$\textsc{hypothesis}$>$

$<$\textsc{hypothesis}$>$ cannot also be $<$\textsc{premise}$>$

Either $<$\textsc{premise}$>$ or $<$\textsc{hypothesis}$>$ 

Either $<$\textsc{premise}$>$ or $<$\textsc{hypothesis}$>$ not both at the same time

$<$\textsc{premise}$>$ or $<$\textsc{hypothesis}$>$ not both at the same time.

%% file: appendix_inconsist.tex
\chapter{Templates Identified in e-SNLI}
\label{assec:templates}

This Appendix presents the list of templates that I manually found to match most of the e-SNLI explanations. During the collection of the dataset, no template was imposed, they were a natural consequence of the task and SNLI dataset.

``\emph{subphrase1}/\emph{subphrase2}/...'' means that a separate template is to be considered for each of the subphrases. X and Y are the key elements that we want to identify and use in the other templates in order to create inconsistencies. ``[...]'' is a placeholder for any string, and its value is not relevant. Subphrases placed between parentheses (for example, ``(the)'' or ``(if)'') are optional, and two distinct templates are formed one with and one without that subphrase.

\paragraph{Entailment Templates}

\begin{itemize}
\item \ X is/are a type of Y
\item \ X implies Y
\item \ X is/are the same as Y
\item \ X is a rephrasing of Y 
\item \ X is a another form of Y
\item \ X is synonymous with Y
\item \ X and Y are synonyms/synonymous
\item \ X and Y is/are the same thing
\item \ (if) X , then Y
\item \ X so Y
\item \ X must be Y
\item \ X has/have to be Y
\item \ X is/are Y
\end{itemize}

% NEUTRAL TEMPLATES
\paragraph{Neutral Templates}

\begin{itemize}
\item \ not all    X are Y
\item \ not every    X is Y
\item \ just because    X does not/n't mean/imply Y
\item \   X is/are not necessarily Y
\item \   X does not/n't have to be Y
\item \   X does not/n't imply/mean Y

\end{itemize}

%CONTRADICTION TEMPLATES
\paragraph{Contradiction Templates}

\begin{itemize}
\item \ ([...]) cannot/can not/ca n't (be) X and Y at the same time/simultaneously 
\item \ ([...]) cannot/can not/ca n't (be) X and at the same time Y
\item \   X is/are not (the) same as Y
\item \ ([...]) is/are either X or Y
\item \   X is/are not Y
\item \   X is/are the opposite of Y
\item \ ([...]) cannot/can not/ca n't (be) X if (is/are)~Y
\item \   X is/are different than Y
\item \   X and Y are different ([...])
\end{itemize}

\section*{Running Example}
\label{run-ex}

Below, I present a running example for creating inconsistencies by swapping between templates of explanations.

Consider the explanation $\bfe$: ``Dog is a type of animal.'' which may arise from a model explaining the instance $\bfx$: (premise: ``A dog is in the park.'', hypothesis: ``An animal is in the park.''). One identifies that $\bfe$ matches the template ``X is/are a type of Y'' with X = ``dog'' and Y = ``animal''. The list $\mathcal{I}_{\bfe}$ is then generated by replacing X and Y in each of the neutral and contradictory templates listed above with the exception of those that contain ``[...]'' in order to avoid guessing the placeholder. %$\mathcal{I}_{\bfe}$ is formed of:

\textbf{Neutral inconsistencies}
\begin{itemize}
\item \ not all dog are animal
\item \ not every dog is animal
\item \ just because dog does not/n't mean/imply animal
\item \   dog is/are not necessarily animal
\item \   dog does not/n't have to be animal
\item \   dog does not/n't imply/mean animal
\end{itemize}

\textbf{Contradiction inconsistencies}
\begin{itemize}
\item \ cannot/can not/ca n't (be) dog and animal at the same time/simultaneously 
\item \ cannot/can not/ca n't (be) dog and at the same time animal
\item \   dog is/are not (the) same as animal
\item \ is/are either dog or animal
\item \   dog is/are not animal
\item \   dog is/are the opposite of animal
\item \  cannot/can not/ca n't (be) dog if (is/are) animal
\item \   dog is/are different than animal
\item \   dog and animal are different
\end{itemize}